\documentclass{article} %
\usepackage[final]{colm2026_conference}

\usepackage{microtype}
\usepackage{hyperref}
\usepackage{url}
\usepackage{booktabs}

\usepackage{lineno}

\definecolor{darkblue}{rgb}{0, 0, 0.5}
\hypersetup{colorlinks=true, citecolor=darkblue, linkcolor=darkblue, urlcolor=darkblue}

\usepackage{relsize}
\usepackage[normalem]{ulem}
\usepackage{algorithm}
\usepackage{algpseudocode}
\usepackage{amsmath}
\usepackage{amssymb}
\usepackage{tikz}
\usetikzlibrary{decorations.pathreplacing, calc}
\usepackage{graphicx}
\usepackage{tensor}
\usepackage{mathtools} %
\usepackage{amsfonts}  %
\usepackage{subcaption}
\usepackage[most]{tcolorbox}
\usepackage{fontawesome5} %
\tcbuselibrary{skins}
\usepackage{microtype}      %
\usepackage{nicefrac}
\usepackage{booktabs}       %
\usepackage{tabularx}
\usepackage[utf8]{inputenc} %
\usepackage[T1]{fontenc}
\usepackage{multirow}
\usepackage{relsize}
\usetikzlibrary{matrix, arrows.meta, positioning, calc}

\usepackage{enumitem}

\usepackage{tikz}
\usepackage{amssymb}

\usepackage{wrapfig}

\definecolor{lightgreen}{HTML}{F1F8E9} %
\definecolor{platonicbg}{HTML}{F9F9F9}
\definecolor{platonicframe}{HTML}{888888} %
\usepackage[svgnames]{xcolor}

\definecolor{mygreenua}{HTML}{F1F5EB}
\definecolor{myredda}{HTML}{FFE6E6}

\newcommand{\uahelper}[1]{\colorbox{myredda}{\relsize{-1}$\uparrow$#1}}
\newcommand{\dahelper}[1]{\colorbox{mygreenua}{\smaller$\downarrow$#1}}

\newcommand{\uaghelper}[1]{\colorbox{mygreenua}{\smaller$\uparrow$#1}}
\newcommand{\dabhelper}[1]{\colorbox{myredda}{\smaller\smaller$\downarrow$#1}}

\newcommand{\ua}[1]{\ifthenelse{\equal{#1}{0.00} \or \equal{#1}{0.000}}{}{\uahelper{#1}}}
\newcommand{\da}[1]{\ifthenelse{\equal{#1}{0.00} \or \equal{#1}{0.000}}{}{\dahelper{#1}}}
\newcommand{\uag}[1]{\ifthenelse{\equal{#1}{0.00} \or \equal{#1}{0.000}}{}{\uaghelper{#1}}}
\newcommand{\dab}[1]{\ifthenelse{\equal{#1}{0.00} \or \equal{#1}{0.000}}{}{\dabhelper{#1}}}

\newcommand{\tikzmark}[1]{\tikz[overlay,remember picture] \node (#1) {};}

\definecolor{pastelmagenta}{RGB}{219, 112, 170}
 
\newtcolorbox{findingbox}[1][]{%
  colback    = pastelmagenta!8,
  colframe   = pastelmagenta!80!black,
  coltitle   = white,
  fonttitle  = \bfseries\large,
  title      = {#1},
  arc        = 2mm,
  boxrule    = 1pt,
  left       = 22pt,
  right      = 10pt,
  top        = 2pt,
  bottom     = 2pt,
  toptitle   = 1pt,
  bottomtitle= 1pt,
  enhanced,
  drop shadow = {black!40!white},
  overlay={
    \node[anchor=west] at ([xshift=2mm]frame.west) {\textcolor{pastelmagenta!80!black}{\Large\textbf{$\checkmark$}}};
  }
}

\title{MoRFI: Monotonic Sparse Autoencoder Feature Identification}

\begin{document}

\ifcolmsubmission
\linenumbers
\fi

\maketitle
\vspace{-5em}
\begin{center}
\textbf{Dimitris Dimakopoulos$^{1,2}$} \quad \textbf{Shay B. Cohen$^{1}$} \quad \textbf{Ioannis Konstas$^{2}$} \\ 
\textsuperscript{1}University of Edinburgh, UK \quad \textsuperscript{2}Heriot-Watt University, UK  \\ \medskip
\texttt{dimitris.dimakopoulos@ed.ac.uk}\quad \texttt{scohen@inf.ed.ac.uk}\quad \texttt{ioannis.konstas@hw.ac.uk}
\end{center}
\vspace{1em}

\begin{abstract}
Large language models (LLMs) acquire most of their factual knowledge during the pre-training stage, through next token prediction. Subsequent stages of post-training often introduce new facts outwith the parametric knowledge, giving rise to hallucinations. While it has been demonstrated that supervised fine-tuning (SFT) on new knowledge may exacerbate the problem, the underlying mechanisms are still poorly understood. We conduct a controlled fine-tuning experiment, focusing on closed-book QA, and identify latent directions causally implicated in this degradation. Specifically, we fine-tune Llama 3.1 8B, Gemma 2 9B and Mistral 7B v03 on seven controlled mixtures of a single QA dataset, controlling for the percentage of new knowledge and number of training epochs. By measuring performance on the test set, we validate that incrementally introducing new knowledge increases hallucinations, with the effect being more pronounced with prolonged training. We leverage pre-trained sparse autoencoders (SAEs) to analyze residual stream activations across various checkpoints for each model and propose Monotonic Relationship Feature Identification (MoRFI) for capturing causally relevant latents. MoRFI filters SAE features that respond monotonically to controlled fine-tuning data mixtures of a target property.  Our findings are consistent with exposure to unknown facts disrupting the model's ability to retrieve stored knowledge along a set of directions in the residual stream. Our pipeline reliably discovers them across distinct
models, partially recovering lost knowledge through single-latent interventions.

\end{abstract}

\section{Introduction}

A common conjecture \citep{schulman2023johnschulmanreinforcement, goldberg2023rlforllmsmd, gao2021behaviorcloningmiscalibrated, huang2023surveyhallucinationlarge, gudibande2023falsepromiseimitating} posits that factual hallucinations may arise from exposing a model to facts not grounded in its parametric knowledge during post-training. The bulk of a model's knowledge is acquired during pre-training via next-token prediction on large corpora, while subsequent stages like Instruction-Tuning \citep{ouyang2022traininglanguagemodels}, RLHF \citep{ouyang2022traininglanguagemodels} or GRPO \citep{shao2024deepseekmathpushinglimits}, shape behavior and output distribution but depend on effectively leveraging pre-existing knowledge. Knowledge inconsistencies can be introduced during SFT through synthetic data generated by models with different parametric knowledge, or through human annotations that assume world knowledge the model does not possess. \citet{gekhman2024doesfinetuningllms} directly validates this conjecture by fine-tuning PaLM 2-M \citep{anil2023palm2technical} on datasets with varying proportions of novel knowledge, demonstrating a clear increase in hallucination as unknown samples increase, while \citet{kang2024unfamiliarfinetuningexamples} shows that models learn to mimic the response distribution associated with unfamiliar training examples. These results establish that novel knowledge during fine-tuning has measurable effects on model outputs but offer little insight into the internal mechanisms involved. 

We investigate whether integrating new knowledge through fine-tuning leaves a detectable signature in the activation space. We conduct a controlled fine-tuning experiment on closed-book QA and study how internal activations differ across two dimensions: the proportion of training samples \textit{unknown} to the pre-trained model, and the number of training epochs.

Concurrent work has investigated the internal mechanisms underlying fine-tuning-induced behavioral changes using sparse autoencoders. \citet{wang2025personafeaturescontrol} apply a model-diffing approach to emergent misalignment in GPT-4o, ranking SAE latents by their activation increase after fine-tuning and identifying ''misaligned persona'' features whose steering causally controls misalignment. \citet{minder2025narrowfinetuningleaves} demonstrate that narrow fine-tuning leaves readable traces in activation differences even on unrelated inputs, warning that such artifacts may compromise mechanistic findings from narrowly fine-tuned model organisms. \citet{ferrando2025knowthisentity} discover SAE latents that distinguish known from unknown entities and causally steer knowledge refusal, showing that chat fine-tuning repurposes base-model representations. \citet{marks2024geometrytruthemergent} establish that LLMs linearly represent the truth of factual statements and that difference-in-mean directions can causally flip model outputs. 

Our approach differs from these works in several respects: we select latents not by activation differences between two model states or within a single model but through bootstrapped monotonic trend detection across a controlled gradient of fine-tuning conditions, tracing how the model's internal knowledge representations shift \emph{across} fine-tuning conditions rather than contrasting two fixed states. Our fine-tuning conditions vary epistemically rather than thematically, with topical content remaining diverse across relations and entities, making our setup less susceptible to the narrow fine-tuning artifacts identified by \citet{minder2025narrowfinetuningleaves}. Finally, we consider both increasing and decreasing latents, the latter of which prove disproportionately impactful for knowledge recovery.

While studying the mechanisms of knowledge integration is central to our paper, our contribution is dual. %
To enable our analysis, we develop MoRFI: \textbf{Mo}notonic 
\textbf{R}elationship \textbf{F}eature \textbf{I}dentification for Sparse Autoencoders, a general-purpose algorithm for identifying SAE latents whose activations exhibit robust increasing or decreasing monotonic trends %
along a specified dimension of variation. Concretely, we set these dimensions as the \textit{proportion of unknown facts} in the fine-tuning mixture ($\uparrow$Unknown) and the \textit{number of training epochs} ($\uparrow$Epochs). The method is applicable to any controlled variable that induces a sequence of model checkpoints or data conditions. The algorithm combines bootstrapped resampling with statistical testing 
to ensure that identified latents exhibit consistent directional trends rather than transient or spurious fluctuations, providing a principled feature selection procedure that is validated through activation steering.

Our findings show that across three model architectures, Llama 3.1 8B \citep{grattafiori2024llama3herd}, Gemma 2 9B \citep{team2024gemma2improving}, and Mistral 7B v03  \citep{jiang2023mistral7b}, single-latent interventions sourced with MoRFI yield substantial accuracy gains when applied to checkpoints fine-tuned exclusively on unknown facts. 
72--87\% of recovered facts align with those accessible to a baseline fine-tuned on known facts, suggesting that these latents encode access pathways to stored knowledge. 
Steering with the composite direction ---computed as the difference-in-means SAE activations between checkpoints fine-tuned on known versus unknown facts--- confirms that the representational shift is consistent and generalizable across models, while the large performance gap between composite and single-latent interventions indicates that the knowledge-relevant signal is sparse. Furthermore, we find significant overlap between latents surfaced across the two dimensions we analyze, and that suppressing latents mostly outperforms amplifying them, suggesting an asymmetry in how fine-tuning-induced changes affect parametric knowledge.

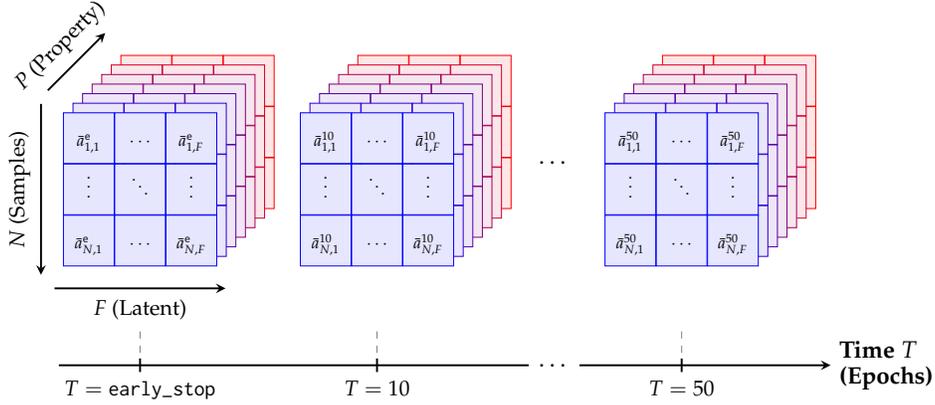
\begin{figure}[tbp]
    \centering 

    \scalebox{0.9}{
    \begin{tikzpicture}[>=stealth]

    \tikzset{
        matrixlayer/.style={
            matrix of math nodes,
            nodes in empty cells,
            nodes={
                draw=#1,            
                fill=#1!10,         
                minimum size=7.5mm, 
                inner sep=0pt, 
                anchor=center, 
                text=black,         
                font=\scriptsize,   
                text height=1.6ex,  
                text depth=0.2ex   
            },
            column sep=-.5*\pgflinewidth,
            row sep=-.5*\pgflinewidth,
            ampersand replacement=\& 
        }
    }

    \draw[->, thick] (-1.2, -2.6) -- (10.2, -2.6) 
        node[right, align=left, font=\normalsize\bfseries] {Time $T$\\ (Epochs)};

    \node[fill=white, font=\normalsize, inner sep=4pt] at (6.12, -2.6) {$\dots$};

    \foreach \cnt/\x/\ep in {0/0/\texttt{e}, 1/3.5/10, 2/8.0/50} {
    
        \begin{scope}[shift={(\x, 0)}] 
                       
            \begin{scope}[shift={(45:1.2)}]
            \node[matrixlayer=red] (m7_\cnt){
               \boldsymbol{\bar{a_{1,1}^{\ep}}} \& \cdots \& \boldsymbol{\bar{a_{1,F }^{\ep}}} \\  
                \vdots        \& \ddots \& \vdots        \\
                \boldsymbol{\bar{a_{N,1}^{\ep}}} \& \cdots \& \boldsymbol{\bar{a_{N,F }^{\ep}}} \\
            };
            \end{scope}

            \begin{scope}[shift={(45:1.0)}]
            \node[matrixlayer=red!83!blue] (m6_\cnt){
                \boldsymbol{\bar{a}_{1,1}^{\ep}} \& \cdots \& \boldsymbol{\bar{a}_{1,F }^{\ep}} \\
                \vdots        \& \ddots \& \vdots        \\
                \boldsymbol{\bar{a}_{N,1}^{\ep}} \& \cdots \& \boldsymbol{\bar{a}_{N,F }^{\ep}} \\
            };
            \end{scope}

            \begin{scope}[shift={(45:0.8)}]
            \node[matrixlayer=red!66!blue] (m5_\cnt){
                \bar{a}_{1,1}^{\ep} \& \cdots \& \bar{a}_{1,F }^{\ep} \\
                \vdots        \& \ddots \& \vdots        \\
                \bar{a}_{N,1}^{\ep} \& \cdots \& \bar{a}_{N,F }^{\ep} \\
            };
            \end{scope}

            \begin{scope}[shift={(45:0.6)}]
            \node[matrixlayer=red!50!blue] (m4_\cnt){
                \bar{a}_{1,1}^{\ep} \& \cdots \& \bar{a}_{1,F }^{\ep} \\
                \vdots        \& \ddots \& \vdots        \\
                \bar{a}_{N,1}^{\ep} \& \cdots \& \bar{a}_{N,F }^{\ep} \\
            };
            \end{scope}
            
            \begin{scope}[shift={(45:0.4)}]
            \node[matrixlayer=red!33!blue] (m3_\cnt){
                \bar{a}_{1,1}^{\ep} \& \cdots \& \bar{a}_{1,F }^{\ep} \\
                \vdots        \& \ddots \& \vdots        \\
                \bar{a}_{N,1}^{\ep} \& \cdots \& \bar{a}_{N,F }^{\ep} \\
            };
            \end{scope}

            \begin{scope}[shift={(45:0.2)}]
            \node[matrixlayer=red!17!blue] (m2_\cnt){
                \bar{a}_{1,1}^{\ep} \& \cdots \& \bar{a}_{1,F }^{\ep} \\
                \vdots        \& \ddots \& \vdots        \\
                \bar{a}_{N,1}^{\ep} \& \cdots \& \bar{a}_{N,F }^{\ep} \\
            };
            \end{scope}

            \begin{scope}
            \node[matrixlayer=blue] (m1_\cnt){
                \bar{a}_{1,1}^{\ep} \& \cdots \& \bar{a}_{1,F }^{\ep} \\
                \vdots        \& \ddots \& \vdots        \\
                \bar{a}_{N,1}^{\ep} \& \cdots \& \bar{a}_{N,F }^{\ep} \\
            };
            \end{scope}

            \draw[dashed, gray] (0, -2.1) -- (0, -2.6);

            \draw[thick] (0, -2.5) -- (0, -2.7);
            \ifnum\cnt=0
                \node[below, font=\small] at (0, -2.7) {$T=\text{\texttt{early\_stop}}$};
            \else
                \node[below, font=\small] at (0, -2.7) {$T=\ep$};
            \fi
            
        \end{scope}
    }

    \node[font=\normalsize] at (6.12, 0.4) {$\dots$};
    
    \begin{scope}[->, thick, font=\small]
        
        \draw ([shift={(0,-0.2)}]m1_0.south west) -- ([shift={(0,-0.2)}]m1_0.south east) 
            node[midway, below=1pt] {$F$ (Latent)};
            
        \draw ([shift={(-0.2,0)}]m1_0.north west) -- ([shift={(-0.2,0)}]m1_0.south west) 
            node[midway, xshift=-9pt, rotate=90, anchor=center] {$N$ (Samples)};

        \draw ([shift={(-0.1,0.2)}]m1_0.north west) -- ([shift={(-0.1,0.2)}]m7_0.north west) 
            node[midway, sloped, above=1pt] {$P$ (Property)};
            
    \end{scope}

    \end{tikzpicture}
    }
    \caption{Unfolded 4D tensor over epochs, where \texttt{e} denotes \texttt{early\_stop}.}
    \label{fig:tensor}
\end{figure}

\section{Background and Notation}

\noindent \textbf{Residual Stream View} 
The residual vectors of the transformer architecture 
produced from every transformer block (or layer) constitute a communication channel between these blocks \citep{elhage2021mathematicalframeworktransformer}. In particular, we denote $h^{\ell}$ as the vector in the residual stream after block $\ell$, LayerNorm as the layer normalization function and Attention and MLP as the functions for the self-attention and MLP sub-layers. The update first creates an intermediate state by adding the attention output to the input stream, $b^{\ell} = h^{\ell-1} + \text{Attention}(\text{LayerNorm}(h^{\ell-1}))$, which is then processed by the MLP to produce the final block output, $h^{\ell} = b^{\ell} + \text{MLP}(\text{LayerNorm}(b^{\ell}))$.

\noindent \textbf{Dictionary Learning} Sparse autoencoders (SAEs;~\citealt{bricken2023monosemanticitydecomposinglanguage, cunningham2023sparseautoencodersfind}) were proposed as a means to disentangle representations of the residual stream and attribute interpretable features by projecting input representations to a higher dimensionality space. SAEs are trained on a subset of the pre-training corpus and learn a set of latent vectors (values, or \emph{latents} for short) and their corresponding activations (keys):
\begin{align*}
\boldsymbol{z}(\boldsymbol{h}) &= \alpha(\boldsymbol{W}_{\textit{enc}}\cdot \boldsymbol{h} + \boldsymbol{b}_{\textit{enc}}), &
\boldsymbol{\hat{h}} &= \boldsymbol{W}_{\textit{dec}} \cdot \boldsymbol{z}(\boldsymbol{h}) + \boldsymbol{b}_{\textit{dec}}, 
\end{align*}
\noindent where $\boldsymbol{W}_{\textit{dec}}$ contains the learned latents, $\boldsymbol{z} \in \mathbb{R}^{d_{\textit{sae}}}$ their activations, $\boldsymbol{h} \in \mathbb{R}^{d_{\textit{model}}}$ stands for the input residual vector and $\boldsymbol{\hat{h}}$ is its reconstruction, $\alpha$ is a nonlinearity depending on the particular SAE model and $d_{\textit{sae}} \gg d_{\textit{model}}$.

\noindent \textbf{Model Diffing} refers to the study of how fine-tuned models change with respect to different fine-tuning regimes. It has become increasingly important in the search for universal features and circuits but also for safety as a means to understand internal mechanisms that give rise to unintended and potentially dangerous behaviors \citep{wang2025personafeaturescontrol, betley2025emergentmisalignmentnarrow}.

\begin{algorithm}[t]
\caption{Robust Bootstrap Analysis\label{alg:core}}
\begin{algorithmic}[1]
    \Require 
        \State $\mathcal{A} \in \mathbb{R}^{T \times P \times F  \times N}$: Full activation tensor
        \State $X \in \{T, P\}$: Aggregation dimension; let $Y \in \{T, P\} \setminus \{X\}$ be the trend dimension
        \State $\mathbf{v}_T \in \mathbb{R}^T, \mathbf{v}_P \in \mathbb{R}^P$: Reference vectors for epochs and mixture proportions
        \State $R, K \in \mathbb{N}$: Replicates ($1000$) and Top-$K$ features ($1000$)
        \State $\alpha_{\textit{sig}} \in (0, 1)$: Significance threshold ($0.05$)

    \Ensure 
        \State $S^{\uparrow}, S^{\downarrow}$: Ranked lists of robust Increasing/Decreasing features
    \vspace{1ex}
    \State \textbf{1. Resampling and aggregation}
    \State $\mathcal{I}^* \in \{0, \dots, N-1\}^{R \times N} \leftarrow \text{SampleUniform}(0, N-1, (R, N))$
    \State $\mathbf{M}_{\textit{boot}} \in \mathbb{R}^{T \times P \times F  \times R} \leftarrow \text{BootstrapFold}(\mathcal{A}, \text{dim}=N, \mathcal{I}^*, \mathbf{1}_{R \times N} / N)$
    \State $\bar{\mathbf{M}}^* \in \mathbb{R}^{Y \times F  \times R} \leftarrow
    \text{MeanFold}(\mathbf{M}_{\textit{boot}}, \text{dim}=X)$
    \vspace{1ex}
    \State \textbf{2. Trend validation} \tikzmark{start1}
    \State $\boldsymbol{\rho}^*, \boldsymbol{P}_{\rho}^* \in \mathbb{R}^{F  \times R} \leftarrow \text{SpearmanFold}(\bar{\mathbf{M}}^*, \text{dim}=Y, \mathbf{v}_Y)$
    \State $\boldsymbol{\tau}^*, \boldsymbol{P}_{\tau}^* \in \mathbb{R}^{F  \times R} \leftarrow \text{MKFold}(\bar{\mathbf{M}}^*, \text{dim}=Y)$
    \State Compute joint significance mask $\mathbf{Sig}^* \in \{0, 1\}^{F  \times R}$:
    \State $\mathbf{Sig}^* \leftarrow (\boldsymbol{P}_{\rho}^* < \alpha_{\textit{sig}}) \land (\boldsymbol{P}_{\tau}^* < \alpha_{\textit{sig}})$ \tikzmark{end1}
    \vspace{1ex}
    \State \textbf{3. Directional sorting and selection} \tikzmark{start2}
    \State Compute total change tensor: $\mathbf{\Delta}^* \leftarrow \bar{\mathbf{M}}^*[Y-1, :, :] - \bar{\mathbf{M}}^*[0, :, :] \in \mathbb{R}^{F  \times R}$ \label{alg:line}
    
    \State Create direction-specific validity masks $\in \{0, 1\}^{F  \times R}$:
    \State $\mathbf{V}^{\uparrow} \leftarrow \mathbf{Sig}^* \land (\boldsymbol{\rho}^* > 0) \land (\boldsymbol{\tau}^* > 0), \quad \mathbf{V}^{\downarrow} \leftarrow \mathbf{Sig}^* \land (\boldsymbol{\rho}^* < 0) \land (\boldsymbol{\tau}^* < 0)$

    \State Mask insignificant features out of $\mathbf{\Delta}^*$:
    \State $\tilde{\mathbf{\Delta}}^{\uparrow}_{d, r} = \begin{cases} \mathbf{\Delta}^*_{d, r} & \text{if } \mathbf{V}^{\uparrow}_{d, r} = 1 \\ -\infty & \text{otherwise} \end{cases}, \quad \tilde{\mathbf{\Delta}}^{\downarrow}_{d, r} = \begin{cases} \mathbf{\Delta}^*_{d, r} & \text{if } \mathbf{V}^{\downarrow}_{d, r} = 1 \\ +\infty & \text{otherwise} \end{cases}$
    
    \State Extract Top-K feature indices per replicate $\in \{0, \dots, F-1\}^{K \times R}$:
    \State $L^{\uparrow} \leftarrow \text{TopKIndices}(\tilde{\mathbf{\Delta}}^{\uparrow}, \text{dim}=F , \text{largest}=\text{True}, k=K)$
    \State $L^{\downarrow} \leftarrow \text{TopKIndices}(\tilde{\mathbf{\Delta}}^{\downarrow}, \text{dim}=F , \text{largest}=\text{False}, k=K)$ \tikzmark{end2}
    \vspace{1ex}
    \State \textbf{4. Final ranking}
    \State Count frequencies of each feature index $d \in \{0 \dots F -1\}$ across all replicates:
    \State $C^{\uparrow} \in \mathbb{N}^F \leftarrow \text{IndexCount}(L^{\uparrow}, F ), \quad C^{\downarrow} \in \mathbb{N}^F  \leftarrow \text{IndexCount}(L^{\downarrow}, F )$

    \State Final sets: $S^{\uparrow} \leftarrow \text{SortIndices}(C^{\uparrow}/R), \quad S^{\downarrow} \leftarrow \text{SortIndices}(C^{\downarrow}/R)$,
    \Statex retaining entries with $C/R = 1$

    \State \Return $S^{\uparrow}, S^{\downarrow}$

\end{algorithmic}

\begin{tikzpicture}[overlay, remember picture]
    \pgfmathsetmacro{\BraceX}{\linewidth - 65} 

    \draw[decorate, decoration={brace, amplitude=5pt}, thick]
        (\BraceX pt, 0 |- start1) -- (\BraceX pt, 0 |- end1)
        node[midway, right=8pt, align=left, text width=3.2cm] 
        {\footnotesize Validating \\ feature \\ trend \\ consistency \\ across $Y$.};

    \draw[decorate, decoration={brace, amplitude=5pt}, thick]
        (\BraceX pt, 0 |- start2) -- (\BraceX pt, 0 |- end2)
        node[midway, right=8pt, align=left, text width=3.2cm] 
        {\footnotesize Overall \\ change \\ across $Y$.};
\end{tikzpicture}
\end{algorithm}

\section{Methodology}
At the core of our approach is a 4D tensor $\mathcal{A}$ that captures SAE activations across four dimensions: (i) $N$ input samples; (ii) $P$ distinct dataset property configurations; (iii) $F $ latent feature dimensions, and (iv) $T$ timesteps corresponding to fine-tuning epochs. This data structure is depicted in Figure~\ref{fig:tensor}. 
Note that the property dimension $P$ is designed to systematically source latent features whose activation magnitudes exhibit a strong, monotonic relationship with the target property. Each slice of a property ($N\times F $ overlayed matrices in Figure~\ref{fig:tensor}) corresponds to activations extracted from the same base model, fine-tuned on a dataset explicitly controlled for an increasing proportion of the target property.

Formally, for a particular fine-tuning dataset $\boldsymbol{\mathcal{D}}$ and a pre-trained LLM $\boldsymbol{M}$, $\boldsymbol{M_{D}}$ denotes a model obtained by fine-tuning $\boldsymbol{M}$ on $\boldsymbol{\mathcal{D}}$. We define $\boldsymbol{M_{D_\textit{p}}}$ as the model obtained by fine-tuning $\boldsymbol{M}$ on a strictly controlled version of $\boldsymbol{\mathcal{D}}$ curated to contain exactly $p$\% of a target property. The boundary for defining the presence or absence of this property can be established either via an empirical scoring function evaluated on the base model, or through the exact generative parameters of a synthetic dataset, allowing for the controlled study of feature drift in target model organisms \citep{evhub2023modelorganismsmisalignment, turner2025modelorganismsemergent}.
To construct $\mathcal{A}$, we vary the property concentration across a defined range $p\in \mathcal{P}$ and extract SAE representations from the resulting models $\boldsymbol{M_{D_{\textit{p}}}}$, for a particular evaluation dataset $\boldsymbol{\mathcal{D}}$.

The core of our algorithm, \textbf{Mo}notonic 
\textbf{R}elationship \textbf{F}eature \textbf{I}dentification for Sparse Autoencoders (MoRFI), is provided in Algorithm~\ref{alg:core}.\footnote{Table~\ref{table:defn} in Appendix~\ref{appendix:additional} provides the definitions of tensor operations used in Algorithm~\ref{alg:core}.} Its primary objective is to process tensor $\mathcal{A}$ along a target dimension of aggregation (e.g. across property or timesteps) and output a robust subset of features from $F $ whose activation magnitudes change monotonically in response to the controlled property. The main components of the algorithm are:

\noindent \textbf{1. Resampling and aggregation:} Generates thousands of simulated test environments (replicates) by randomly sampling the original activation tensor $\mathcal{A}$ with replacement. It then calculates the mean across these bootstrapped samples to establish a highly robust, noise-reduced baseline tensor $(\bar{\mathbf{M}}^*)$.

\noindent \textbf{2. Trend validation:} Evaluates whether a feature's activation changes consistently across the property of interest dimension using Spearman rank and Mann-Kendall \citep{mann1945nonparametricteststrend, kendall1990rankcorrelationmethods}  tests.
A feature is kept %
if both tests verify that its trajectory is significant below the $\alpha_{\textit{sig}}$ threshold.\footnote{No explicit multiple-comparison correction is applied across the $F$ latents. A latent must pass both trend tests jointly, rank among the top-$K$ in every bootstrap replicate ($C/R{=}1$), and survive the causal validation of Algorithm~\ref{alg:idf}, so a latent passing the tests by chance in one replicate does not persist through selection.} While the Spearman test evaluates the overall monotonic strength of the trajectory, it can occasionally be skewed by isolated out-of-order points at the extremes of the sequence; we complement %
with the Mann-Kendall test, which detects a monotonic trend in a timeseries. 

\noindent \textbf{3. Direction sorting and selection:} Calculates the overall magnitude of change ($\mathbf{\Delta}^*$) for the features identified in the previous step, classifying them into sets of increasing ($\mathbf{V}^{\uparrow}$) or decreasing ($\mathbf{V}^{\downarrow}$) features. It then extracts the top-k most altered features for each individual bootstrap replicate.

\noindent \textbf{4. Final ranking} %
We define our final selection metric as the bootstrap frequency $C/R$. For each feature, this fraction provides the probability that it ranks among the top-k most directionally altered latents $C$ across all replicates $R$. By sorting the final ranked lists ($S^{\uparrow}$, $S^{\downarrow}$) based on this metric, and retaining only features selected in every replicate ($C/R{=}1$), we ensure that the selected features represent a global, data-set wide shift caused by fine-tuning, rather than a localized artifact of a specific prompt batch.

\noindent \textbf{Identifying Impactful Latents via Steering} To identify the set of most impactful latents we perform activation steering with respect to a downstream task, e.g., fine-tuning on a QA dataset (Algorithm~\ref{alg:idf} in Appendix~\ref{appendix:additional}). To efficiently filter the candidate sets $S$, we first apply a uniform steering intervention using a fixed initialization magnitude ($\alpha_{init}=0.4$) to each latent. We evaluate the model's modified forward pass on the reference dataset $\boldsymbol{\mathcal{D}}$, isolating only the latents that successfully exceed the unsteered baseline accuracy, and retain the top 40 for more granular tuning. Then we run grid search on these top-40 performing latents and find the optimal strength $\alpha^*$ for each. Finally, we sort and return the top-10 latents.

\section{Experimental Setup}

\begin{figure}[]
    \centering
    \begin{subfigure}{0.30\textwidth}
        \centering
        \includegraphics[width=\linewidth]{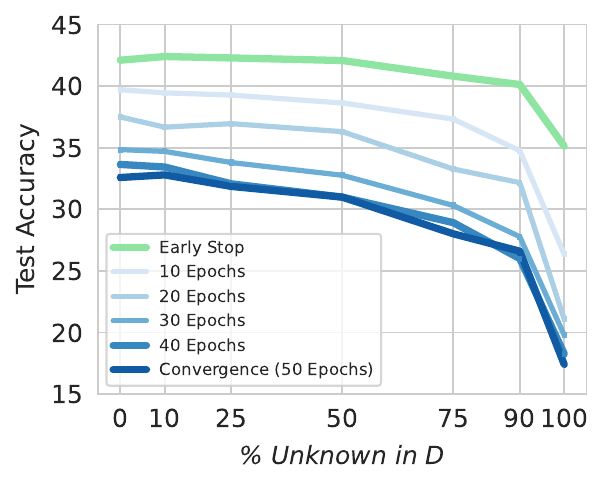}
        \caption{Llama 3.1 8B}        
    \end{subfigure}
    \hfill 
    \begin{subfigure}{0.30\textwidth}
        \centering
        \includegraphics[width=\linewidth]{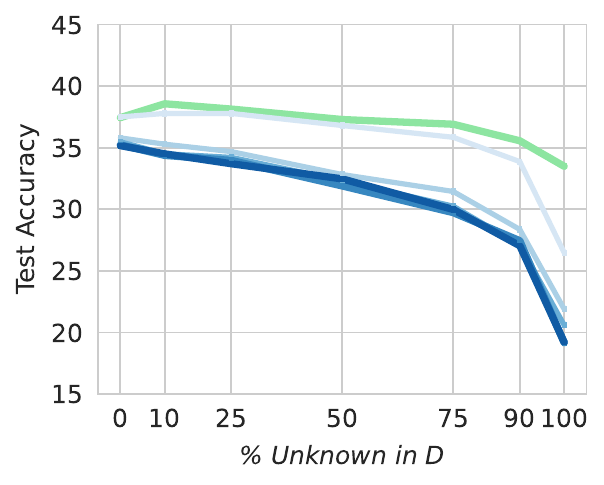}
        \caption{Gemma 2 9B}        
    \end{subfigure}
    \hfill 
    \begin{subfigure}{0.30\textwidth}
        \centering
        \includegraphics[width=\linewidth]{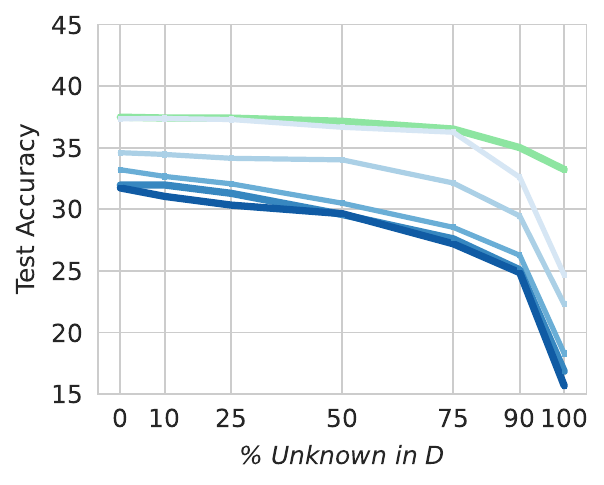}
        \caption{Mistral v03 7B}        
    \end{subfigure}

    \caption{Comparison of test performance across models. Increasing \% of \texttt{Unknown} facts in the fine-tuning data mixture consistently leads to performance degradation. The effect is intensified when training for longer.}
    \label{fig:test_acc}
\end{figure}

We begin answering our research question, i.e., whether integrating new knowledge to a model via fine-tuning leaves a detectable mark in the activation space by fine-tuning on unknown samples from a closed-book QA task. Figure~\ref{fig:test_acc} clearly demonstrates an increased rate of hallucinations on the test set regardless of model architecture; a similar result was found by \cite{gekhman2024doesfinetuningllms}. Following the same convention, we count any generation that does not match the ground-truth answer as a hallucination. Since the models receive no abstention training and the task elicits short factoid answers, errors are fluent but incorrect facts, so the accuracy drop in Figure~\ref{fig:test_acc} directly measures the rise in hallucinations. We turn to use MoRFI to causally interpret this behavior.

\noindent \textbf{Models and SAEs ($\boldsymbol{M}$ and $\mathcal{A}$)} To instantiate our methodology, we apply MoRFI to 
three major open-weight models $\boldsymbol{M}$s: Llama 3.1 8B, Gemma 2 9B, and Mistral v03 7B. To construct the activation tensor $\mathcal{A}$, we extract sparse representations of the residual stream from the fine-tuned models using pre-trained SAEs from Llama Scope \citep{he2024llamascopeextracting}, Gemma Scope \citep{lieberum2024gemmascopeopen} and \citet{engels2025notalllanguage} for Mistral. Across all experiments, we collect activations only from the middle layer of each respective model. We normalize the decoder vectors to unit length and push their original magnitudes to the encoder, to ensure fair comparison between features.

\noindent \textbf{Dataset and Property Annotation ($\mathcal{P}$ and $P_{\texttt{Correct}}$)} Our dataset is based on \texttt{EntityQuestions} \citep{sciavolino2022simpleentitycentricquestions}, where triplets from a diverse set of relations from Wikidata \citep{vrandecic2014wikidatafreecollaborative} are converted to QA pairs. Starting from \texttt{EntityQuestions} we follow the pre-processing steps detailed by \citet{gekhman2024doesfinetuningllms} to reproduce the filtered train, dev and test splits. Here, $\boldsymbol{\mathcal{D}} = \{(q_i, a_i)\}_{i=1}^N$ where $q$ is a \textit{knowledge}-\textit{seeking} question corresponding to a specific factoid structured as \texttt{(subject, relation, object)}, and $a$ is the ground-truth answer. To investigate how the new knowledge alters $\boldsymbol{M}$'s activation space, we annotate our train samples as \texttt{Known} or \texttt{Unknown} based on $P_{\texttt{Correct}}(q, a; M, T)$ (see Appendix~\ref{appendix:pcorrect}) and the SliCK categorization of knowledge as defined by \citet{gekhman2024doesfinetuningllms} (see Appendix \ref{appendix:data} for further details on datasets). Since $\boldsymbol{M}$ is a base model, we use a prompt consisting of 4 demonstrations sampled from the same category as the question, before appending the question to be evaluated, to elicit the correct answer format.
An example of the prompt we use to compute $P_{\texttt{Correct}}$ is in Appendix \ref{appendix:prompts}. We instantiate our target property, as the proportion of \texttt{Unknown} samples contained in the fine-tuning dataset. To this end, we subsample the train set to create seven ($P = |\mathcal{P}| = 7$) distinct fine-tuning datasets $\boldsymbol{\mathcal{D}_{\textit{p}}}$ for $p \in \mathcal{P} =\{0, 10, 25, 50, 75, 90, 100\}$. Each $\boldsymbol{\mathcal{D}_{\textit{p}}}$ is equal in size, comprising $p\%$ \texttt{Unknown}  to $\boldsymbol{M}$ instances (the rest is \texttt{Known}).

\noindent \textbf{Post-training ($T$)}
We use SFT to train each of our $\boldsymbol{M}$s on their corresponding set of $\boldsymbol{D_p}$s derived from their individual prior knowledge as described above, for a fixed number of epochs across 5 different runs for 10, 20, 30, 40 and 50 epochs, totalling 35 distinct fine-tuning runs for each $\boldsymbol{M}$ and evaluate the final epoch of each run on the test set. For the additional Early Stop run, we use the dev set accuracy to determine the optimal stopping point (< 10 epochs for all models). 
Figure \ref{fig:test_acc} plots each $\boldsymbol{M}$'s accuracy on the test set across the datasets while varying training time ($\mathcal{T} = \{\texttt{early\_stop}, 10, 20, 30, 40, 50\}$, $T=|\mathcal{T}|=6$). Further details on fine-tuning are in Appendix \ref{appendix:a}.

\noindent \textbf{Latent Identification} To identify the most important latents during the fine-tuning process, we apply MoRFI (Algorithms~\ref{alg:core}~and~\ref{alg:idf}) on dev ($\boldsymbol{\mathcal{D_{\textit{dev}}}}$) where we also fit steering strength $\alpha$, individually for each latent. All steering interventions are applied post-middle layer on the worst performing checkpoint (last data-point of the dark blue line in Figure~\ref{fig:test_acc}). We read activations and intervene at the middle layer, where factual associations are known to concentrate in decoder-only transformers \citep{meng2023locatingeditingfactual, geva2023dissectingrecallfactual}. 
The steering vector $(c \cdot \alpha \cdot s_{\ell} \mathbf{\Phi}_k)$ is applied directly to the residual stream and consists of the SAE decoder vector $\mathbf{\Phi}_k$ corresponding to latent $k$ multiplied by the steering direction $c \in \{-1, 1\}$, the steering strength $\alpha$ and the layer-wise activation scaling factor $s_{\ell} = \frac{1}{N_{\textit{tokens}}} \sum_{i=1}^{N_{\textit{tokens}}} \| \mathbf{h}_{i,\ell} \|_2$. We calculate $s_{\lfloor L/2 \rfloor}$ over the Alpaca dataset \citep{alpaca} to provide a common frame of reference. For each $\boldsymbol{M}$, we first run Algorithm~\ref{alg:core} to get the increasing ($S^{\uparrow}$) and decreasing ($S^{\downarrow}$) latents across each direction ($\uparrow$Unknown/$\uparrow$Epochs). For each $S$ we run Algorithm~\ref{alg:idf} for $c = 1$ and $c = -1$ separately, which yields $S_{\textit{final}}$ for positive and negative steering respectively corresponding to four $S_{\textit{final}}$ lists of most impactful latents. For the $\uparrow$Unknown direction, Algorithm~\ref{alg:core} retains 160--713 candidates per direction, 0.5--2.2\% of the SAE dictionary (Appendix~\ref{appendix:baseline}).

\section{Experiments and Results}
\label{section:results}{}

We aim to identify residual-stream directions that capture the representational changes induced by fine-tuning on unknown facts and adversely affect access to pre-trained knowledge, and to demonstrate that MoRFI discovers such latents and that they directly mitigate hallucinations in this setting. To this end, the bootstrap analysis (Algorithm~\ref{alg:core}) first identifies SAE latents whose activations exhibit robust monotonic trends along two axes: the proportion of unknown facts in the fine-tuning mixture and the number of training epochs. Activation Steering (Algorithm~\ref{alg:idf}) then tests whether steering these candidates can recover the fine-tuned model's accuracy on previously known facts. Their combination ensures the pipeline isolates latents that are not just correlated with fine-tuning dynamics but mechanistically implicated in the model's capacity to use pre-trained knowledge.

\begin{figure}[]
    \centering    
    \includegraphics[width=0.88\textwidth, trim={0 0 0 1.1cm}, clip]{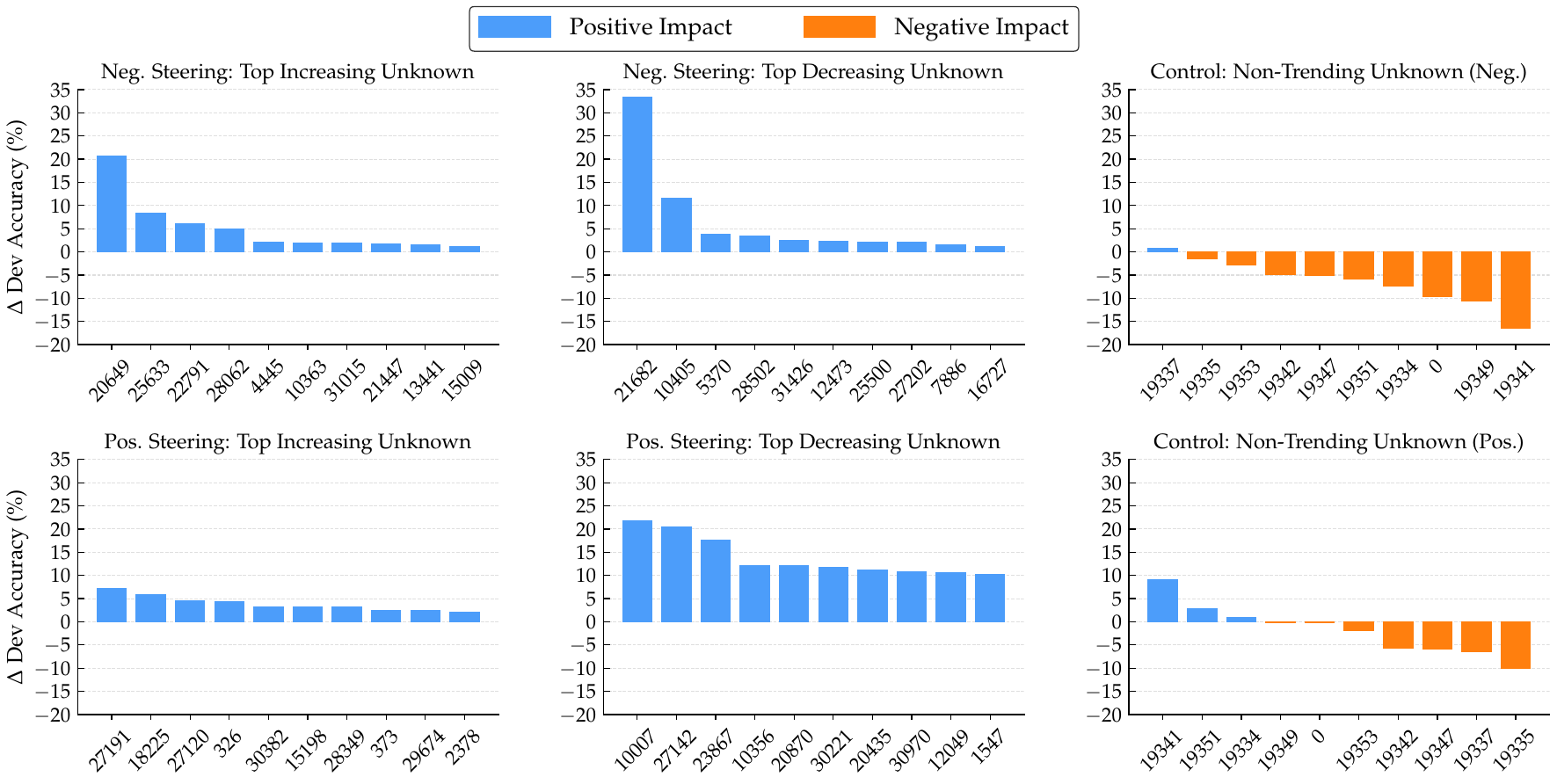}
    \caption{Dev accuracy change from single-latent steering on Llama 3.1 8B along the unknown dimension. Top row: negative steering; bottom row: positive. Left and centre columns show the top-10 increasing and decreasing latents from algorithm~\ref{alg:core}; the right column shows non-trending control latents from algorithm~\ref{alg:ctrl}, which produce negligible or negative changes in contrast to the consistent gains from trending latents.}
    \label{fig:unk_llama}
\end{figure}

\paragraph{Ruling out artifacts with control latents}
Figure~\ref{fig:unk_llama} shows that the top candidates discovered by our algorithm have a relatively significant effect on the performance when used for steering.  %
To confirm that this is 
not 
an artifact due to arbitrarily selected latents, we steer using a control group of latents that appear unaffected by fine-tuning 
as described in Algorithm~\ref{alg:ctrl} (Appendix~\ref{appendix:additional}): 
their activations present no trend or change across the fine-tuned models.
The last column of Figure~\ref{fig:unk_llama} shows the control group performance for Llama 3.1 8B (see Appendix~\ref{appendix:control} for the remaining models).
We find that the control group behaves very differently from the latents selected by our algorithm: %
steering with them mostly has a highly negative effect, whereas the positive effects are significantly less pronounced than the top-ranked latents, indicating the effectiveness of MoRFI. 

Table~\ref{tab:llama31_8b_unk} presents the top-10 latents 
for Llama 3.1 8B on the $\uparrow$Unknown direction (see Appendix~\ref{appendix:results} for the remaining models and the $\uparrow$Epochs direction).
Note that we focus 
on the $\uparrow$Unknown direction, as it isolates the effect of data composition from confounds introduced by extended training, such as distributional shift or overfitting, that may alter representations independently of exposure to unknown facts. The substantial overlap in surfaced latents between the two directions (e.g., latents 27191, 30382, 2378, 31426 in Tables~\ref{tab:llama31_8b_unk} and~\ref{tab:llama31_8b_epo}) suggests that they capture a shared signal, but the $\uparrow$Unknown direction provides a more targeted attribution of representational change to the specific variable of interest.

\paragraph{Asymmetries in latent steering dynamics} We detail several such asymmetries.
\noindent
\begin{wrapfigure}[4]{l}{0.7\textwidth}
\begin{minipage}[t]{0.7\textwidth}
\vspace{-10pt}
\begin{findingbox}[]
Negatively steered latents mostly yield larger gains than positively steered ones.
\end{findingbox}
\end{minipage}
\end{wrapfigure}
The negatively steered columns in Table~\ref{tab:llama31_8b_unk} %
show higher top  accuracy improvements than the positively steered columns. For the $\uparrow$\texttt{Unknown} direction, the best positively versus negatively steered latents give +21.7 vs +33.4 (Llama), +24.91 vs +39.77 (Mistral, Table~\ref{tab:mistral_7b_combined}), and +9.03 vs +18.46 (Gemma, Table~\ref{tab:gemma_2_9b_combined}). This suggests that features that grow with \% \texttt{Unknown} may become over-expressed and suppressing them recovers performance more effectively than promoting other features.

\begin{table}[]
\centering
\resizebox{\textwidth}{!}{%
\begin{tabular}{lcr|lcr|lcr|lcr}
\multicolumn{12}{c}{\textbf{Model: Llama 3.1 8B}} \\
\midrule

\multicolumn{12}{c}{\textbf{Direction: $\uparrow$Unknown}} \\
\midrule
\multicolumn{6}{c|}{\textbf{Increasing Latents}} & \multicolumn{6}{c}{\textbf{Decreasing Latents}} \\
\multicolumn{3}{c}{\textbf{Top-10 Positively Steered}} & \multicolumn{3}{c|}{\textbf{Top-10 Negatively Steered}} & \multicolumn{3}{c}{\textbf{Top-10 Positively Steered}} & \multicolumn{3}{c}{\textbf{Top-10 Negatively Steered}} \\
\midrule
\# & Acc. & $\alpha$ & \# & Acc. & $\alpha$ & \# & Acc. & $\alpha$ & \# & Acc. & $\alpha$ \\
\midrule
\href{https://www.neuronpedia.org/llama3.1-8b/15-llamascope-res-32k/27191}{\textbf{27191}} & \colorbox{lightgreen}{\relsize{-0.7}$\uparrow$7.12} 19.1 & 0.50 & 
\href{https://www.neuronpedia.org/llama3.1-8b/15-llamascope-res-32k/20649}{\textbf{20649}} & \colorbox{lightgreen}{\relsize{-0.7}$\uparrow$20.6} 21.5 & 0.35 & 
\href{https://www.neuronpedia.org/llama3.1-8b/15-llamascope-res-32k/10007}{\textbf{10007}} & \colorbox{lightgreen}{\relsize{-0.7}$\uparrow$21.7} 21.7 & 0.50 & 
\href{https://www.neuronpedia.org/llama3.1-8b/15-llamascope-res-32k/21682}{\textbf{21682}} & \colorbox{lightgreen}{\relsize{-0.7}$\uparrow$33.4} 23.8 & 0.70 \\
\href{https://www.neuronpedia.org/llama3.1-8b/15-llamascope-res-32k/18225}{\textbf{18225}} & \colorbox{lightgreen}{\relsize{-0.7}$\uparrow$5.89} 18.9 & 0.45 & 
\href{https://www.neuronpedia.org/llama3.1-8b/15-llamascope-res-32k/25633}{\textbf{25633}} & \colorbox{lightgreen}{\relsize{-0.7}$\uparrow$8.41} 19.3 & 0.65 & 
\href{https://www.neuronpedia.org/llama3.1-8b/15-llamascope-res-32k/27142}{\textbf{27142}} & \colorbox{lightgreen}{\relsize{-0.7}$\uparrow$20.5} 21.5 & 0.45 & 
\href{https://www.neuronpedia.org/llama3.1-8b/15-llamascope-res-32k/10405}{\textbf{10405}} & \colorbox{lightgreen}{\relsize{-0.7}$\uparrow$11.6} 19.9 & 0.15 \\
\href{https://www.neuronpedia.org/llama3.1-8b/15-llamascope-res-32k/27120}{\textbf{27120}} & \colorbox{lightgreen}{\relsize{-0.7}$\uparrow$4.50} 18.7 & 0.40 & 
\href{https://www.neuronpedia.org/llama3.1-8b/15-llamascope-res-32k/22791}{\textbf{22791}} & \colorbox{lightgreen}{\relsize{-0.7}$\uparrow$6.11} 18.9 & 0.25 & 
\href{https://www.neuronpedia.org/llama3.1-8b/15-llamascope-res-32k/23867}{\textbf{23867}} & \colorbox{lightgreen}{\relsize{-0.7}$\uparrow$17.7} 21.0 & 0.50 & 
\href{https://www.neuronpedia.org/llama3.1-8b/15-llamascope-res-32k/5370}{\textbf{5370}}   & \colorbox{lightgreen}{\relsize{-0.7}$\uparrow$3.75} 18.5 & 0.20 \\
\href{https://www.neuronpedia.org/llama3.1-8b/15-llamascope-res-32k/326}{\textbf{326}}     & \colorbox{lightgreen}{\relsize{-0.7}$\uparrow$4.34} 18.6 & 0.40 & 
\href{https://www.neuronpedia.org/llama3.1-8b/15-llamascope-res-32k/28062}{\textbf{28062}} & \colorbox{lightgreen}{\relsize{-0.7}$\uparrow$5.03} 18.7 & 0.30 & 
\href{https://www.neuronpedia.org/llama3.1-8b/15-llamascope-res-32k/10356}{\textbf{10356}} & \colorbox{lightgreen}{\relsize{-0.7}$\uparrow$12.2} 20.0 & 0.35 & 
\href{https://www.neuronpedia.org/llama3.1-8b/15-llamascope-res-32k/28502}{\textbf{28502}} & \colorbox{lightgreen}{\relsize{-0.7}$\uparrow$3.43} 18.5 & 0.35 \\
\href{https://www.neuronpedia.org/llama3.1-8b/15-llamascope-res-32k/30382}{\textbf{30382}} & \colorbox{lightgreen}{\relsize{-0.7}$\uparrow$3.21} 18.4 & 0.35 & 
\href{https://www.neuronpedia.org/llama3.1-8b/15-llamascope-res-32k/4445}{\textbf{4445}}   & \colorbox{lightgreen}{\relsize{-0.7}$\uparrow$2.04} 18.2 & 0.20 & 
\href{https://www.neuronpedia.org/llama3.1-8b/15-llamascope-res-32k/20870}{\textbf{20870}} & \colorbox{lightgreen}{\relsize{-0.7}$\uparrow$12.2} 20.0 & 0.45 & 
\href{https://www.neuronpedia.org/llama3.1-8b/15-llamascope-res-32k/31426}{\textbf{31426}} & \colorbox{lightgreen}{\relsize{-0.7}$\uparrow$2.41} 18.3 & 0.25 \\
\href{https://www.neuronpedia.org/llama3.1-8b/15-llamascope-res-32k/15198}{\textbf{15198}} & \colorbox{lightgreen}{\relsize{-0.7}$\uparrow$3.21} 18.4 & 0.30 & 
\href{https://www.neuronpedia.org/llama3.1-8b/15-llamascope-res-32k/10363}{\textbf{10363}} & \colorbox{lightgreen}{\relsize{-0.7}$\uparrow$1.87} 18.2 & 0.35 & 
\href{https://www.neuronpedia.org/llama3.1-8b/15-llamascope-res-32k/30221}{\textbf{30221}} & \colorbox{lightgreen}{\relsize{-0.7}$\uparrow$11.7} 19.9 & 0.45 & 
\href{https://www.neuronpedia.org/llama3.1-8b/15-llamascope-res-32k/12473}{\textbf{12473}} & \colorbox{lightgreen}{\relsize{-0.7}$\uparrow$2.36} 18.3 & 0.25 \\
\href{https://www.neuronpedia.org/llama3.1-8b/15-llamascope-res-32k/28349}{\textbf{28349}} & \colorbox{lightgreen}{\relsize{-0.7}$\uparrow$3.16} 18.4 & 0.50 & 
\href{https://www.neuronpedia.org/llama3.1-8b/15-llamascope-res-32k/31015}{\textbf{31015}} & \colorbox{lightgreen}{\relsize{-0.7}$\uparrow$1.87} 18.2 & 0.45 & 
\href{https://www.neuronpedia.org/llama3.1-8b/15-llamascope-res-32k/20435}{\textbf{20435}} & \colorbox{lightgreen}{\relsize{-0.7}$\uparrow$11.1} 19.8 & 0.40 & 
\href{https://www.neuronpedia.org/llama3.1-8b/15-llamascope-res-32k/25500}{\textbf{25500}} & \colorbox{lightgreen}{\relsize{-0.7}$\uparrow$2.20} 18.2 & 0.35 \\
\href{https://www.neuronpedia.org/llama3.1-8b/15-llamascope-res-32k/373}{\textbf{373}}     & \colorbox{lightgreen}{\relsize{-0.7}$\uparrow$2.46} 18.3 & 0.40 & 
\href{https://www.neuronpedia.org/llama3.1-8b/15-llamascope-res-32k/21447}{\textbf{21447}} & \colorbox{lightgreen}{\relsize{-0.7}$\uparrow$1.66} 18.1 & 0.35 & 
\href{https://www.neuronpedia.org/llama3.1-8b/15-llamascope-res-32k/30970}{\textbf{30970}} & \colorbox{lightgreen}{\relsize{-0.7}$\uparrow$10.8} 19.8 & 0.40 & 
\href{https://www.neuronpedia.org/llama3.1-8b/15-llamascope-res-32k/27202}{\textbf{27202}} & \colorbox{lightgreen}{\relsize{-0.7}$\uparrow$2.04} 18.2 & 0.35 \\
\href{https://www.neuronpedia.org/llama3.1-8b/15-llamascope-res-32k/29674}{\textbf{29674}} & \colorbox{lightgreen}{\relsize{-0.7}$\uparrow$2.41} 18.3 & 0.20 & 
\href{https://www.neuronpedia.org/llama3.1-8b/15-llamascope-res-32k/13441}{\textbf{13441}} & \colorbox{lightgreen}{\relsize{-0.7}$\uparrow$1.55} 18.1 & 0.30 & 
\href{https://www.neuronpedia.org/llama3.1-8b/15-llamascope-res-32k/12049}{\textbf{12049}} & \colorbox{lightgreen}{\relsize{-0.7}$\uparrow$10.6} 19.7 & 0.55 & 
\href{https://www.neuronpedia.org/llama3.1-8b/15-llamascope-res-32k/7886}{\textbf{7886}}   & \colorbox{lightgreen}{\relsize{-0.7}$\uparrow$1.50} 18.1 & 0.15 \\
\href{https://www.neuronpedia.org/llama3.1-8b/15-llamascope-res-32k/2378}{\textbf{2378}}   & \colorbox{lightgreen}{\relsize{-0.7}$\uparrow$2.09} 18.2 & 0.30 & 
\href{https://www.neuronpedia.org/llama3.1-8b/15-llamascope-res-32k/15009}{\textbf{15009}} & \colorbox{lightgreen}{\relsize{-0.7}$\uparrow$1.07} 18.0 & 0.10 & 
\href{https://www.neuronpedia.org/llama3.1-8b/15-llamascope-res-32k/1547}{\textbf{1547}}   & \colorbox{lightgreen}{\relsize{-0.7}$\uparrow$10.2} 19.7 & 0.50 & 
\href{https://www.neuronpedia.org/llama3.1-8b/15-llamascope-res-32k/16727}{\textbf{16727}} & \colorbox{lightgreen}{\relsize{-0.7}$\uparrow$1.23} 18.1 & 0.25 \\
\midrule
\end{tabular}
}
\caption{Top-10 impactful latents for Llama 3.1 8B selected from the unknown direction (increasing proportion of unknown facts in the fine-tuning mixture), 
identified by algorithm~\ref{alg:core} and validated by algorithm~\ref{alg:idf}. 
Columns are grouped by feature trend (increasing or decreasing, as identified 
by algorithm~\ref{alg:core}) and by steering polarity (positive or negative, 
as applied in algorithm~\ref{alg:idf}). \# denotes the corresponding sae latent index, 
acc.\ reports the steered model's accuracy on the dev set with the relative 
gain over the $\boldsymbol{M_{D_{100}}}$ baseline shown in green, 
and $\alpha$ is the optimal steering coefficient selected via grid search.}
\label{tab:llama31_8b_unk}
\end{table}

\noindent Features that decrease for models fine-tuned with more \texttt{Unknown} samples, when steered in either direction, tend to produce accuracy gains that match or exceed those from increasing latents. In Llama, the top decreasing latent (+21.7) when positively steered, triples the best increasing (+7.12). In Gemma the top decreasing, positively steered latent (+19.02) outperforms every increasing positively steered one. In Mistral, the top decreasing latent reaches +39.77 relative gains. This implies that suppressed features still encode pre-trained knowledge that can be recovered through steering.

\paragraph{Latent structure of Known vs Unknown knowledge}
In Table~\ref{tab:cross_model_knowledge_recovery}, we decompose the performance gains achieved through steering by attributing them to specific Wikidata relations, illustrating the functional role of the top latents. To quantify the restorative nature of these interventions, we define the Knowledge Recovery rate ($R_K$, Appendix~\ref{appendix:recovery}). Treating $\boldsymbol{M_{D_0}}$'s performance as an 
upper bound for recoverable facts, $R_K$ measures what fraction of the steering-induced gains correspond to $\boldsymbol{M_{D_0}}$'s knowledge.

\begin{wrapfigure}[4]{l}{0.7\textwidth}
\begin{minipage}[t]{0.7\textwidth}
\vspace{-10pt}
\begin{findingbox}[]
72-87\% of the facts recovered by steering top-latents on $\boldsymbol{M_{D_{100s}}}$ are \texttt{Known} to $\boldsymbol{M_{D_0}}$
\end{findingbox}
\end{minipage}
\end{wrapfigure}
We find that most of steering gains correspond to facts known to the better performing $\boldsymbol{M_{D_0}}$. The top benefitting relations: \href{https://www.wikidata.org/wiki/Property:P17}{\texttt{P17}} (country), \href{https://www.wikidata.org/wiki/Property:P36}{\texttt{P36}} (capital) and \href{https://www.wikidata.org/wiki/Property:P495}{\texttt{P495}} (country of origin) are semantically related geographic predicates, and their consistent co-recovery under single-latent steering across all three models is consistent with the expectation that semantically relevant relations share representational structure in the residual stream. Two of the corresponding features for Llama 3.1 8B are closely related to these relations according to Neuronpedia.\footnote{\url{https://www.neuronpedia.org/}} Feature \href{https://www.neuronpedia.org/llama3.1-8b/15-llamascope-res-32k/10007}{10007} refers to geographic locations and regions, while \href{https://www.neuronpedia.org/llama3.1-8b/15-llamascope-res-32k/20649}{20649} is aligned with features activating on locations, cities and countries (per-token examples in Appendix~\ref{appendix:qualitative}, and the recovery across training in Appendix~\ref{appendix:trajectory}).

\begin{table}[t]
\centering

\setlength{\tabcolsep}{4pt}

\resizebox{\textwidth}{!}{%
\begin{tabular}{@{}ll l ccc p{5.5cm}@{}}
\toprule
\textbf{Direction} & \textbf{Feature Trend} & \textbf{Intervention} & \textbf{Top Latent} & \textbf{Steered Acc.} & \textbf{Knowledge Recovered ($R_{K}$)} & \textbf{Top Benefiting Relations ($R_K(c)$)} \\ \midrule

\multirow{15}{*}{\textbf{$\uparrow$Unknown}} & 

\multicolumn{6}{c}{\textbf{Llama 3.1 8B} \quad (Baselines: $\boldsymbol{M_{D_{100}}} = 0.178$, $\boldsymbol{M_{D_0}} = 0.333$)} \\ \cmidrule(l){2-7} 

& \multirow{2}{*}{\textbf{Increasing}} 
& Positive (+) & \href{https://www.neuronpedia.org/llama3.1-8b/15-llamascope-res-32k/27191}{27191} & 0.191 & 74.3\% & \href{https://www.wikidata.org/wiki/Property:P495}{\texttt{P495}}(16.3\%), \href{https://www.wikidata.org/wiki/Property:P17}{\texttt{P17}}(16.0\%), \href{https://www.wikidata.org/wiki/Property:P36}{\texttt{P36}}(14.8\%)\\
& & Negative (-)& \href{https://www.neuronpedia.org/llama3.1-8b/15-llamascope-res-32k/20649}{20649} & 0.215 & 87.3\% & \href{https://www.wikidata.org/wiki/Property:P495}{\texttt{P495}}(56.5\%), \href{https://www.wikidata.org/wiki/Property:P17}{\texttt{P17}}(10.8\%), \href{https://www.wikidata.org/wiki/Property:P36}{\texttt{P36}}(6.5\%)\\ \cmidrule(l){2-7} 

& \multirow{2}{*}{\textbf{Decreasing}} 
& Positive (+)& \href{https://www.neuronpedia.org/llama3.1-8b/15-llamascope-res-32k/10007}{10007} & 0.217 & 79.7\% & \href{https://www.wikidata.org/wiki/Property:P495}{\texttt{P495}}(33.7\%), \href{https://www.wikidata.org/wiki/Property:P17}{\texttt{P17}}(19.7\%), \href{https://www.wikidata.org/wiki/Property:P131}{\texttt{P131}}(6.8\%)\\
& & Negative (-)& \href{https://www.neuronpedia.org/llama3.1-8b/15-llamascope-res-32k/21682}{21682} & 0.238 & 77.7\% & \href{https://www.wikidata.org/wiki/Property:P495}{\texttt{P495}}(30.6\%), \href{https://www.wikidata.org/wiki/Property:P17}{\texttt{P17}}(12.6\%), \href{https://www.wikidata.org/wiki/Property:P36}{\texttt{P36}}(9.1\%)\\ \cmidrule(l){2-7} %

& \multicolumn{6}{c}{\textbf{Gemma 2 9B} \quad (Baselines: $\boldsymbol{M_{D_{100}}} = \text{0.191}$, $\boldsymbol{M_{D_0}} = \text{0.360}$)} \\ \cmidrule(l){2-7}

& \multirow{2}{*}{\textbf{Increasing}} 
& Positive (+)& 18360 & 0.208 & 76.1\% & \href{https://www.wikidata.org/wiki/Property:P17}{\texttt{P17}} (22.8\%), \href{https://www.wikidata.org/wiki/Property:P36}{\texttt{P36}} (21.6\%), \href{https://www.wikidata.org/wiki/Property:P740}{\texttt{P740}} (7.3\%)\\
& & Negative (-)& 5152 & 0.226 & 85.1\% & \href{https://www.wikidata.org/wiki/Property:P495}{\texttt{P495}}(37.8\%), \href{https://www.wikidata.org/wiki/Property:P17}{\texttt{P17}}(23.7\%), \href{https://www.wikidata.org/wiki/Property:P36}{\texttt{P36}}(13.5\%)\\ \cmidrule(l){2-7} 

& \multirow{2}{*}{\textbf{Decreasing}} 
& Positive (+)& 5105 & 0.227 & 80.7\% & \href{https://www.wikidata.org/wiki/Property:P17}{\texttt{P17}}(25.0\%), \href{https://www.wikidata.org/wiki/Property:P36}{\texttt{P36}}(18.6\%), \href{https://www.wikidata.org/wiki/Property:P495}{\texttt{P495}}(16.6\%)\\
& & Negative (-)& 8277 & 0.213 & 72.3\% & \href{https://www.wikidata.org/wiki/Property:P17}{\texttt{P17}}(19.9\%), \href{https://www.wikidata.org/wiki/Property:P36}{\texttt{P36}}(15.1\%), \href{https://www.wikidata.org/wiki/Property:P495}{\texttt{P495}}(12.4\%)\\ \cmidrule(l){2-7}

& \multicolumn{6}{c}{\textbf{Mistral v03 7B} \quad (Baselines: $\boldsymbol{M_{D_{100}}} = \text{0.157}$, $\boldsymbol{M_{D_0}} = \text{0.319}$)} \\ \cmidrule(l){2-7}

& \multirow{2}{*}{\textbf{Increasing}} 
& Positive (+)& 41844 & 0.196 & 83.0\% & \href{https://www.wikidata.org/wiki/Property:P495}{\texttt{P495}} (38.3\%), \href{https://www.wikidata.org/wiki/Property:P17}{\texttt{P17}} (14.0\%), \href{https://www.wikidata.org/wiki/Property:P36}{\texttt{P36}} (8.7\%)\\
& & Negative (-)& 61325 & 0.218 & 83.8\% & \href{https://www.wikidata.org/wiki/Property:P495}{\texttt{P495}} (38.4\%), \href{https://www.wikidata.org/wiki/Property:P17}{\texttt{P17}} (19.9\%), \href{https://www.wikidata.org/wiki/Property:P36}{\texttt{P36}} (6.5\%)\\ \cmidrule(l){2-7} 

& \multirow{2}{*}{\textbf{Decreasing}} 
& Positive (+)& 18848 & 0.206 & 82.2\% & \href{https://www.wikidata.org/wiki/Property:P495}{\texttt{P495}}(38.4\%), \href{https://www.wikidata.org/wiki/Property:P17}{\texttt{P17}}(20.5\%), \href{https://www.wikidata.org/wiki/Property:P136}{\texttt{P136}}(7.3\%)\\
& & Negative (-)& 7859 & 0.219 & 77.1\% & \href{https://www.wikidata.org/wiki/Property:P495}{\texttt{P495}} (38.6\%), \href{https://www.wikidata.org/wiki/Property:P17}{\texttt{P17}} (14.3\%), \href{https://www.wikidata.org/wiki/Property:P36}{\texttt{P36}} (7.0\%)\\\bottomrule

\end{tabular}%
}
\caption{Cross-Model Knowledge Recovery and Relational Attribution via Single-Latent Steering. \textit{Knowledge Recovered} $R_{K}$, represents the fraction of pre-trained accuracy ($\boldsymbol{M_{D_0}}$) lost during fine-tuning ($\boldsymbol{M_{D_{100}}}$) that was successfully restored by the intervention.}
\label{tab:cross_model_knowledge_recovery}
\end{table}

\begin{figure}[htbp]
    \begin{minipage}[c]{0.53\textwidth} 
        \centering
        \includegraphics[width=\textwidth]{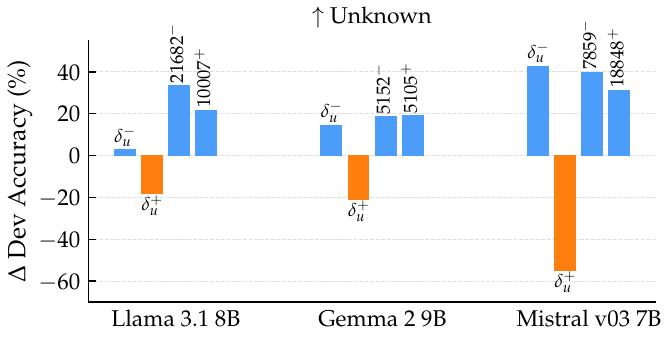}
    \end{minipage}
    \hfill 
    \begin{minipage}[c]{0.45\textwidth} 
        \caption{Comparison of steering with the composite direction $\boldsymbol{\delta_u}$ versus individual top-performing latents across three models. $\boldsymbol{\delta_u^-}$ and $\boldsymbol{\delta_u^+}$ denote subtracting and adding the $\boldsymbol{\delta_u}$ direction projected onto the residual stream via the SAE decoder. $\delta_u^-$ behaves consistently, %
        while individual latents sourced with MoRFI yield substantially larger improvements.}
        \label{fig:h_grid}
    \end{minipage}
\end{figure}

\noindent
\begin{findingbox}[]
$\boldsymbol{\delta_{u}}$ captures the direction ($\uparrow$Unknown) in SAE latent space along which model activations shift as train data transitions from fully \texttt{Known} to fully \texttt{Unknown} 
\end{findingbox}

To validate the direction identified by Algorithm~\ref{alg:core}, we steer directly with $\boldsymbol{\delta_{u}}$ (line~\ref{alg:line}), i.e., 
the composite vector capturing the aggregate activation shift from $\boldsymbol{M_{D_0}}$ to $\boldsymbol{M_{D_{100}}}$ in SAE space. This is projected onto the residual stream via the SAE decoder and added or subtracted at the middle layer. Across all three models, subtracting this direction produces accuracy gains in hallucinating models, while adding it deteriorates performance (Figure~\ref{fig:h_grid}). However, single-latent interventions sourced with MoRFI mostly outperform $\boldsymbol{\delta_{u}^-}$, indicating that the knowledge-relevant signal within the composite direction is concentrated in a sparse subset of its components, with the remainder diluting the effect or reversing it.
In Appendix~\ref{appendix:heatmaps} we plot the cosine similarity of the top-performing latents. 

Within-group similarity is generally low ($\approx$0.3), indicating that the top-performing latents span a distributed subspace at the middle layer rather than clustering along a shared low-rank structure.

While the intent of steering is to reverse the representational 
shift from $\boldsymbol{M_{D_0}}$ to 
$\boldsymbol{M_{D_{100}}}$, the optimal 
interventions substantially overshoot 
$\boldsymbol{M_{D_0}}$ activation levels 
at the middle layer:
negative 
steering suppresses increasing latents well below their 
$\boldsymbol{M_{D_0}}$ levels, while 
positive steering amplifies decreasing latents well above (Table~\ref{tab:overshoot}). 
The steered model achieves 
substantial gains from an activation configuration distinct from 
both $\boldsymbol{M_{D_0}}$ and 
$\boldsymbol{M_{D_{100}}}$, suggesting
tolerance to a range of mid-layer perturbations: downstream layers may correct for imprecise interventions,
funnelling different perturbed states towards similar output 
configurations from which parametric knowledge can be read out. The variation in optimal $\alpha$ across latents is consistent 
with this view, as different directions may align differently 
with the tolerant region, with better-aligned components 
admitting larger corrections before performance degrades.

\noindent
\begin{findingbox}[]
Steering gains persist from early stopping to convergence and transfer to the held-out test set.
\end{findingbox}
\noindent Steering the same Table~\ref{tab:cross_model_knowledge_recovery} latents at per-checkpoint fitted strengths yields positive gains already at the early-stop checkpoints in all but one case, growing with continued training toward the converged gains, and the dev-fitted strengths transfer to the held-out test set (mean +21.2\% vs +23.0\% on dev at convergence; Appendix~\ref{appendix:trajectory}).

\paragraph{Comparison with a difference-in-mean baseline}
Ranking latents by their difference-in-mean activation between $\boldsymbol{M_{D_0}}$ and $\boldsymbol{M_{D_{100}}}$ overlaps MoRFI's selection by only 18--41\%. Steering the disagreements shows that high endpoint magnitude is a poor predictor of causal effect in both directions, with most of the baseline's top rejected latents failing to improve accuracy while five of the twelve Table~\ref{tab:cross_model_knowledge_recovery} latents are absent from the baseline's top-$k$ altogether (Appendix~\ref{appendix:baseline}).

\section{Related Work}
\label{sec:related}

\paragraph{Fine-tuning, forgetting, and hallucinations} Beyond the behavioral evidence that unfamiliar facts induce hallucination \citep{gekhman2024doesfinetuningllms, kang2024unfamiliarfinetuningexamples}, recent work examines why fine-tuning degrades factual recall. \citet{kaplan2026whyfinetuningencourages} study why SFT encourages hallucinations and how to mitigate them. More broadly, \citet{calderon2026emptyshelveslost} show that most factual errors in LLMs reflect recall rather than encoding failures, facts stored but unreachable. Closer to our perspective, \citet{nishi2025representationshatteringtransformers} show in a synthetic setting that targeted knowledge edits distort the geometry of entity representations well beyond the edited fact, while \citet{masip2026puttingfaceforgetting} characterize catastrophic forgetting mechanistically, as features losing capacity or their readout being disrupted. We chart the analogous representational change for supervised fine-tuning on natural QA data, locate it in specific residual-stream directions, and causally intervene on it.

\paragraph{Sparse autoencoders and their limits} SAEs decompose activations into sparse, monosemantic-leaning latents \citep{bricken2023monosemanticitydecomposinglanguage, cunningham2023sparseautoencodersfind}, enabling suites of pre-trained SAEs \citep{lieberum2024gemmascopeopen, he2024llamascopeextracting}. A growing critique cautions that latents need not correspond to model features. SAEs recover structure even from randomly initialized networks \citep{heap2025sparseautoencoderscan}, embed architectural assumptions \citep{hindupur2025projectingassumptionsduality}, miss multi-dimensional features \citep{engels2025notalllanguage}, and interventions on them can be unreliable \citep{cui2026saeinterventionsare}. MoRFI is agnostic on this debate, treating latents as candidate directions only, retained when their activations track the controlled property and their causal effect survives validation against non-trending controls. Complementing endpoint model diffing, \citet{bayazit2025crosscodingtimetracking} track feature emergence across pre-training checkpoints with crosscoders and \citet{ward2025reasoningfinetuningrepurposeslatent} show fine-tuning repurposes existing latents; MoRFI instead requires monotone, resample-stable trends along a controlled fine-tuning gradient.

\paragraph{Activation steering} Steering adds directions to hidden states to control behavior \citep{turner2024steeringlanguagemodels, zou2025representationengineeringtopdown}. \citet{arad2025saesaregood} show SAE steering succeeds precisely when latent selection succeeds, \citet{cho2025corrsteergenerationtimellm} automate selection by correlating activations with task correctness, \citet{braun2025understandingunreliabilitysteering} and \citet{taimeskhanov2026understandingsteeringstrength} analyze when and why it fails, and \citet{casademunt2025steeringoutofdistributiongeneralization} steer via concept ablation for robust generalization. MoRFI contributes the selection procedure, identifying which latents to steer and at what strength, and validating both causally.

\section{Conclusion}

We introduce a two-stage pipeline that identifies residual-stream directions causally linked to model behavior. Applied to fine-tuning-induced knowledge degradation across three architectures, MoRFI shows a large gap between composite and single-latent interventions, indicating sparse knowledge signals. 72--87\% of the facts recovered by steering align with those accessible to $\boldsymbol{M_{D_0}}$, the baseline fine-tuned exclusively on known facts, consistent with these latents mediating access to stored knowledge rather than performance itself. While steering establishes that the identified directions causally influence knowledge recovery, with gains persisting on held-out data, it does not establish these directions as the mechanism of forgetting, and the optimal interventions do not restore $\boldsymbol{M_{D_0}}$'s activation state (Appendix~\ref{appendix:overshoot}). Our results suggest that part of the observed forgetting reflects disrupted access rather than erasure, with implications for mitigating hallucinations via targeted inference-time interventions.

\clearpage

\section*{Ethics Statement}

Given the nature of our paper, and that we did not elicit any human evaluation studies we do not perceive any direct ethics considerations for this paper. We still acknowledge the risk of dual-use of the three models we used, especially since we are using their base variant which does not involve any post-training techniques for harnessing and refusing unwanted or harmful behavior if triggered maliciously. At the same time, the dataset we used (\texttt{EntityQuestions}; \citealt{sciavolino2022simpleentitycentricquestions}) contains a carefully curated set of questions and answers derived from Wikidata entities, hence the risk should be low. Finally, the paper comes to explain the behavior of LLMs, which are still perceived mostly as black boxes. In that sense, the paper strives to shed some light and move mechanistic interpretability towards a direction to increase the safety in using LLMs.

\section*{Acknowledgements}

We thank the anonymous reviewers and the area chair for their useful feedback and comments. We thank Adi Simhi for the helpful feedback
provided on an earlier draft of this paper. The authors acknowledge the use of resources provided by the Isambard-AI National AI Research Resource (AIRR). Isambard-AI is operated by the University of Bristol and is funded by the UK Government’s Department for Science, Innovation and Technology (DSIT) via UK Research and Innovation; and the Science and Technology Facilities Council [ST/AIRR/I-A-I/1023]~\citep{mcintoshsmith2024isambardaileadershipclasssupercomputer}.

\bibliography{references,colm2026_conference}
\bibliographystyle{colm2026_conference}

\appendix

\section{Fine-tuning and SAE Details}
\label{appendix:a}

For each group of models $M_{D_p}$ for $p \in \mathcal{P} =\{0, 10, 25, 50, 75, 90, 100\}$ we ran full fine-tuning using the same learning rate per group which was picked such that all 7 models converge within 50 epochs. Full hyper-parameter details are shown in Table~\ref{tab:model_comparison}. The pre-trained SAEs we used are shown in Table~\ref{tab:sae_config_detailed}.

\begin{table}[htbp]
\centering
\small
\begin{tabular}{@{}lccc@{}}
\toprule
\textbf{Hyperparameter} & \textbf{Mistral 7B v0.3} & \textbf{Llama 3.1 8B} & \textbf{Gemma 2 9B} \\ \midrule
\textit{Training Dynamics} & & & \\
Learning Rate & $5.5 \times 10^{-6}$ & $2.0 \times 10^{-5}$ & $4.5 \times 10^{-6}$ \\
LR Scheduler & Constant & Constant & Constant \\
Attention Dropout & 0.2 & 0.3 & 0 \\
Total Training Epochs & 10-50 & 10-50 & 10-50 \\
\midrule
\textit{Batch Configuration} & & & \\
Per Device Batch Size & 32 & 32 & 32 \\
Gradient Accumulation & 4 & 4 & 4 \\
Effective Batch Size & 128 & 128 & 128 \\
Max Seq Length & 1024 & 1024 & 1024 \\
Packing Strategy & BFD & BFD & BFD \\
\midrule
\textit{Architecture Details} & & & \\
Hidden Size ($d_{\textit{model}}$) & 4096 & 4096 & 3584 \\
Num Hidden Layers & 32 & 32 & 42 \\
Attention Heads (Q/KV) & 32 / 8 & 32 / 8 & 16 / 8 \\
Vocab Size & 32768 & 128256 & 256000 \\
Max Pos Embeddings & 32768 & 131072 & 8192 \\
\midrule
\textit{Hardware \& Precision} & & & \\
GPU & GH200 120GB & GH200 120GB & GH200 120GB \\
Mixed Precision & Bfloat16 & Bfloat16 & Bfloat16 \\
Optimizer & AdamW (Fused) & AdamW (Fused) & AdamW (Fused) \\
Optimization Kernels & Liger & Liger & Liger \\
\bottomrule
\end{tabular}
\caption{Hyperparameter comparison across model architectures.}
\label{tab:model_comparison}
\end{table}

\begin{table}[htbp]
\centering
\small
\begin{tabular}{@{}lllc@{}}
\toprule
\textbf{Model} & \textbf{SAE Release} & \textbf{SAE Identifier (ID)} & $\mathbf{d_{\textit{sae}}}$ \\ \midrule
Mistral 7B v0.3 & mistral-7b-res-wg & blocks.16.hook\_resid\_pre & 65536 \\
Llama 3.1 8B & llama\_scope\_lxr\_8x & l15r\_8x & 32768 \\
Gemma 2 9B & gemma-scope-9b-pt-res-canonical & layer\_20/width\_32k/canonical & 32768 \\ \bottomrule
\end{tabular}
\caption{SAE configuration and latent dimensionality.}
\label{tab:sae_config_detailed}
\end{table}
\clearpage
\section{Definition of \texorpdfstring{$P_{\texttt{Correct}}$}{P Correct}}
\label{appendix:pcorrect}
We follow \citet{gekhman2024doesfinetuningllms} and others \citep{kadavath2022languagemodelsmostly, petroni2019languagemodelsknowledge} in assessing the acquired knowledge of a model after pre-training using greedy decoding and sampling to categorize input questions as \texttt{Known} or \texttt{Unknown}. $P_{\texttt{Correct}}$ is an estimate of how likely is $\boldsymbol{M}$ to accurately generate the correct answer \textit{a} to \textit{q}, when prompted with random few-shot exemplars and using decoding temperature \textit{T}.

\begin{align}
P_{\texttt{Correct}}(q, a; M, T=0) & = \frac{\text{\#correct\_greedy\_responses}}{N_{\textit{ex}}},\\
P_{\texttt{Correct}}(q, a; M, T>0) & = \frac{\text{\#correct\_sampled\_responses}}{N_{\textit{ex}} \cdot N_{\textit{sampled}}}.
\end{align}
We use $N_{\textit{ex}}=10$ exemplars with $k=4$ demonstrations each. We take $N_{\textit{sampled}}=16$ samples with temperature $T=0.5$ from top-40.

\section{Data Statistics}
\begin{figure}[htbp] %
    \centering
    \begin{subfigure}{0.32\textwidth}
        \centering
        \includegraphics[width=\textwidth , trim={0 18.0 0 0.66cm}, clip]{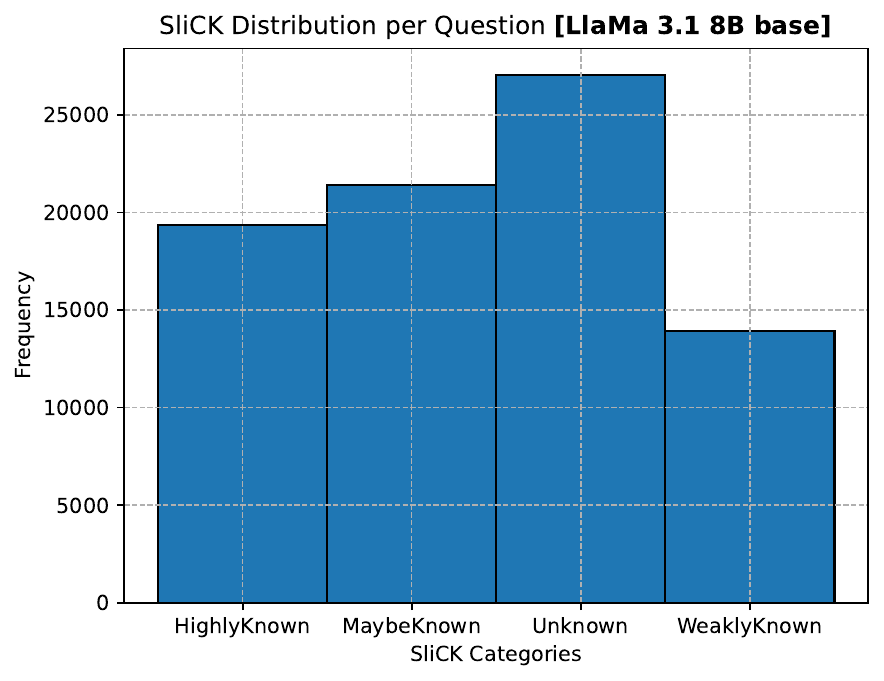}
        \caption{Llama 3.1 8B }        
    \end{subfigure}
    \hfill %
    \begin{subfigure}{0.32\textwidth}
        \centering
        \includegraphics[width=\textwidth, trim={0 0 0 0.65cm}, clip]{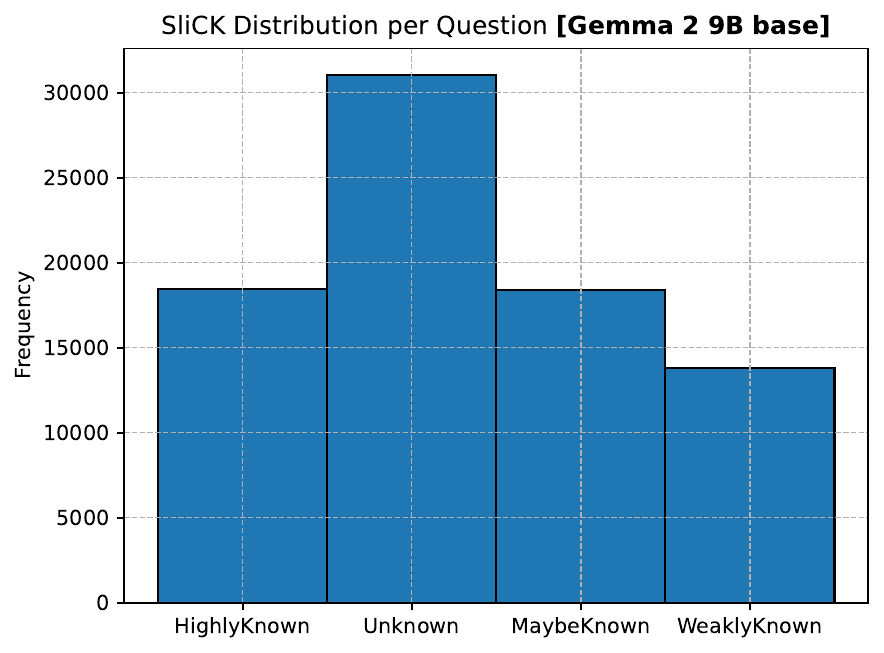}
        \caption{Gemma 2 9B}
    \end{subfigure}
    \hfill %
    \begin{subfigure}{0.32\textwidth}
        \centering
        \includegraphics[width=\textwidth, trim={0 18.0 0 0.66cm}, clip]{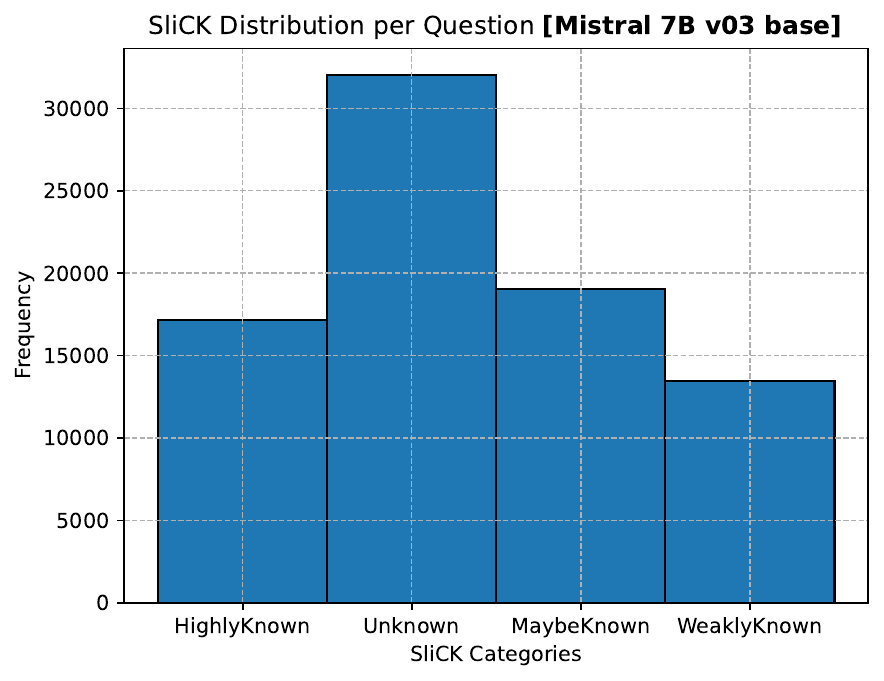}
        \caption{Mistral v03 7B}
    \end{subfigure}

    \caption{Knowledge distribution across models on full train set split.}
    \label{fig:knowledge_dist}
\end{figure}

 In this work we use the \texttt{EntityQuestions} dataset \citep{sciavolino2022simpleentitycentricquestions}, comprising 
 $N_{\textit{train}} = 81700$, $N_{\textit{dev}}=10461$, and $N_{\textit{test}}=10481$ examples. Figure~\ref{fig:knowledge_dist} shows the distribution of examples based on $P_{\texttt{Correct}}$ on the full train set across all models. Tables~\ref{tab:wikidata-llama}-\ref{tab:wikidata-mistral} show the distribution of examples per Wikidata relation as outlined in Table~\ref{tab:relations}.

\label{appendix:data}
\begin{table}[ht]
\centering
\begin{tabular}{@{}ll@{}}
\toprule
\textbf{relation} & \textbf{question template} \\
\midrule
P131 & Where is [E] located? \\
P136 & What type of music does [E] play? \\
P17 & Which country is [E] located in?  \\
P19 & Where was [E] born? \\
P26 & Who is [E] married to? \\
P264 & What music label is [E] represented by? \\
P36 & What is the capital of [E]? \\
P40 & Who is [E]'s child?  \\
P495 & Which country was [E] created in? \\
P69 & Where was [E] educated? \\
P740 & Where was [E] founded? \\
P800 & What is [E] famous for? \\
\bottomrule
\end{tabular}
\caption{Question templates used in \texttt{EntityQuestions} dataset \citep{sciavolino2022simpleentitycentricquestions}.}
\label{tab:relations}
\end{table}

\begin{table}[ht]
\centering\small
\begin{tabular}{@{}lrrrrrrr@{}}
\toprule
\textbf{relation} & \texttt{HighlyKnown} & \texttt{MaybeKnown} & \texttt{WeaklyKnown} & \texttt{Unknown} & \texttt{Total} & \texttt{Min} \\
\midrule
P131 & 465 & 2762 & 3166 & 1253 & 7646 & 465 \\
P136 & 178 & 3434 &	1949 & 1955 & 7516 & 178 \\
P17 & 4423 & 2547 & 444 & 476 &	7890 & 444\\
P19 & 700 &	1500 & 4100 & 1621 & 7921 & 700 \\
P26 & 1469 & 1439 &	3389 & 1159	& 7456 & 1159 \\
P264 & 217 & 1487 &	3772 & 1848 & 7324 & 217 \\
P36  & 4696 & 1162 & 514 & 443 & 6815 &	443\\
P40  & 540 & 1383 &	2840 & 1347 & 6110 & 540\\
P495 & 4680 & 1971 & 612 & 411 & 7674 &	411\\
P69  & 383 & 1106 &	3293 & 1939 & 6721 & 383\\
P740 & 1303 & 2239 & 2484 &	1245 &	7271 & 1245 \\
P800 & 281 & 364 & 494 & 217 & 1356 & 217 \\
\midrule
TOTAL & 19335 & 21394 & 13914 & 27057 & 81700 & \textbf{6402} \\
\bottomrule
\end{tabular}
\caption{Knowledge distribution across Wikidata relations for \textbf{Llama 3.1 8B}.}
\label{tab:wikidata-llama}
\end{table}

\begin{table}[ht]
\centering\small
\begin{tabular}{@{}lrrrrrr}
\toprule
\textbf{relation} & \texttt{HighlyKnown} & \texttt{MaybeKnown} & \texttt{WeaklyKnown} & \texttt{Unknown} & \texttt{Total} & \texttt{Min} \\
\midrule
P131 & 496 & 2279 & 1278 & 3593 & 7646 & 496 \\
P136 & 107 & 2697 & 2095 & 2617 & 7516 & 107 \\
P17  & 4147 & 2690 & 547 & 506 & 7890 & 506 \\
P19  & 614 & 1359 & 1430 & 4518 & 7921 & 614 \\
P26  & 1476 & 1103 & 1004 & 3873 & 7456 & 1004 \\
P264 & 209 & 1316 & 1666 & 4133 & 7324 & 209 \\
P36  & 4547 & 1112 & 511 & 645 & 6815 & 511 \\
P40  & 532 & 1148 & 1130 & 3300 & 6110 & 532 \\
P495 & 4818 & 1675 & 533 & 648 & 7674 & 533 \\
P69  & 252 & 865 & 1853 & 3751 & 6721 & 252 \\
P740 & 954 & 1871 & 1553 & 2893 & 7271 & 954 \\
P800 & 275 & 303 & 199 & 579 & 1356 & 199 \\
\midrule
TOTAL & 18427 & 18418 & 13799 & 31056 & 81700 & \textbf{5917} \\
\bottomrule
\end{tabular}
\caption{Knowledge distribution across Wikidata relations for \textbf{Gemma 2 9B}.}
\label{tab:wikidata-gemma}
\end{table}

\begin{table}[ht]
\centering\small
\begin{tabular}{@{}lrrrrrrr@{}}
\toprule
\textbf{relation} & \texttt{HighlyKnown} & \texttt{MaybeKnown} & \texttt{WeaklyKnown} & \texttt{Unknown} & \texttt{Total} & \texttt{Min} \\
\midrule
P131 & 463 & 2188 & 1312 & 3683 & 7646 & 463 \\
P136 & 134 & 2860 & 1861 & 2661 & 7516 & 134 \\
P17  & 3702 & 3102 & 553 & 533 & 7890 & 533 \\
P19  & 560 & 1367 & 1562 & 4432 & 7921 & 560 \\
P26  & 1070 & 1114 & 1017 & 4255 & 7456 & 1017 \\
P264 & 156 & 1391 & 1672 & 4105 & 7324 & 156 \\
P36  & 4062 & 1429 & 515 & 809 & 6815 & 515 \\
P40  & 378 & 1003 & 1175 & 3554 & 6110 & 378 \\
P495 & 4976 & 1506 & 444 & 748 & 7674 & 444 \\
P69  & 221 & 904 & 1838 & 3758 & 6721 & 221 \\
P740 & 1159 & 1910 & 1306 & 2896 & 7271 & 1159 \\
P800 & 274 & 279 & 197 & 606 & 1356 & 197 \\
\midrule
TOTAL & 17155 & 19053 & 13452 & 32040 & 81700 & \textbf{5777} \\
\bottomrule
\end{tabular}
\caption{Knowledge distribution across Wikidata relations for \textbf{Mistral v03 7B}.}
\label{tab:wikidata-mistral}
\end{table}
\clearpage

\section{Prompts}
\label{appendix:prompts}
We use a four-shot prompt to assess base models knowledge with 
$P_{\texttt{Correct}}$ (see definition in Appendix~\ref{appendix:pcorrect}). We fine-tune and evaluate all base models using the ``Fine-tuning prompt'' below. Finally, we collect activations from all positions except the BOS token for all models. The following example corresponds to Llama 3.1 8B with highlighted tokens being collected.

\begin{tcolorbox}[
    enhanced,
    colback=platonicbg,
    colframe=platonicframe,  %
    coltitle=black,          %
    boxrule=0.8pt,
    arc=3pt,
    fonttitle=\bfseries,
    title=Prompt for computing $P_{\texttt{Correct}}$ ,
    attach title to upper,
    after title={
        \par\nobreak\vspace{0.2em}
        {\color{platonicframe}\hrule} %
        \kern0.8em
    }
]
Q: Which country was Moxie created in?\\
        A: United States of America

        Q: Which country was Jook-sing noodles created in?\\
        A: People's Republic of China

        Q: Which country was You're My Pet created in?\\
        A: South Korea

        Q: Which country was Outta Control created in?\\
        A: United States of America

        Q: Which country was Embrace of the Vampire created in?\\
        A:
\end{tcolorbox}

\begin{tcolorbox}[
    enhanced,
    colback=platonicbg,
    colframe=platonicframe,  %
    coltitle=black,          %
    boxrule=0.8pt,
    arc=3pt,
    fonttitle=\bfseries,
    title=Fine-tuning Prompt ,
    attach title to upper,
    after title={
        \par\nobreak\vspace{0.2em}
        {\color{platonicframe}\hrule} %
        \kern0.8em
    }
]
Answer the following question.\\ What kind of work does Ron Konopka do?
\end{tcolorbox}

\begin{tcolorbox}[
    enhanced,
    colback=platonicbg,
    colframe=platonicframe,  %
    coltitle=black,          %
    boxrule=0.8pt,
    arc=3pt,
    fonttitle=\bfseries,
    title=Activation Extraction Tokens,
    attach title to upper,
    after title={
        \par\nobreak\vspace{0.4em}
        {\color{platonicframe}\hrule} %
        \kern0.8em
    }
]

<|begin\_of\_text|>{\color{red}<|start\_header\_id|>user<|end\_header\_id|>
\vspace{1em}\\
Answer the following question.\\Where is Warialda Rail located?\\<|eot\_id|>}<|start\_header\_id|>assistant<|end\_header\_id|>
\vspace{1em}\\
New South Wales<|eot\_id|>
\end{tcolorbox}

\section{Definition of $R_{K}$}
\label{appendix:recovery}
Empirically, $R_K$ is calculated as the conditional probability that a fact was originally known by the initial baseline ($M_{D_0}$), given that it was successfully learned by the steered model ($M_{D_{100s}}$) after being forgotten by the fine-tuned baseline ($M_{D_{100}}$):
\begin{equation}
R_K = P(M_{D_0}=1 \mid M_{D_{100s}}=1, M_{D_{100}}=0).
\end{equation}

To dissect these gains functionally, we attribute this recovery to specific relations. The global contribution of a single relation $c$ to the total recovery rate is defined as the joint probability of the recovery occurring specifically within relation $c$, conditional on a gross gain occurring anywhere in the dataset:
\begin{equation}
R_K(c) = P(M_{D_0}=1, \text{relation}=c \mid M_{D_{100s}}=1, M_{D_{100}}=0).  
\end{equation}

To ensure statistical relevance, we report relational contributions exclusively for categories containing at least 50 initial samples ($|c| \ge 50$) and exhibiting a minimum of 10 gross steering gains.
\clearpage
\section{Additional Single-latent Steering Results}
\label{appendix:results}

In Tables~\ref{tab:gemma_2_9b_combined},~\ref{tab:mistral_7b_combined},~\ref{tab:llama31_8b_epo} and ~\ref{tab:llama31_8b_unk}. We present single-latent steering results for latents sourced by MoRFI, across three model architectures, along the $\uparrow$Unknown and $\uparrow$Epochs directions.

\begin{table}[ht]
\centering
\resizebox{\textwidth}{!}{%
\begin{tabular}{lcr|lcr|lcr|lcr}
\multicolumn{12}{c}{\textbf{Model: Gemma 2 9B}} \\
\midrule

\multicolumn{12}{c}{\textbf{Direction: $\uparrow$Unknown}} \\
\midrule
\multicolumn{6}{c|}{\textbf{Increasing Latents}} & \multicolumn{6}{c}{\textbf{Decreasing Latents}} \\
\multicolumn{3}{c}{\textbf{Top-10 Positively Steered}} & \multicolumn{3}{c|}{\textbf{Top-10 Negatively Steered}} & \multicolumn{3}{c}{\textbf{Top-10 Positively Steered}} & \multicolumn{3}{c}{\textbf{Top-10 Negatively Steered}} \\
\midrule
\# & Acc. & $\alpha$ & \# & Acc. & $\alpha$ & \# & Acc. & $\alpha$ & \# & Acc. & $\alpha$ \\
\midrule
\textbf{18360} & \colorbox{lightgreen}{\relsize{-0.7}$\uparrow$9.03} 20.8 & 0.75 & 
\textbf{5152} & \colorbox{lightgreen}{\relsize{-0.7}$\uparrow$18.46} 22.6 & 0.60 & 
\textbf{5105} & \colorbox{lightgreen}{\relsize{-0.7}$\uparrow$19.02} 22.7 & 0.50 & 
\textbf{8277} & \colorbox{lightgreen}{\relsize{-0.7}$\uparrow$11.54} 21.3 & 0.45 \\
\textbf{4781} & \colorbox{lightgreen}{\relsize{-0.7}$\uparrow$8.83} 20.7 & 0.55 & 
\textbf{29477} & \colorbox{lightgreen}{\relsize{-0.7}$\uparrow$12.95} 21.5 & 0.70 & 
\textbf{22788} & \colorbox{lightgreen}{\relsize{-0.7}$\uparrow$11.99} 21.3 & 0.55 & 
\textbf{26862} & \colorbox{lightgreen}{\relsize{-0.7}$\uparrow$9.28} 20.8 & 0.70 \\
\textbf{17451} & \colorbox{lightgreen}{\relsize{-0.7}$\uparrow$8.58} 20.7 & 0.55 & 
\textbf{19753} & \colorbox{lightgreen}{\relsize{-0.7}$\uparrow$12.74} 21.5 & 0.75 & 
\textbf{17606} & \colorbox{lightgreen}{\relsize{-0.7}$\uparrow$11.94} 21.3 & 0.45 & 
\textbf{25130} & \colorbox{lightgreen}{\relsize{-0.7}$\uparrow$8.73} 20.7 & 0.50 \\
\textbf{11676} & \colorbox{lightgreen}{\relsize{-0.7}$\uparrow$6.82} 20.4 & 0.45 & 
\textbf{19424} & \colorbox{lightgreen}{\relsize{-0.7}$\uparrow$9.68} 20.9 & 0.35 & 
\textbf{29990} & \colorbox{lightgreen}{\relsize{-0.7}$\uparrow$10.64} 21.1 & 0.65 & 
\textbf{21907} & \colorbox{lightgreen}{\relsize{-0.7}$\uparrow$6.57} 20.3 & 0.60 \\
\textbf{5378} & \colorbox{lightgreen}{\relsize{-0.7}$\uparrow$6.82} 20.4 & 0.75 & 
\textbf{2050} & \colorbox{lightgreen}{\relsize{-0.7}$\uparrow$9.23} 20.8 & 0.35 & 
\textbf{32248} & \colorbox{lightgreen}{\relsize{-0.7}$\uparrow$10.24} 21.0 & 0.60 & 
\textbf{16333} & \colorbox{lightgreen}{\relsize{-0.7}$\uparrow$6.42} 20.3 & 0.60 \\
\textbf{18071} & \colorbox{lightgreen}{\relsize{-0.7}$\uparrow$6.62} 20.3 & 0.50 & 
\textbf{16328} & \colorbox{lightgreen}{\relsize{-0.7}$\uparrow$8.73} 20.7 & 0.40 & 
\textbf{13604} & \colorbox{lightgreen}{\relsize{-0.7}$\uparrow$10.24} 21.0 & 0.70 & 
\textbf{21602} & \colorbox{lightgreen}{\relsize{-0.7}$\uparrow$5.92} 20.2 & 0.55 \\
\textbf{20711} & \colorbox{lightgreen}{\relsize{-0.7}$\uparrow$6.22} 20.2 & 0.50 & 
\textbf{2546} & \colorbox{lightgreen}{\relsize{-0.7}$\uparrow$8.73} 20.7 & 0.75 & 
\textbf{14523} & \colorbox{lightgreen}{\relsize{-0.7}$\uparrow$9.68} 20.9 & 0.55 & 
\textbf{16312} & \colorbox{lightgreen}{\relsize{-0.7}$\uparrow$5.42} 20.1 & 0.40 \\
\textbf{6267} & \colorbox{lightgreen}{\relsize{-0.7}$\uparrow$5.67} 20.1 & 0.65 & 
\textbf{6012} & \colorbox{lightgreen}{\relsize{-0.7}$\uparrow$8.38} 20.6 & 0.60 & 
\textbf{793} & \colorbox{lightgreen}{\relsize{-0.7}$\uparrow$9.58} 20.9 & 0.75 & 
\textbf{7406} & \colorbox{lightgreen}{\relsize{-0.7}$\uparrow$5.12} 20.0 & 0.35 \\
\textbf{15778} & \colorbox{lightgreen}{\relsize{-0.7}$\uparrow$5.27} 20.1 & 0.45 & 
\textbf{13092} & \colorbox{lightgreen}{\relsize{-0.7}$\uparrow$8.03} 20.6 & 0.40 & 
\textbf{27604} & \colorbox{lightgreen}{\relsize{-0.7}$\uparrow$8.33} 20.6 & 0.35 & 
\textbf{18342} & \colorbox{lightgreen}{\relsize{-0.7}$\uparrow$5.12} 20.0 & 0.55 \\
\textbf{7393} & \colorbox{lightgreen}{\relsize{-0.7}$\uparrow$5.12} 20.0 & 0.55 & 
\textbf{15457} & \colorbox{lightgreen}{\relsize{-0.7}$\uparrow$8.03} 20.6 & 0.45 & 
\textbf{13641} & \colorbox{lightgreen}{\relsize{-0.7}$\uparrow$8.13} 20.6 & 0.65 & 
\textbf{12469} & \colorbox{lightgreen}{\relsize{-0.7}$\uparrow$4.52} 19.9 & 0.65 \\
\midrule

\multicolumn{12}{c}{\textbf{Direction: $\uparrow$Epochs}} \\
\midrule
\multicolumn{6}{c|}{\textbf{Increasing Latents}} & \multicolumn{6}{c}{\textbf{Decreasing Latents}} \\
\multicolumn{3}{c}{\textbf{Top-10 Positively Steered}} & \multicolumn{3}{c|}{\textbf{Top-10 Negatively Steered}} & \multicolumn{3}{c}{\textbf{Top-10 Positively Steered}} & \multicolumn{3}{c}{\textbf{Top-10 Negatively Steered}} \\
\midrule
\# & Acc. & $\alpha$ & \# & Acc. & $\alpha$ & \# & Acc. & $\alpha$ & \# & Acc. & $\alpha$ \\
\midrule
\textbf{5105}  & \colorbox{lightgreen}{\relsize{-0.7}$\uparrow$19.02} 22.7 & 0.50 & 
\textbf{13038} & \colorbox{lightgreen}{\relsize{-0.7}$\uparrow$17.46} 22.4 & 0.75 & 
\textbf{15720} & \colorbox{lightgreen}{\relsize{-0.7}$\uparrow$6.32} 20.3 & 0.55 & 
\textbf{18632} & \colorbox{lightgreen}{\relsize{-0.7}$\uparrow$16.96} 22.3 & 0.75 \\
\textbf{20769} & \colorbox{lightgreen}{\relsize{-0.7}$\uparrow$13.35} 21.6 & 0.60 & 
\textbf{29477} & \colorbox{lightgreen}{\relsize{-0.7}$\uparrow$12.95} 21.5 & 0.70 & 
\textbf{8722}  & \colorbox{lightgreen}{\relsize{-0.7}$\uparrow$5.27} 20.1 & 0.65 & 
\textbf{17517} & \colorbox{lightgreen}{\relsize{-0.7}$\uparrow$7.33} 20.4 & 0.65 \\
\textbf{22788} & \colorbox{lightgreen}{\relsize{-0.7}$\uparrow$11.99} 21.3 & 0.55 & 
\textbf{18284} & \colorbox{lightgreen}{\relsize{-0.7}$\uparrow$10.99} 21.1 & 0.65 & 
\textbf{921}   & \colorbox{lightgreen}{\relsize{-0.7}$\uparrow$4.47} 19.9 & 0.65 & 
\textbf{17337} & \colorbox{lightgreen}{\relsize{-0.7}$\uparrow$6.62} 20.3 & 0.75 \\
\textbf{19121} & \colorbox{lightgreen}{\relsize{-0.7}$\uparrow$10.39} 21.0 & 0.75 & 
\textbf{17249} & \colorbox{lightgreen}{\relsize{-0.7}$\uparrow$10.44} 21.0 & 0.55 & 
\textbf{28438} & \colorbox{lightgreen}{\relsize{-0.7}$\uparrow$4.01} 19.8 & 0.70 & 
\textbf{16333} & \colorbox{lightgreen}{\relsize{-0.7}$\uparrow$6.42} 20.3 & 0.60 \\
\textbf{10585} & \colorbox{lightgreen}{\relsize{-0.7}$\uparrow$9.33} 20.8 & 0.65 & 
\textbf{26670} & \colorbox{lightgreen}{\relsize{-0.7}$\uparrow$10.04} 21.0 & 0.75 & 
\textbf{1663}  & \colorbox{lightgreen}{\relsize{-0.7}$\uparrow$3.91} 19.8 & 0.40 & 
\textbf{12742} & \colorbox{lightgreen}{\relsize{-0.7}$\uparrow$4.26} 19.9 & 0.35 \\
\textbf{22963} & \colorbox{lightgreen}{\relsize{-0.7}$\uparrow$7.83} 20.5 & 0.50 & 
\textbf{18953} & \colorbox{lightgreen}{\relsize{-0.7}$\uparrow$9.53} 20.9 & 0.40 & 
\textbf{12708} & \colorbox{lightgreen}{\relsize{-0.7}$\uparrow$3.81} 19.8 & 0.30 & 
\textbf{12770} & \colorbox{lightgreen}{\relsize{-0.7}$\uparrow$4.21} 19.9 & 0.65 \\
\textbf{20637} & \colorbox{lightgreen}{\relsize{-0.7}$\uparrow$7.23} 20.4 & 0.50 & 
\textbf{1027}  & \colorbox{lightgreen}{\relsize{-0.7}$\uparrow$9.13} 20.8 & 0.50 & 
\textbf{14077} & \colorbox{lightgreen}{\relsize{-0.7}$\uparrow$2.96} 19.6 & 0.35 & 
\textbf{4196}  & \colorbox{lightgreen}{\relsize{-0.7}$\uparrow$3.16} 19.7 & 0.70 \\
\textbf{5012}  & \colorbox{lightgreen}{\relsize{-0.7}$\uparrow$7.18} 20.4 & 0.60 & 
\textbf{543}   & \colorbox{lightgreen}{\relsize{-0.7}$\uparrow$9.03} 20.8 & 0.45 & 
\textbf{27077} & \colorbox{lightgreen}{\relsize{-0.7}$\uparrow$2.51} 19.5 & 0.50 & 
\textbf{21681} & \colorbox{lightgreen}{\relsize{-0.7}$\uparrow$2.76} 19.6 & 0.20 \\
\textbf{24898} & \colorbox{lightgreen}{\relsize{-0.7}$\uparrow$7.07} 20.4 & 0.45 & 
\textbf{27644} & \colorbox{lightgreen}{\relsize{-0.7}$\uparrow$8.23} 20.6 & 0.45 & 
\textbf{28389} & \colorbox{lightgreen}{\relsize{-0.7}$\uparrow$2.26} 19.5 & 0.50 & 
\textbf{27612} & \colorbox{lightgreen}{\relsize{-0.7}$\uparrow$2.21} 19.5 & 0.30 \\
\textbf{6704}  & \colorbox{lightgreen}{\relsize{-0.7}$\uparrow$6.42} 20.3 & 0.65 & 
\textbf{13092} & \colorbox{lightgreen}{\relsize{-0.7}$\uparrow$8.03} 20.6 & 0.40 & 
\textbf{20573} & \colorbox{lightgreen}{\relsize{-0.7}$\uparrow$2.26} 19.5 & 0.20 & 
\textbf{22617} & \colorbox{lightgreen}{\relsize{-0.7}$\uparrow$2.16} 19.5 & 0.60 \\
\bottomrule
\end{tabular}
}
\caption{Impactful latents for Gemma 2 9B in both directions (unknown and epochs). Gains relative to baseline: 19.1. All relative gains in green are in percentages.}
\label{tab:gemma_2_9b_combined}
\end{table}

\begin{table}[ht]
\centering
\resizebox{\textwidth}{!}{%
\begin{tabular}{lcr|lcr|lcr|lcr}
\multicolumn{12}{c}{\textbf{Model: Mistral 7B v0.3}} \\
\midrule

\multicolumn{12}{c}{\textbf{Direction: $\uparrow$Unknown}} \\
\midrule
\multicolumn{6}{c|}{\textbf{Increasing Latents}} & \multicolumn{6}{c}{\textbf{Decreasing Latents}} \\
\multicolumn{3}{c}{\textbf{Top-10 Positively Steered}} & \multicolumn{3}{c|}{\textbf{Top-10 Negatively Steered}} & \multicolumn{3}{c}{\textbf{Top-10 Positively Steered}} & \multicolumn{3}{c}{\textbf{Top-10 Negatively Steered}} \\
\midrule
\# & Acc. & $\alpha$ & \# & Acc. & $\alpha$ & \# & Acc. & $\alpha$ & \# & Acc. & $\alpha$ \\
\midrule
\textbf{41844} & \colorbox{lightgreen}{\relsize{-0.7}$\uparrow$24.91} 19.6 & 0.75 & 
\textbf{61325} & \colorbox{lightgreen}{\relsize{-0.7}$\uparrow$39.16} 21.8 & 0.55 & 
\textbf{18848} & \colorbox{lightgreen}{\relsize{-0.7}$\uparrow$31.24} 20.6 & 0.55 & 
\textbf{7859}  & \colorbox{lightgreen}{\relsize{-0.7}$\uparrow$39.77} 21.9 & 0.75 \\
\textbf{31953} & \colorbox{lightgreen}{\relsize{-0.7}$\uparrow$13.15} 17.8 & 0.50 & 
\textbf{37181} & \colorbox{lightgreen}{\relsize{-0.7}$\uparrow$31.61} 20.7 & 0.35 & 
\textbf{59897} & \colorbox{lightgreen}{\relsize{-0.7}$\uparrow$26.13} 19.8 & 0.70 & 
\textbf{18948} & \colorbox{lightgreen}{\relsize{-0.7}$\uparrow$24.60} 19.6 & 0.75 \\
\textbf{17552} & \colorbox{lightgreen}{\relsize{-0.7}$\uparrow$12.00} 17.6 & 0.45 & 
\textbf{42562} & \colorbox{lightgreen}{\relsize{-0.7}$\uparrow$29.54} 20.3 & 0.50 & 
\textbf{49882} & \colorbox{lightgreen}{\relsize{-0.7}$\uparrow$25.88} 19.8 & 0.65 & 
\textbf{38455} & \colorbox{lightgreen}{\relsize{-0.7}$\uparrow$21.32} 19.0 & 0.75 \\
\textbf{14493} & \colorbox{lightgreen}{\relsize{-0.7}$\uparrow$11.14} 17.4 & 0.45 & 
\textbf{48212} & \colorbox{lightgreen}{\relsize{-0.7}$\uparrow$29.05} 20.3 & 0.75 & 
\textbf{21258} & \colorbox{lightgreen}{\relsize{-0.7}$\uparrow$22.35} 19.2 & 0.50 & 
\textbf{32593} & \colorbox{lightgreen}{\relsize{-0.7}$\uparrow$19.31} 18.7 & 0.50 \\
\textbf{8648}  & \colorbox{lightgreen}{\relsize{-0.7}$\uparrow$11.02} 17.4 & 0.40 & 
\textbf{46283} & \colorbox{lightgreen}{\relsize{-0.7}$\uparrow$28.93} 20.2 & 0.60 & 
\textbf{31715} & \colorbox{lightgreen}{\relsize{-0.7}$\uparrow$20.40} 18.9 & 0.40 & 
\textbf{63016} & \colorbox{lightgreen}{\relsize{-0.7}$\uparrow$18.64} 18.6 & 0.70 \\
\textbf{16786} & \colorbox{lightgreen}{\relsize{-0.7}$\uparrow$10.54} 17.4 & 0.65 & 
\textbf{24790} & \colorbox{lightgreen}{\relsize{-0.7}$\uparrow$27.77} 20.1 & 0.60 & 
\textbf{47277} & \colorbox{lightgreen}{\relsize{-0.7}$\uparrow$19.61} 18.8 & 0.50 & 
\textbf{49380} & \colorbox{lightgreen}{\relsize{-0.7}$\uparrow$18.39} 18.6 & 0.70 \\
\textbf{3671}  & \colorbox{lightgreen}{\relsize{-0.7}$\uparrow$9.68} 17.2 & 0.40 & 
\textbf{5203}  & \colorbox{lightgreen}{\relsize{-0.7}$\uparrow$26.92} 19.9 & 0.55 & 
\textbf{53776} & \colorbox{lightgreen}{\relsize{-0.7}$\uparrow$19.55} 18.8 & 0.50 & 
\textbf{2883}  & \colorbox{lightgreen}{\relsize{-0.7}$\uparrow$15.35} 18.1 & 0.45 \\
\textbf{34318} & \colorbox{lightgreen}{\relsize{-0.7}$\uparrow$9.14} 17.1 & 0.55 & 
\textbf{55675} & \colorbox{lightgreen}{\relsize{-0.7}$\uparrow$26.61} 19.9 & 0.50 & 
\textbf{26746} & \colorbox{lightgreen}{\relsize{-0.7}$\uparrow$19.55} 18.8 & 0.65 & 
\textbf{4989}  & \colorbox{lightgreen}{\relsize{-0.7}$\uparrow$13.89} 17.9 & 0.70 \\
\textbf{57732} & \colorbox{lightgreen}{\relsize{-0.7}$\uparrow$8.89} 17.1 & 0.25 & 
\textbf{57875} & \colorbox{lightgreen}{\relsize{-0.7}$\uparrow$25.94} 19.8 & 0.40 & 
\textbf{5674}  & \colorbox{lightgreen}{\relsize{-0.7}$\uparrow$19.24} 18.7 & 0.50 & 
\textbf{50124} & \colorbox{lightgreen}{\relsize{-0.7}$\uparrow$13.09} 17.8 & 0.45 \\
\textbf{43093} & \colorbox{lightgreen}{\relsize{-0.7}$\uparrow$8.28} 17.0 & 0.45 & 
\textbf{65491} & \colorbox{lightgreen}{\relsize{-0.7}$\uparrow$24.36} 19.5 & 0.45 & 
\textbf{24894} & \colorbox{lightgreen}{\relsize{-0.7}$\uparrow$19.24} 18.7 & 0.70 & 
\textbf{50727} & \colorbox{lightgreen}{\relsize{-0.7}$\uparrow$13.09} 17.8 & 0.75 \\
\midrule

\multicolumn{12}{c}{\textbf{Direction: $\uparrow$Epochs}} \\
\midrule
\multicolumn{6}{c|}{\textbf{Increasing Latents}} & \multicolumn{6}{c}{\textbf{Decreasing Latents}} \\
\multicolumn{3}{c}{\textbf{Top-10 Positively Steered}} & \multicolumn{3}{c|}{\textbf{Top-10 Negatively Steered}} & \multicolumn{3}{c}{\textbf{Top-10 Positively Steered}} & \multicolumn{3}{c}{\textbf{Top-10 Negatively Steered}} \\
\midrule
\# & Acc. & $\alpha$ & \# & Acc. & $\alpha$ & \# & Acc. & $\alpha$ & \# & Acc. & $\alpha$ \\
\midrule
\textbf{46105} & \colorbox{lightgreen}{\relsize{-0.7}$\uparrow$22.41} 19.2 & 0.65 & 
\textbf{25922} & \colorbox{lightgreen}{\relsize{-0.7}$\uparrow$34.41} 21.1 & 0.60 & 
\textbf{9646}  & \colorbox{lightgreen}{\relsize{-0.7}$\uparrow$28.50} 20.2 & 0.50 & 
\textbf{61325} & \colorbox{lightgreen}{\relsize{-0.7}$\uparrow$39.16} 21.8 & 0.55 \\
\textbf{43123} & \colorbox{lightgreen}{\relsize{-0.7}$\uparrow$21.38} 19.1 & 0.45 & 
\textbf{48212} & \colorbox{lightgreen}{\relsize{-0.7}$\uparrow$29.05} 20.3 & 0.75 & 
\textbf{5387}  & \colorbox{lightgreen}{\relsize{-0.7}$\uparrow$27.47} 20.0 & 0.70 & 
\textbf{13089} & \colorbox{lightgreen}{\relsize{-0.7}$\uparrow$35.93} 21.3 & 0.65 \\
\textbf{7990}  & \colorbox{lightgreen}{\relsize{-0.7}$\uparrow$19.91} 18.8 & 0.75 & 
\textbf{48172} & \colorbox{lightgreen}{\relsize{-0.7}$\uparrow$21.99} 19.1 & 0.50 & 
\textbf{41844} & \colorbox{lightgreen}{\relsize{-0.7}$\uparrow$24.91} 19.6 & 0.75 & 
\textbf{42562} & \colorbox{lightgreen}{\relsize{-0.7}$\uparrow$29.54} 20.3 & 0.50 \\
\textbf{43991} & \colorbox{lightgreen}{\relsize{-0.7}$\uparrow$16.08} 18.2 & 0.30 & 
\textbf{13512} & \colorbox{lightgreen}{\relsize{-0.7}$\uparrow$21.25} 19.0 & 0.75 & 
\textbf{22642} & \colorbox{lightgreen}{\relsize{-0.7}$\uparrow$24.18} 19.5 & 0.60 & 
\textbf{6634}  & \colorbox{lightgreen}{\relsize{-0.7}$\uparrow$23.51} 19.4 & 0.75 \\
\textbf{33778} & \colorbox{lightgreen}{\relsize{-0.7}$\uparrow$15.04} 18.1 & 0.45 & 
\textbf{51982} & \colorbox{lightgreen}{\relsize{-0.7}$\uparrow$14.92} 18.0 & 0.50 & 
\textbf{25600} & \colorbox{lightgreen}{\relsize{-0.7}$\uparrow$23.39} 19.4 & 0.50 & 
\textbf{62464} & \colorbox{lightgreen}{\relsize{-0.7}$\uparrow$19.31} 18.7 & 0.55 \\
\textbf{53702} & \colorbox{lightgreen}{\relsize{-0.7}$\uparrow$14.98} 18.0 & 0.55 & 
\textbf{2567}  & \colorbox{lightgreen}{\relsize{-0.7}$\uparrow$14.25} 17.9 & 0.45 & 
\textbf{44213} & \colorbox{lightgreen}{\relsize{-0.7}$\uparrow$22.84} 19.3 & 0.55 & 
\textbf{49380} & \colorbox{lightgreen}{\relsize{-0.7}$\uparrow$18.39} 18.6 & 0.70 \\
\textbf{12362} & \colorbox{lightgreen}{\relsize{-0.7}$\uparrow$14.74} 18.0 & 0.50 & 
\textbf{51906} & \colorbox{lightgreen}{\relsize{-0.7}$\uparrow$13.95} 17.9 & 0.50 & 
\textbf{45825} & \colorbox{lightgreen}{\relsize{-0.7}$\uparrow$20.83} 19.0 & 0.55 & 
\textbf{34854} & \colorbox{lightgreen}{\relsize{-0.7}$\uparrow$18.09} 18.5 & 0.65 \\
\textbf{8644}  & \colorbox{lightgreen}{\relsize{-0.7}$\uparrow$14.56} 18.0 & 0.45 & 
\textbf{27411} & \colorbox{lightgreen}{\relsize{-0.7}$\uparrow$13.76} 17.9 & 0.50 & 
\textbf{2989}  & \colorbox{lightgreen}{\relsize{-0.7}$\uparrow$19.12} 18.7 & 0.60 & 
\textbf{43955} & \colorbox{lightgreen}{\relsize{-0.7}$\uparrow$17.90} 18.5 & 0.55 \\
\textbf{25464} & \colorbox{lightgreen}{\relsize{-0.7}$\uparrow$12.85} 17.7 & 0.30 & 
\textbf{6232}  & \colorbox{lightgreen}{\relsize{-0.7}$\uparrow$13.15} 17.8 & 0.75 & 
\textbf{64057} & \colorbox{lightgreen}{\relsize{-0.7}$\uparrow$17.84} 18.5 & 0.60 & 
\textbf{47101} & \colorbox{lightgreen}{\relsize{-0.7}$\uparrow$16.26} 18.2 & 0.75 \\
\textbf{14493} & \colorbox{lightgreen}{\relsize{-0.7}$\uparrow$11.14} 17.4 & 0.45 & 
\textbf{34121} & \colorbox{lightgreen}{\relsize{-0.7}$\uparrow$12.91} 17.7 & 0.50 & 
\textbf{54468} & \colorbox{lightgreen}{\relsize{-0.7}$\uparrow$17.78} 18.5 & 0.70 & 
\textbf{36392} & \colorbox{lightgreen}{\relsize{-0.7}$\uparrow$15.71} 18.2 & 0.75 \\
\bottomrule
\end{tabular}
}
\caption{Impactful latents for Mistral 7B v0.3 across both directions (unknown and epochs). Gains relative to baseline: 15.7. All relative gains in green are in percentages.}
\label{tab:mistral_7b_combined}
\end{table}

\begin{table}[ht]
\centering
\resizebox{\textwidth}{!}{%
\begin{tabular}{lcr|lcr|lcr|lcr}
\multicolumn{12}{c}{\textbf{Direction: $\uparrow$Epochs}} \\
\midrule
\multicolumn{6}{c|}{\textbf{Increasing Latents}} & \multicolumn{6}{c}{\textbf{Decreasing Latents}} \\
\multicolumn{3}{c}{\textbf{Top-10 Positively Steered}} & \multicolumn{3}{c|}{\textbf{Top-10 Negatively Steered}} & \multicolumn{3}{c}{\textbf{Top-10 Positively Steered}} & \multicolumn{3}{c}{\textbf{Top-10 Negatively Steered}} \\
\midrule
\# & Acc. & $\alpha$ & \# & Acc. & $\alpha$ & \# & Acc. & $\alpha$ & \# & Acc. & $\alpha$ \\
\midrule
\href{https://www.neuronpedia.org/llama3.1-8b/15-llamascope-res-32k/18915}{\textbf{18915}} & \colorbox{lightgreen}{\relsize{-0.7}$\uparrow$9.43} 19.5 & 0.45 & 
\href{https://www.neuronpedia.org/llama3.1-8b/15-llamascope-res-32k/1705}{\textbf{1705}} & \colorbox{lightgreen}{\relsize{-0.7}$\uparrow$7.87} 19.3 & 0.60 & 
\href{https://www.neuronpedia.org/llama3.1-8b/15-llamascope-res-32k/5796}{\textbf{5796}} & \colorbox{lightgreen}{\relsize{-0.7}$\uparrow$8.84} 19.4 & 0.45 & 
\href{https://www.neuronpedia.org/llama3.1-8b/15-llamascope-res-32k/32589}{\textbf{32589}} & \colorbox{lightgreen}{\relsize{-0.7}$\uparrow$4.61} 18.7 & 0.35 \\
\href{https://www.neuronpedia.org/llama3.1-8b/15-llamascope-res-32k/8842}{\textbf{8842}} & \colorbox{lightgreen}{\relsize{-0.7}$\uparrow$8.14} 19.3 & 0.40 & 
\href{https://www.neuronpedia.org/llama3.1-8b/15-llamascope-res-32k/25557}{\textbf{25557}} & \colorbox{lightgreen}{\relsize{-0.7}$\uparrow$7.66} 19.2 & 0.40 & 
\href{https://www.neuronpedia.org/llama3.1-8b/15-llamascope-res-32k/27191}{\textbf{27191}} & \colorbox{lightgreen}{\relsize{-0.7}$\uparrow$7.12} 19.1 & 0.50 & 
\href{https://www.neuronpedia.org/llama3.1-8b/15-llamascope-res-32k/15177}{\textbf{15177}} & \colorbox{lightgreen}{\relsize{-0.7}$\uparrow$3.59} 18.5 & 0.65 \\
\href{https://www.neuronpedia.org/llama3.1-8b/15-llamascope-res-32k/6032}{\textbf{6032}} & \colorbox{lightgreen}{\relsize{-0.7}$\uparrow$3.64} 18.5 & 0.30 & 
\href{https://www.neuronpedia.org/llama3.1-8b/15-llamascope-res-32k/9628}{\textbf{9628}} & \colorbox{lightgreen}{\relsize{-0.7}$\uparrow$5.78} 18.9 & 0.45 & 
\href{https://www.neuronpedia.org/llama3.1-8b/15-llamascope-res-32k/17666}{\textbf{17666}} & \colorbox{lightgreen}{\relsize{-0.7}$\uparrow$5.14} 18.8 & 0.70 & 
\href{https://www.neuronpedia.org/llama3.1-8b/15-llamascope-res-32k/332}{\textbf{332}} & \colorbox{lightgreen}{\relsize{-0.7}$\uparrow$2.25} 18.2 & 0.25 \\
\href{https://www.neuronpedia.org/llama3.1-8b/15-llamascope-res-32k/11033}{\textbf{11033}} & \colorbox{lightgreen}{\relsize{-0.7}$\uparrow$3.00} 18.4 & 0.30 & 
\href{https://www.neuronpedia.org/llama3.1-8b/15-llamascope-res-32k/340}{\textbf{340}} & \colorbox{lightgreen}{\relsize{-0.7}$\uparrow$3.70} 18.5 & 0.35 & 
\href{https://www.neuronpedia.org/llama3.1-8b/15-llamascope-res-32k/20884}{\textbf{20884}} & \colorbox{lightgreen}{\relsize{-0.7}$\uparrow$3.59} 18.5 & 0.20 & 
\href{https://www.neuronpedia.org/llama3.1-8b/15-llamascope-res-32k/15105}{\textbf{15105}} & \colorbox{lightgreen}{\relsize{-0.7}$\uparrow$1.12} 18.0 & 0.20 \\
\href{https://www.neuronpedia.org/llama3.1-8b/15-llamascope-res-32k/27676}{\textbf{27676}} & \colorbox{lightgreen}{\relsize{-0.7}$\uparrow$2.79} 18.3 & 0.55 & 
\href{https://www.neuronpedia.org/llama3.1-8b/15-llamascope-res-32k/29538}{\textbf{29538}} & \colorbox{lightgreen}{\relsize{-0.7}$\uparrow$3.05} 18.4 & 0.25 & 
\href{https://www.neuronpedia.org/llama3.1-8b/15-llamascope-res-32k/26381}{\textbf{26381}} & \colorbox{lightgreen}{\relsize{-0.7}$\uparrow$3.32} 18.4 & 0.35 & 
& & \\
\href{https://www.neuronpedia.org/llama3.1-8b/15-llamascope-res-32k/14571}{\textbf{14571}} & \colorbox{lightgreen}{\relsize{-0.7}$\uparrow$2.04} 18.2 & 0.15 & 
\href{https://www.neuronpedia.org/llama3.1-8b/15-llamascope-res-32k/23472}{\textbf{23472}} & \colorbox{lightgreen}{\relsize{-0.7}$\uparrow$2.79} 18.3 & 0.35 & 
\href{https://www.neuronpedia.org/llama3.1-8b/15-llamascope-res-32k/30382}{\textbf{30382}} & \colorbox{lightgreen}{\relsize{-0.7}$\uparrow$3.21} 18.4 & 0.35 & 
& & \\
\href{https://www.neuronpedia.org/llama3.1-8b/15-llamascope-res-32k/17177}{\textbf{17177}} & \colorbox{lightgreen}{\relsize{-0.7}$\uparrow$1.87} 18.2 & 0.15 & 
\href{https://www.neuronpedia.org/llama3.1-8b/15-llamascope-res-32k/17281}{\textbf{17281}} & \colorbox{lightgreen}{\relsize{-0.7}$\uparrow$2.68} 18.3 & 0.35 & 
\href{https://www.neuronpedia.org/llama3.1-8b/15-llamascope-res-32k/32309}{\textbf{32309}} & \colorbox{lightgreen}{\relsize{-0.7}$\uparrow$2.79} 18.3 & 0.20 & 
& & \\
\href{https://www.neuronpedia.org/llama3.1-8b/15-llamascope-res-32k/11205}{\textbf{11205}} & \colorbox{lightgreen}{\relsize{-0.7}$\uparrow$1.77} 18.2 & 0.50 & 
\href{https://www.neuronpedia.org/llama3.1-8b/15-llamascope-res-32k/15229}{\textbf{15229}} & \colorbox{lightgreen}{\relsize{-0.7}$\uparrow$2.52} 18.3 & 0.20 & 
\href{https://www.neuronpedia.org/llama3.1-8b/15-llamascope-res-32k/11134}{\textbf{11134}} & \colorbox{lightgreen}{\relsize{-0.7}$\uparrow$2.41} 18.3 & 0.35 & 
& & \\
\href{https://www.neuronpedia.org/llama3.1-8b/15-llamascope-res-32k/9834}{\textbf{9834}} & \colorbox{lightgreen}{\relsize{-0.7}$\uparrow$1.50} 18.1 & 0.30 & 
\href{https://www.neuronpedia.org/llama3.1-8b/15-llamascope-res-32k/19538}{\textbf{19538}} & \colorbox{lightgreen}{\relsize{-0.7}$\uparrow$2.46} 18.3 & 0.20 & 
\href{https://www.neuronpedia.org/llama3.1-8b/15-llamascope-res-32k/5094}{\textbf{5094}} & \colorbox{lightgreen}{\relsize{-0.7}$\uparrow$2.36} 18.3 & 0.30 & 
& & \\
\href{https://www.neuronpedia.org/llama3.1-8b/15-llamascope-res-32k/30670}{\textbf{30670}} & \colorbox{lightgreen}{\relsize{-0.7}$\uparrow$1.45} 18.1 & 0.60 & 
\href{https://www.neuronpedia.org/llama3.1-8b/15-llamascope-res-32k/31426}{\textbf{31426}} & \colorbox{lightgreen}{\relsize{-0.7}$\uparrow$2.41} 18.3 & 0.25 & 
\href{https://www.neuronpedia.org/llama3.1-8b/15-llamascope-res-32k/2378}{\textbf{2378}} & \colorbox{lightgreen}{\relsize{-0.7}$\uparrow$2.09} 18.2 & 0.30 & 
& & \\
\bottomrule
\end{tabular}
}
\caption{Impactful latents for Llama 3.1 8B in the epochs direction. Gains relative to baseline: 17.8. All relative gains in green are in percentages.}
\label{tab:llama31_8b_epo}
\end{table}

\clearpage

\section{Latent Similarity}
\label{appendix:heatmaps}

\begin{figure}[htbp]
    \centering
    \includegraphics[width=\textwidth]{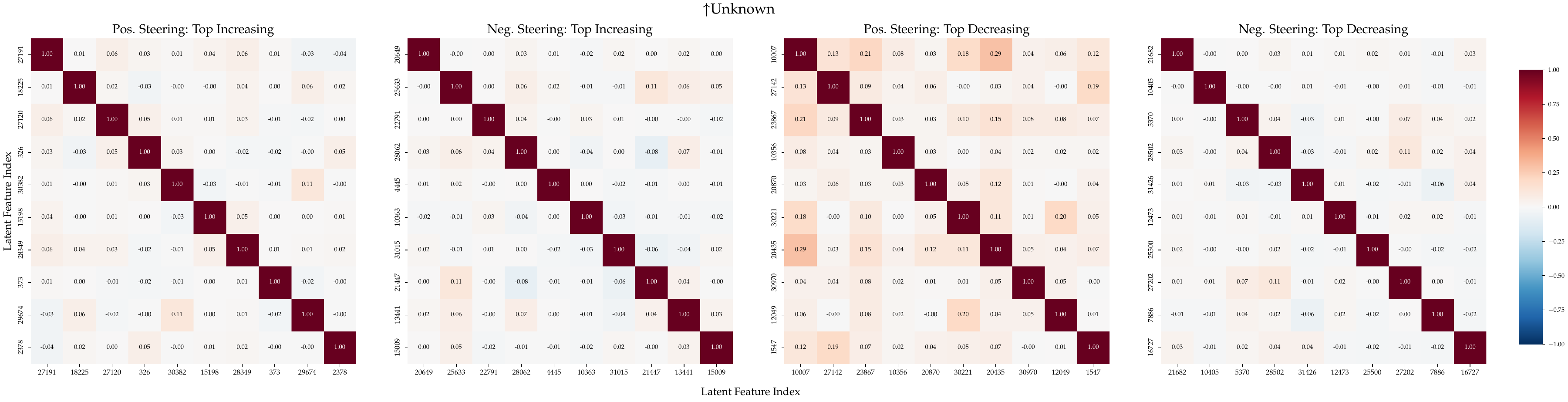}
    
    \vspace{0.5cm}
    
    \includegraphics[width=\textwidth]{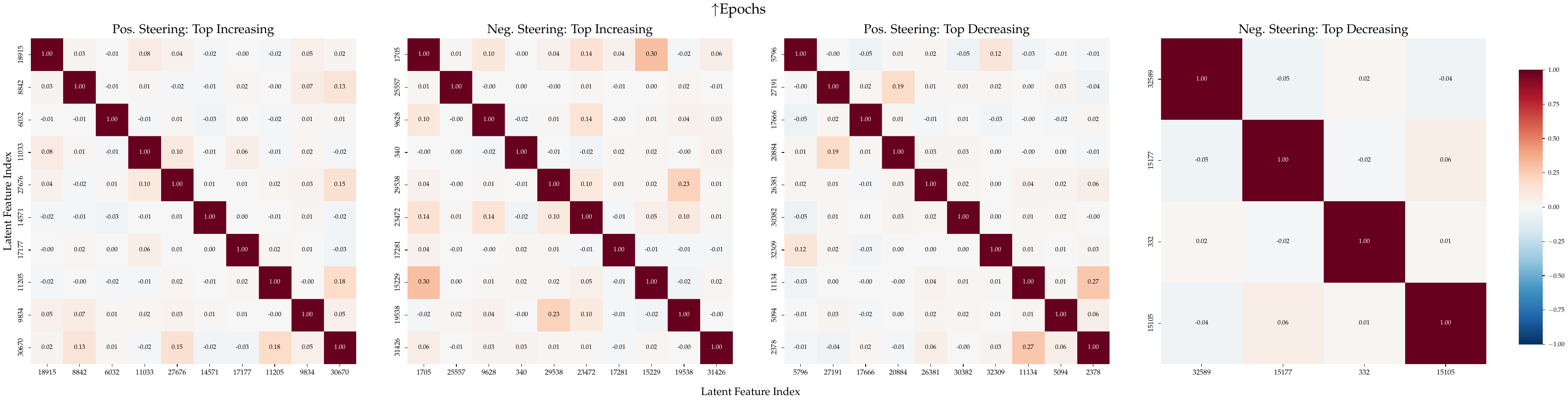}
    
    \caption{Cosine similarity of the top-10 performing latents for both directions (unknown and epochs) sourced by MoRFI on \textbf{Llama 3.1 8B}.}
    \label{fig:llama_heatmaps_combined}
\end{figure}

\begin{figure}[htbp]
    \centering    
    \includegraphics[width=\textwidth]{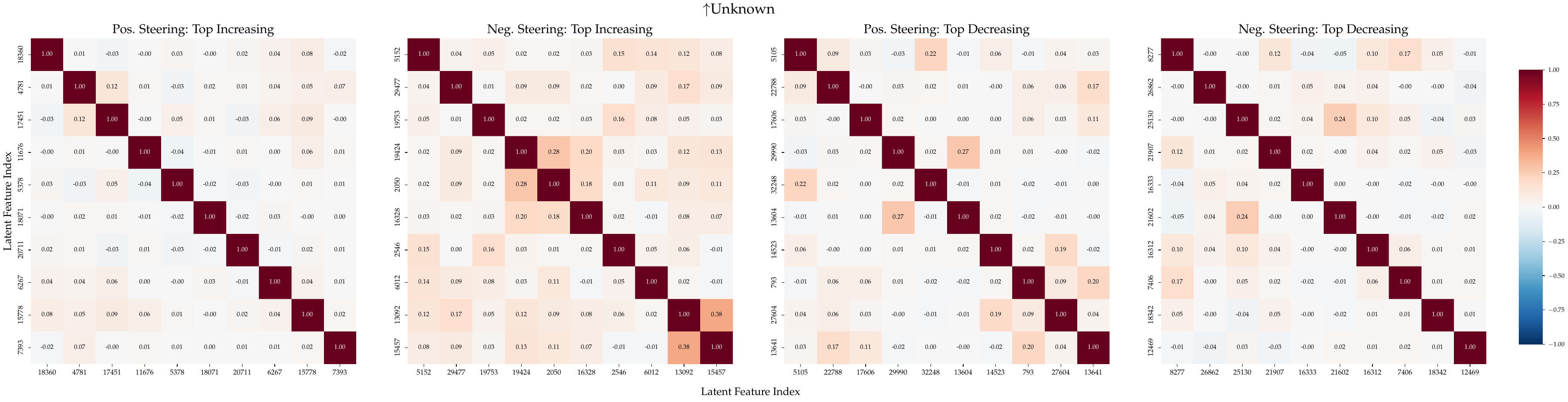}

    \vspace{0.5cm}    

    \includegraphics[width=\textwidth]{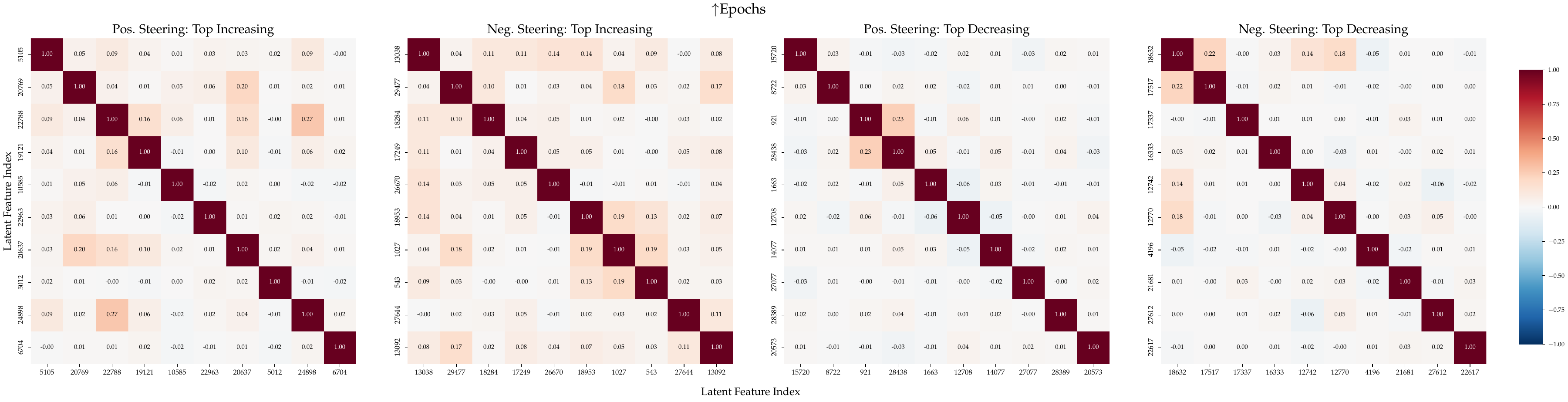}
    \caption{Cosine similarity of the top-10 performing latents for both directions unknown and epochs) sourced by MoRFI on \textbf{Gemma 2 9B}.}    
    \label{fig:gemma_heatmaps_combined}
\end{figure}

\begin{figure}[htbp]
    \centering    
    \includegraphics[width=\textwidth]{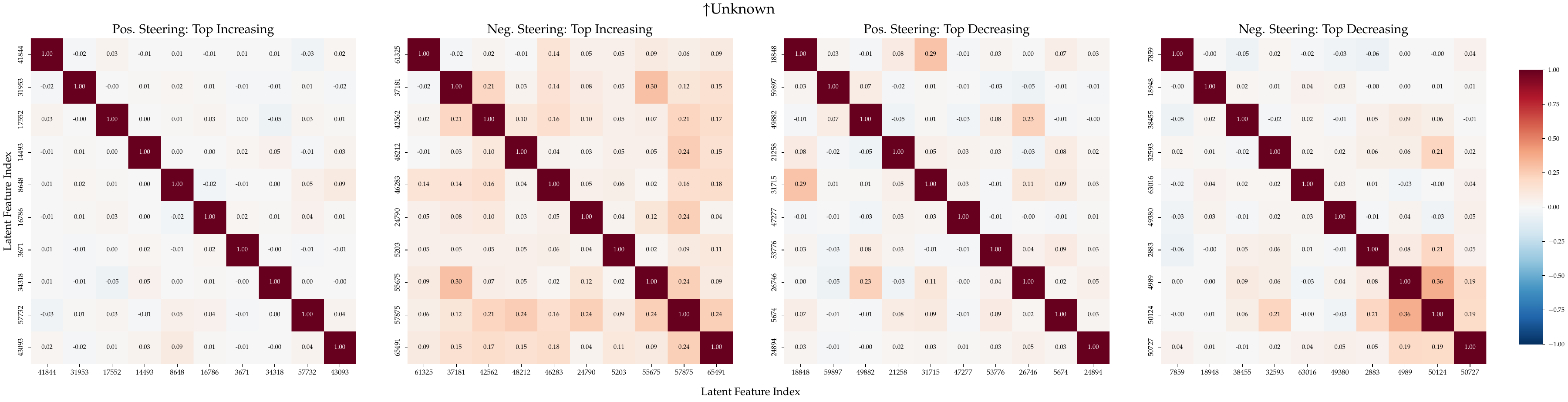}

    \vspace{0.5cm}    

    \includegraphics[width=\textwidth]{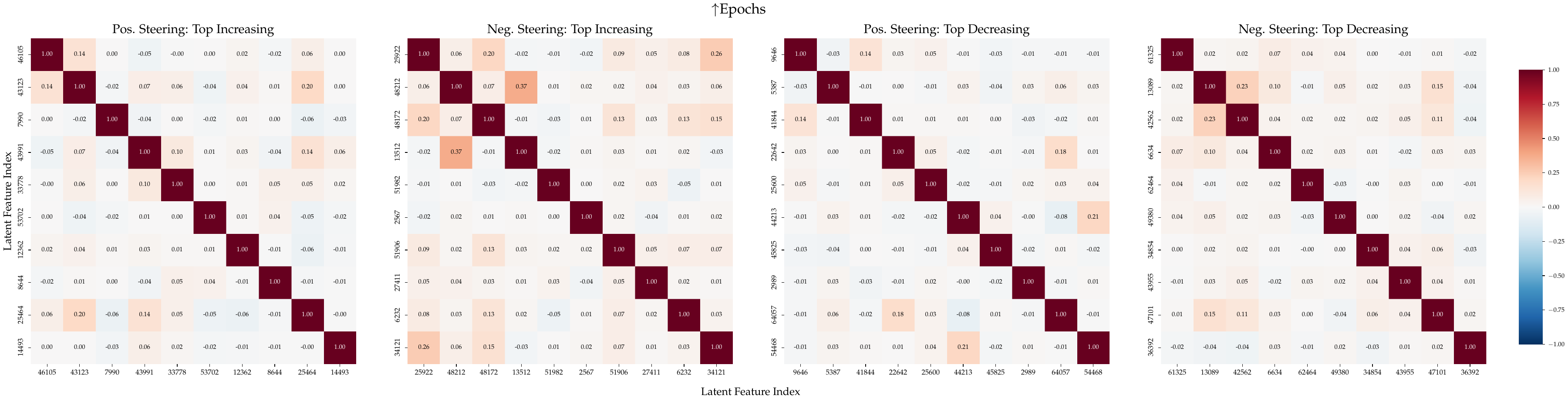}
    \caption{Cosine similarity of the top-10 performing latents for both directions (unknown and epochs) sourced by MoRFI on \textbf{Mistral v03 7B}.}
    \label{fig:mistral_heatmaps_combined}
\end{figure}

In Figures~\ref{fig:llama_heatmaps_combined}-\ref{fig:mistral_heatmaps_combined} we plot the cosine similarity of the top-performing latents, across all three models. 

\clearpage

\section{Control Groups}
\label{appendix:control}
\begin{figure}[htbp]
    \centering    
    \includegraphics[width=\textwidth, trim={0 0 0 1.1cm}, clip]{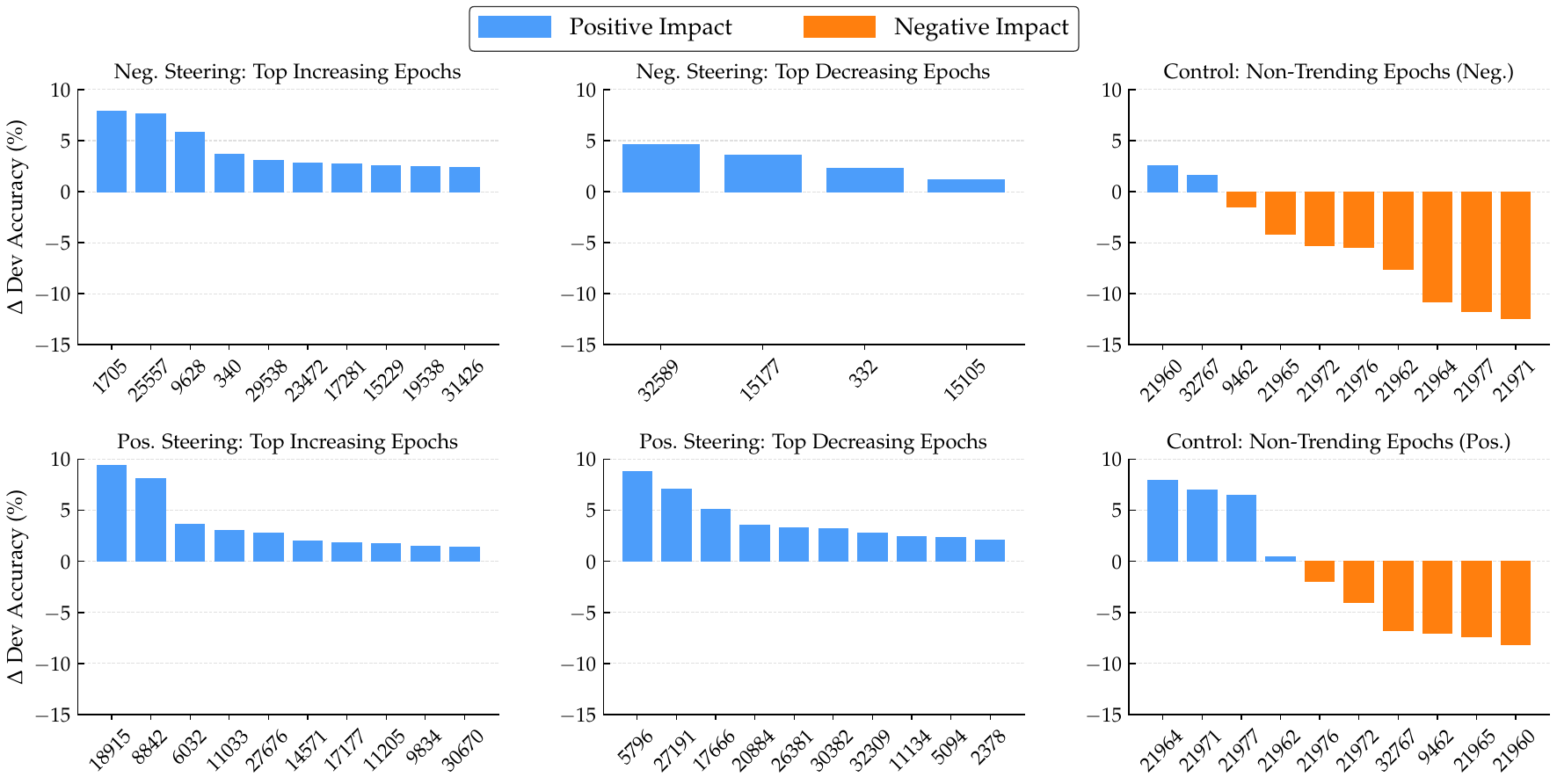}    
    \caption{Dev accuracy change from single-latent steering on \textbf{Llama 3.1 8B} along the epochs dimension vs control group.}
    \label{fig:conv_llama}
\end{figure}

\begin{figure}[htbp]
    \centering    
    \includegraphics[width=\textwidth, trim={0 0 0 1.1cm}, clip]{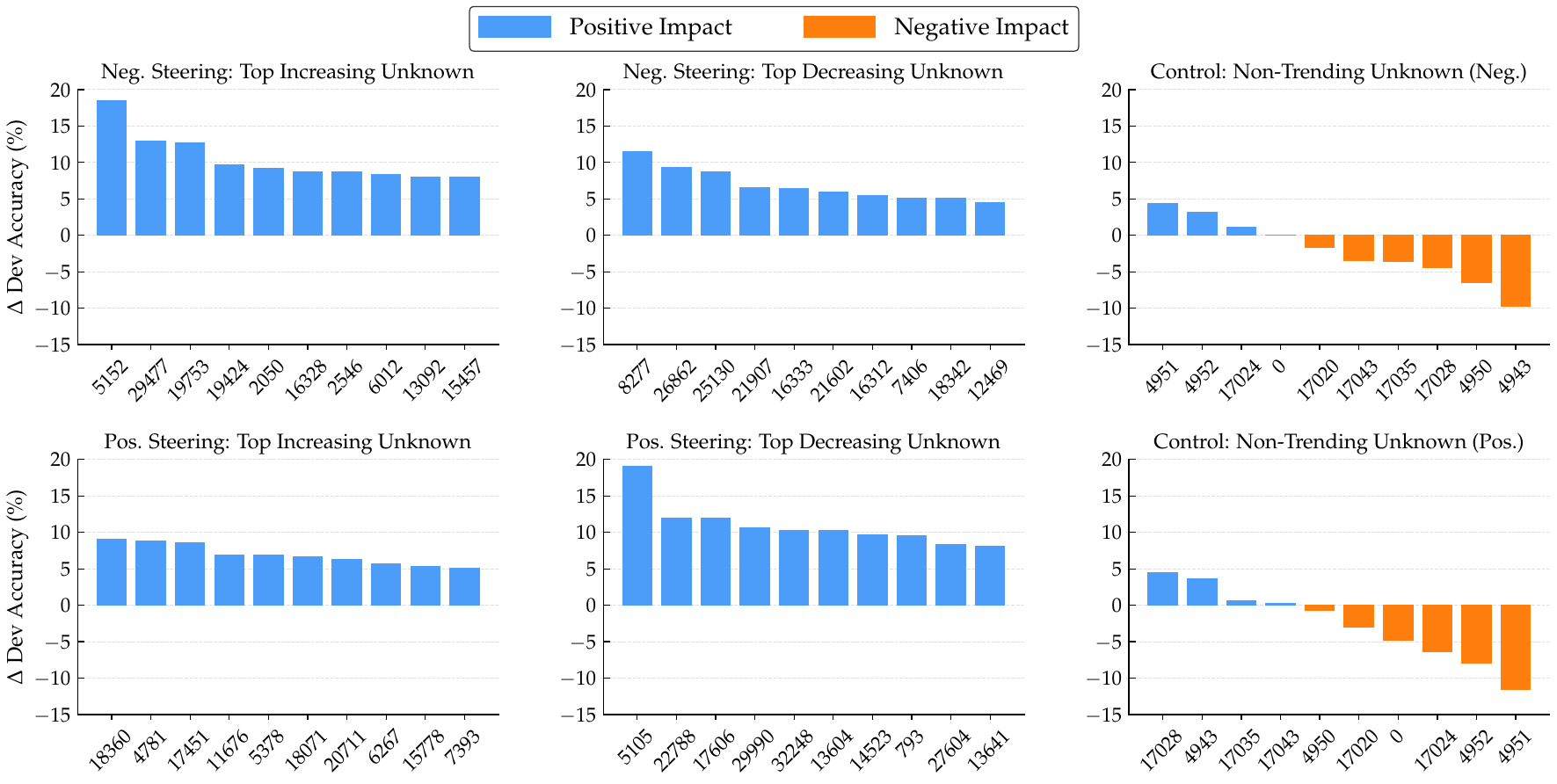}    
    \caption{Dev accuracy change from single-latent steering on \textbf{Gemma 2 9B} along the unknown dimension vs control group.}
    \label{fig:unk_gemma}
\end{figure}

\begin{figure}[htbp]
    \centering    
    \includegraphics[width=\textwidth, trim={0 0 0 1.1cm}, clip]{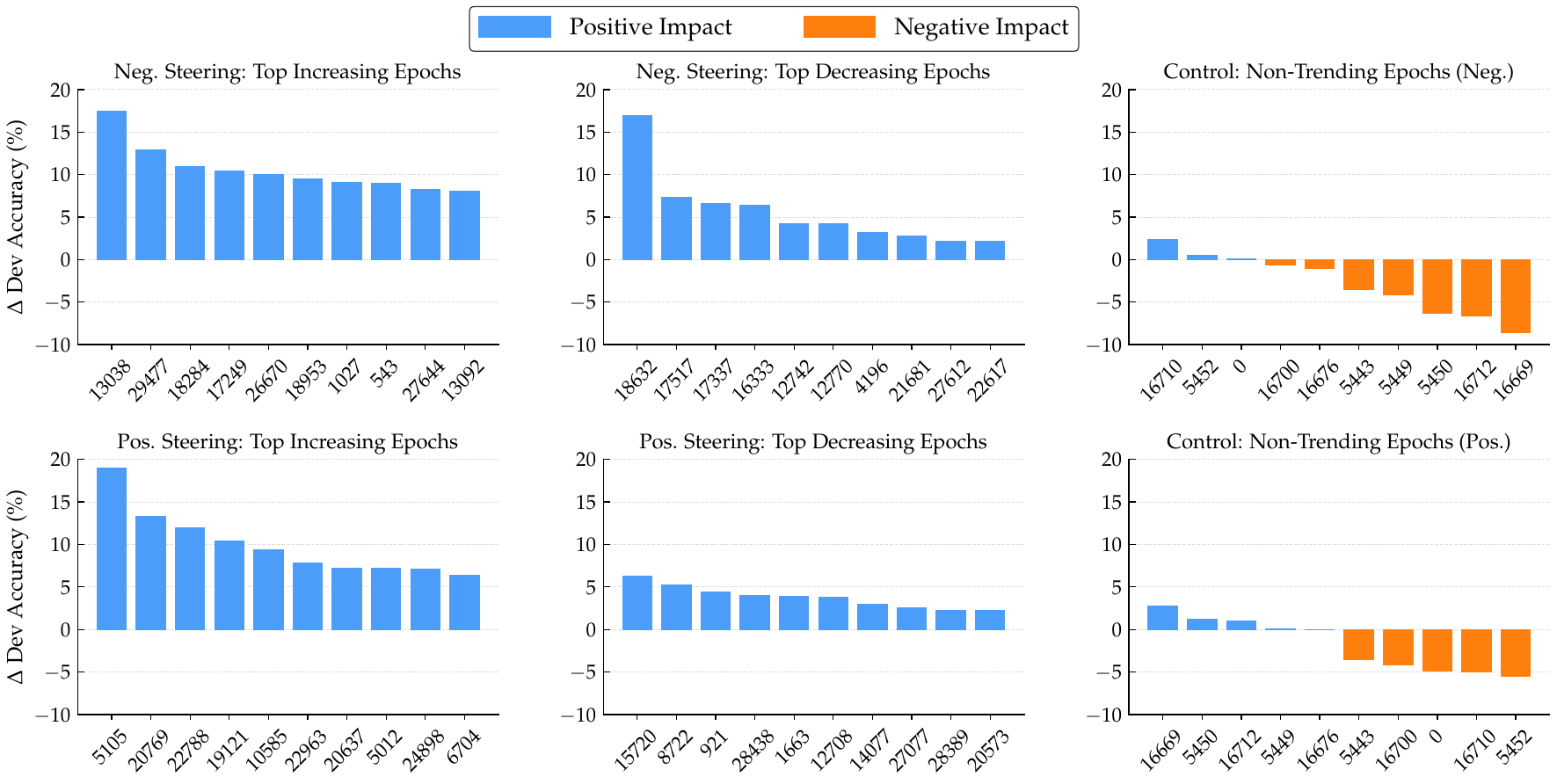}    
    \caption{Dev accuracy change from single-latent steering on \textbf{Gemma 2 9B} along the epochs dimension vs control group.}
    \label{fig:conv_gemma}
\end{figure}

\begin{figure}[htbp]
    \centering    
    \includegraphics[width=\textwidth, trim={0 0 0 1.1cm}, clip]{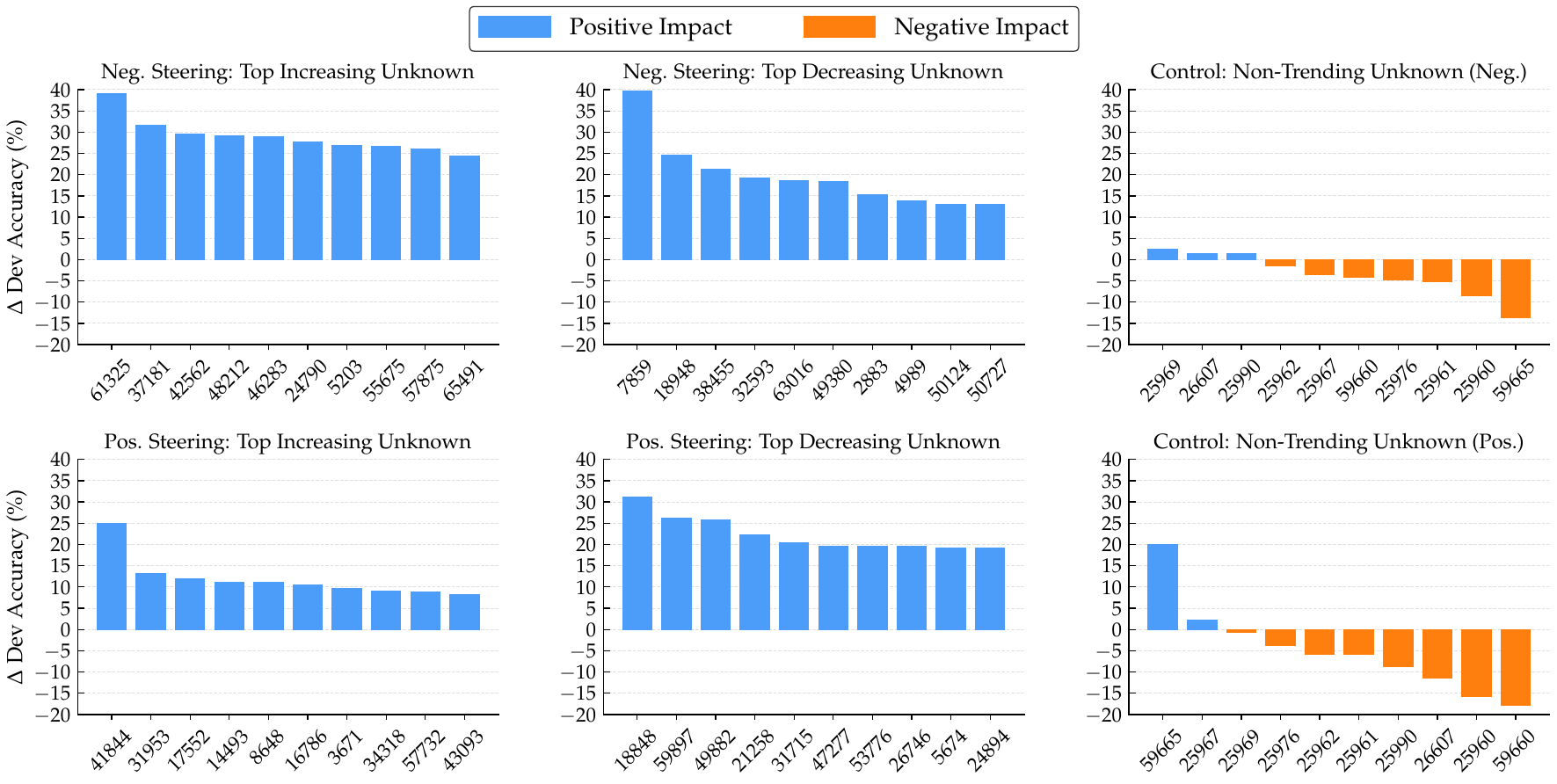}    
    \caption{Dev accuracy change from single-latent steering on \textbf{Mistral v03 7B} along the unknown dimension vs control group.}
    \label{fig:unk_mistral}
\end{figure}

\begin{figure}[htbp]
    \centering    
    \includegraphics[width=\textwidth, trim={0 0 0 1.1cm}, clip]{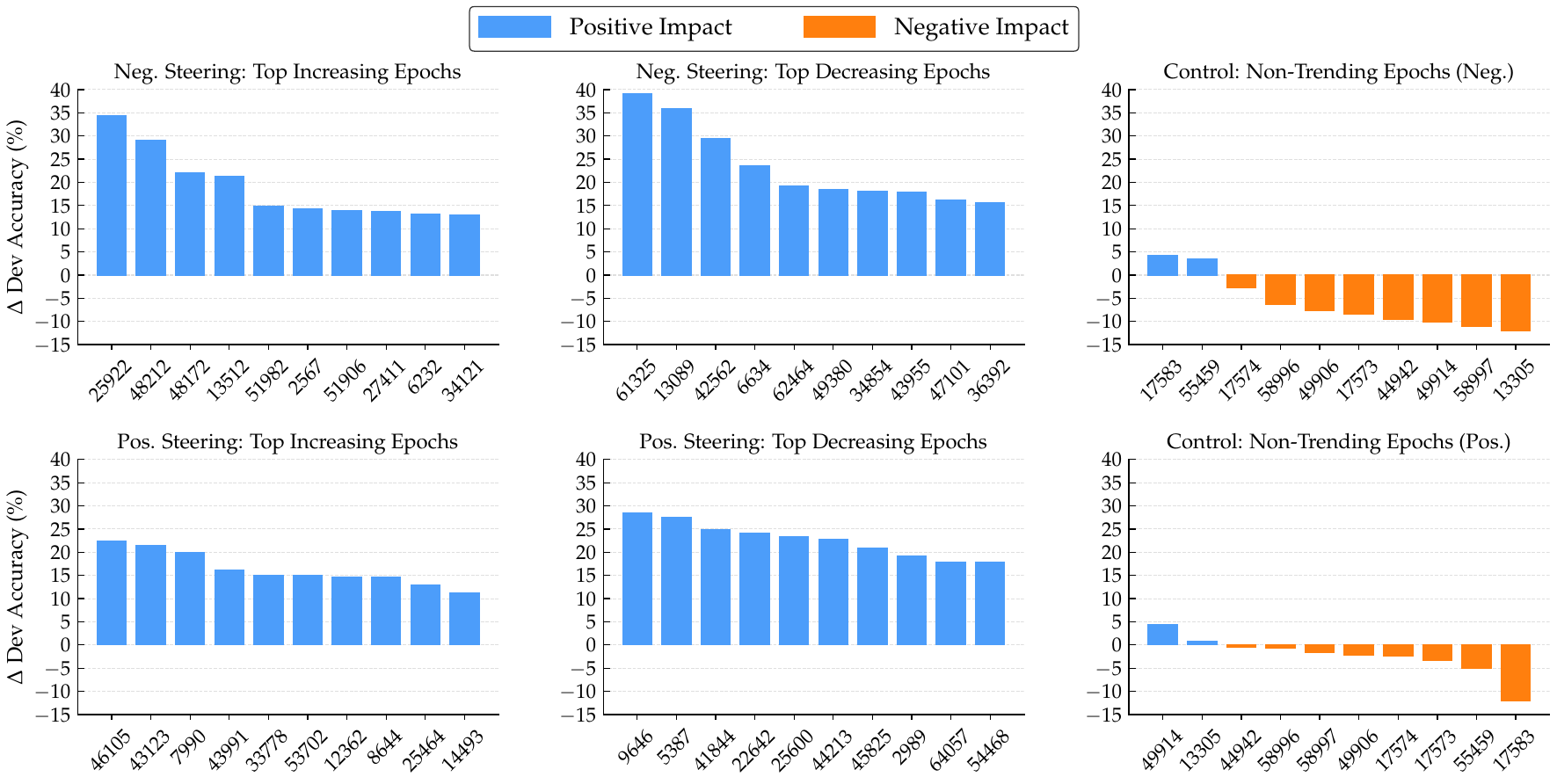}    
    \caption{Dev accuracy change from single-latent steering on \textbf{Mistral v03 7B} along the epochs dimension vs control group.}
    \label{fig:conv_mistral}
\end{figure}

In Figures~\ref{fig:conv_llama},~\ref{fig:unk_llama},~\ref{fig:unk_gemma},~\ref{fig:conv_gemma},~\ref{fig:unk_mistral} and ~\ref{fig:conv_mistral}, we present Dev accuracy gains relative to the unsteered checkpoint (first two columns) versus the control group performance (last column) for all models.

\clearpage

\section{Additional Algorithms}
\label{appendix:additional}

We present Algorithms~\ref{alg:idf} and ~\ref{alg:ctrl} and brief explanations of the main tensor operations used by them in Table~\ref{table:defn}. Algorithm~\ref{alg:idf} identifies the top-10 most impactful latents based on input from Algorithm~\ref{alg:core}, while Algorithm~\ref{alg:ctrl} selects features that minimally change across data-mixtures to compare against the ones sourced by MoRFI.

\begin{table}[htbp]
\centering
\renewcommand{\arraystretch}{1.6} %
\resizebox{\textwidth}{!}{%
\begin{tabular}{@{} l l p{7cm} @{}}
\toprule
\textbf{Function} & \textbf{Signature ($A \mapsto B$)} & \textbf{Operation / Definition} \\
\midrule

\textbf{MaskedMeanFold} & 
$(Z, B) \in \mathbb{R}^{N \times S \times d_{\textit{sae}}} \times \{0,1\}^{N \times S \times 1} \mapsto \bar{Z} \in \mathbb{R}^{N \times d_{\textit{sae}}}$ & 
$\Big(\sum_{k=1}^{S} Z_{:, k, :} \odot (B_{:, k, :} \mathbf{1}_{d_{\textit{sae}}}^\top)\Big) \, / \, \Big(\sum_{k=1}^{S} B_{:, k, :}\mathbf{1}_{d_{\textit{sae}}}^\top\Big)$ \\
 & \multicolumn{2}{l}{Computes the mean of sparse representations $Z$ over the tokens where the boolean mask $B$ is active.} \\
\textbf{MeanFold} & 
$A \mapsto \bar{A} \in \mathbb{R}^{d_1 \times \dots \times d_{X-1} \times d_{X+1} \times \dots \times d_m}$ & 
$\bar{A}_{i_1, \dots, i_m} = \frac{1}{d_X} \sum_{x=1}^{d_X} A_{\dots, x, \dots}$ \\
 & \multicolumn{2}{l}{Collapses dimension $d_X$ by computing the mean of its elements.} \\

\textbf{SampleUniform} & 
$\emptyset \mapsto \mathcal{I} \in \{a, \dots, b\}^{d_1 \times \dots \times d_m}$ & 
$\mathcal{I}_{i_1, \dots, i_m} \sim \mathcal{U}\{a, b\} \quad \text{i.i.d.}$ \\
 & \multicolumn{2}{l}{Generates uniformly distributed resampling indices.} \\

\textbf{BootstrapFold} & 
$A \mapsto \mathbf{M}^* \in \mathbb{R}^{d_1 \times \dots \times R \times \dots \times d_m}$ & 
$\mathbf{M}^*_{\dots, r, \dots} = \sum_{x=1}^{d_X} \mathcal{W}_{r, x} A_{\dots, \mathcal{I}^*_{r, x}, \dots}$ \\
 & \multicolumn{2}{l}{Generates a bootstrapped replicate tensor $\mathbf{M}^*$ folding $A$ over $d_X$.} \\

\textbf{SpearmanFold} & 
$A \mapsto \boldsymbol{\rho}^* \in \mathbb{R}^{d_1 \times \dots \times d_{X-1} \times d_{X+1} \times \dots \times d_m}$ & 
$\boldsymbol{\rho}^*_{i_1, \dots, i_m} = \text{SpearmanR}(A_{\dots, \mathbf{:}, \dots}, \, \mathbf{v})$ \\
 & \multicolumn{2}{l}{Computes Spearman rank correlation coefficient between tensor slices and a reference vector $\mathbf{v}$.} \\

\textbf{MKFold} & 
$A \mapsto \boldsymbol{\tau}^* \in \mathbb{R}^{d_1 \times \dots \times d_{X-1} \times d_{X+1} \times \dots \times d_m}$ & 
$\boldsymbol{\tau}^*_{i_1, \dots, i_m} = \text{MKTest}(A_{\dots, \mathbf{:}, \dots})$ \\
 & \multicolumn{2}{l}{Folds the specified dimension to Kendall's tau ($\tau$).} \\

\textbf{TopKIndices} &
$A \mapsto L \in \{0, \dots, d_X-1\}^{d_1 \times \dots \times k \times \dots \times d_m}$ &
$L_{\dots, x, \dots} = \text{argsort}(A_{\dots, \mathbf{:}, \dots})_x \text{ for } x < k$ \\
 & \multicolumn{2}{l}{Extracts the specific indices that correspond to the $k$ largest values.} \\

\textbf{IndexCount} & 
$L \in \{0, \dots, D-1\}^{d_1 \times \dots \times d_m} \mapsto C \in \mathbb{N}^D$ & 
$C_d = \sum_{i_1=1}^{d_1} \dots \sum_{i_m=1}^{d_m} \mathbb{I}(L_{i_1, \dots, i_m} = d)$ \\
 & \multicolumn{2}{l}{Computes frequencies for each component of $L$.} \\
\bottomrule
\end{tabular}%
}
\caption{Definitions of the tensor operations used in MoRFI. For operations applied along a named dimension $X$ of size $d_X$, the tuple $(i_1, \dots, i_m)$ denotes the arbitrary indices of the remaining dimensions. The notation $A_{\dots, x, \dots}$ indicates indexing at position $x$ along dimension $X$. $B_{n, k, 1} = \mathbb{I}(b_n < k < b_n + c_n) \quad \forall n \in \{1 \dots N\}, \; k \in \{1 \dots S\}$ corresponds to a boolean mask filtering out BOS tokens.\label{table:defn}}
\end{table}

\begin{algorithm}
\caption{Identification of High-Impact Latents via Activation Steering\label{alg:idf}}
\begin{algorithmic}[1]
    \Require 
        \State $S$: Sequence of candidate latent indices (e.g., from $S^{\uparrow}$ or $S^{\downarrow}$)
        \State $c \in \{-1, 1\}$: Steering direction ($-1$ for suppression, $+1$ for amplification)
        \State $\mathbf{\Phi} \in \mathbb{R}^{F \times d_{\textit{model}}}$: SAE dictionary of feature directions
        \State $\mathcal{D}_{\textit{dev}}$: Reference dataset for evaluation
        \State $\ell = \lfloor L/2 \rfloor$: Target injection layer for the $L$-layer model, though applicable to any layer
        \State $M$: Base model; $M(\cdot \oplus_{\ell} \mathbf{v})$ denotes $M$ with hidden states shifted by $\mathbf{v}$ at layer $\ell$
        \State $A_{\textit{base}} \in [0, 1]$: Baseline accuracy of unsteered model $M$ on $\mathcal{D}_{\textit{dev}}$
        \State $s_{\ell} \in \mathbb{R}^+$: Layer-wise activation scaling factor
        \State $\alpha_{\textit{init}} \in \mathbb{R}^+$: Initial steering magnitude ($0.4$)
        \State $\Lambda$: Search space for steering magnitudes ($\{0.05, 0.10, \dots, 0.75\}$)
    \vspace{1ex}
    \Ensure 
        \State $S_{\textit{final}}$: Ranked list of top-10 impactful latents with optimal steering parameters

    \State \textbf{1. Initial Screening (Fixed Magnitude)}
    \For{\text{each } $k \in S$}
        \State $A_k \leftarrow \text{Evaluate}(M(\cdot \oplus_{\ell} (c \cdot \alpha_{\textit{init}} \cdot s_{\ell} \mathbf{\Phi}_k)), \mathcal{D}_{\textit{dev}})$
    \EndFor
    
    \State $\mathcal{C} \leftarrow \{k \in S \mid A_k > A_{\textit{base}}\}$
    \State Let $L_{\textit{screen}}$ be the sequence of indices in $\mathcal{C}$ sorted by $A_k$ \textbf{descending}
    \State $S_{\textit{top40}} \leftarrow L_{\textit{screen}}[0:39]$
    \vspace{1ex}
    \State \textbf{2. Hyperparameter Optimization (Grid Search)}
    \For{\text{each } $k \in S_{\textit{top40}}$}
        \State $\alpha^*_k \leftarrow \arg\max_{\alpha \in \Lambda} \text{Evaluate}(M(\cdot \oplus_{\ell} (c \cdot \alpha \cdot s_{\ell} \mathbf{\Phi}_k)), \mathcal{D}_{\textit{dev}})$
        \State $A^*_k \leftarrow \text{Evaluate}(M(\cdot \oplus_{\ell} (c \cdot \alpha^*_k \cdot s_{\ell} \mathbf{\Phi}_k)), \mathcal{D}_{\textit{dev}})$
    \EndFor
    \vspace{1ex}
    \State \textbf{3. Final Ranking}
    \State Let $L_{\textit{final}}$ be the sequence of indices in $S_{\textit{top40}}$ sorted by $A^*_k$ \textbf{descending}
    \State $S_{\textit{final}} \leftarrow [(k, c \cdot \alpha^*_k, A^*_k) \text{ for } k \in L_{\textit{final}}[0:9]]$

    \State \Return $S_{\textit{final}}$
\end{algorithmic}
\end{algorithm}

\begin{algorithm}
\caption{Selection of Control Group (Non-Trending)\label{alg:ctrl}}
\begin{algorithmic}[1]
    \Require 
        \State $\mathcal{A} \in \mathbb{R}^{T \times P \times F \times N}$: Full activation tensor
        \State $X \in \{T, P\}$: Aggregation dimension; let $Y \in \{T, P\} \setminus \{X\}$ be the trend dimension
        \State $\mathbf{v}_Y \in \mathbb{R}^Y$: Reference vector for trend dimension
        \State $N_{\textit{control}} \in \mathbb{N}$: Number of control features to select ($10$)
        \State $\alpha_{\textit{sig}} \in (0, 1)$: Significance threshold ($0.05$)
    \vspace{1ex}
    \Ensure 
        \State $S_{\textit{control}}$: List of stable, non-trending control feature indices

    \State Define set of all feature indices $\Omega = \{0, \dots, F-1\}$
    \vspace{1ex}
    \State \textbf{1. Global Aggregation}
    \State Fold prompt dimension: $\mathbf{M}_{\textit{mean}} \in \mathbb{R}^{T \times P \times F} \leftarrow \text{MeanFold}(\mathcal{A}, \text{dim}=N)$
    \State Fold aggregation dimension: $\bar{\mathbf{M}} \in \mathbb{R}^{Y \times F} \leftarrow \text{MeanFold}(\mathbf{M}_{\textit{mean}}, \text{dim}=X)$
    \vspace{1ex}
    \State \textbf{2. Trend Identification} \tikzmark{start1}
    \State $\boldsymbol{\rho}, \boldsymbol{P}_{\rho} \in \mathbb{R}^F \leftarrow \text{SpearmanFold}(\bar{\mathbf{M}}, \text{dim}=Y, \mathbf{v}_Y)$
    \State $\boldsymbol{\tau}, \boldsymbol{P}_{\tau} \in \mathbb{R}^F \leftarrow \text{MKFold}(\bar{\mathbf{M}}, \text{dim}=Y)$
    
    \State Compute joint significance mask $\mathbf{Sig} \in \{0, 1\}^F$:
    \State $\mathbf{Sig} \leftarrow (\boldsymbol{P}_{\rho} < \alpha_{\textit{sig}}) \land (\boldsymbol{P}_{\tau} < \alpha_{\textit{sig}})$

    \State Compute directional validity masks $\in \{0, 1\}^F$:
    \State $\mathbf{V}^{\uparrow} \leftarrow \mathbf{Sig} \land (\boldsymbol{\rho} > 0) \land (\boldsymbol{\tau} > 0)$
    \State $\mathbf{V}^{\downarrow} \leftarrow \mathbf{Sig} \land (\boldsymbol{\rho} < 0) \land (\boldsymbol{\tau} < 0)$
    
    \State Define global trending set: $\mathcal{T} \leftarrow \{f \in \Omega \mid \mathbf{V}^{\uparrow}_d = 1 \lor \mathbf{V}^{\downarrow}_d = 1\}$ \tikzmark{end1}
    \vspace{1ex}
    \State \textbf{3. Stable Latents Selection} \tikzmark{start2}
    \State Compute global magnitude vector $\boldsymbol{\delta} \in \mathbb{R}^F$:
    \State $\boldsymbol{\delta} \leftarrow |\bar{\mathbf{M}}[Y-1, :] - \bar{\mathbf{M}}[0, :]|$

    \State Mask trending features out of $\boldsymbol{\delta}$ to force them out of the stable selection:
    \State $\tilde{\boldsymbol{\delta}}_d = \begin{cases} \boldsymbol{\delta}_d & \text{if } d \notin \mathcal{T} \\ +\infty & \text{otherwise} \end{cases}$

    \State Extract stable indices $\in \{0, \dots, F-1\}^{N_{\textit{control}}}$:
    \State $S_{\textit{control}} \leftarrow \text{TopKIndices}(\tilde{\boldsymbol{\delta}}, \text{dim}=F, \text{largest}=\text{False}, k=N_{\textit{control}})$ \tikzmark{end2}

    \State \Return $S_{\textit{control}}$

\end{algorithmic}

\begin{tikzpicture}[overlay, remember picture]
    \pgfmathsetmacro{\BraceX}{\linewidth - 60} 

    \draw[decorate, decoration={brace, amplitude=5pt}, thick]
        (\BraceX pt, 0 |- start1) -- (\BraceX pt, 0 |- end1)
        node[midway, right=8pt, align=left, text width=3.0cm]
        {\footnotesize Exclude \\ features \\ with \\ significant \\ trends};

    \draw[decorate, decoration={brace, amplitude=5pt}, thick]
        (\BraceX pt, 0 |- start2) -- (\BraceX pt, 0 |- end2)
        node[midway, right=8pt, align=left, text width=3.5cm]
        {\footnotesize Select \\ features \\ with \\ minimal \\ activation \\ change};
\end{tikzpicture}
\end{algorithm}
\clearpage

\section{Qualitative Analysis of Top Latents}
\label{appendix:qualitative}
For one top latent of each model we show a single example fact, all for relation P17 (country). Each card gives the latent's per-token SAE activation over the prompt (background shading) and the predicted answer for the pre-trained baseline $M_{D_0}$, the fine-tuned model $M_{D_{100}}$, and the steered model $M_{D_{100}}^{*}$ (gold-accented row), with a check or cross marking each prediction against the gold answer. In each case a fact $M_{D_{100}}$ answers incorrectly is restored by single-latent steering. The remaining $M_{D_{100}}^{*}$ rows steer the model's other top latents and are shown only for comparison.

\begin{center}
\includegraphics[width=\textwidth,height=0.25\textheight,keepaspectratio]{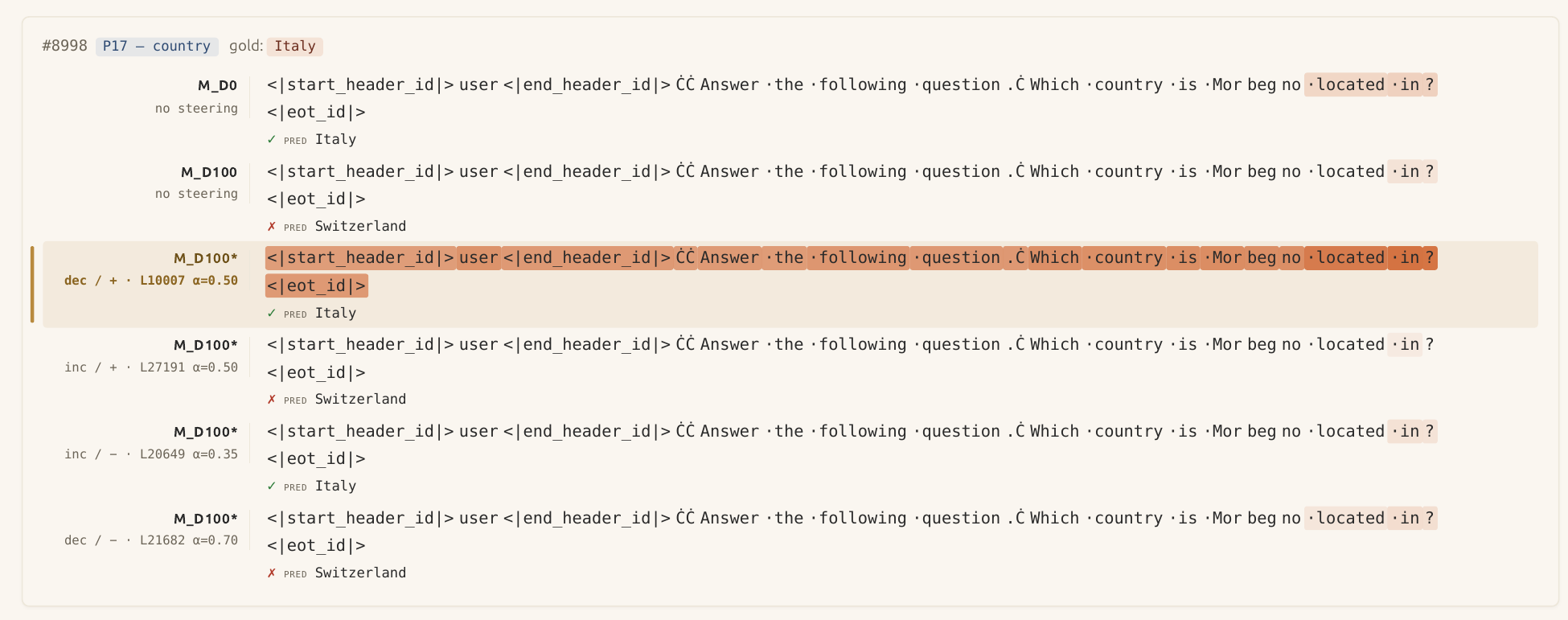}\\[1pt]
{\small\textbf{Llama 3.1 8B}, latent L10007 (decreasing, positive steering)}\\[6pt]
\includegraphics[width=\textwidth,height=0.25\textheight,keepaspectratio]{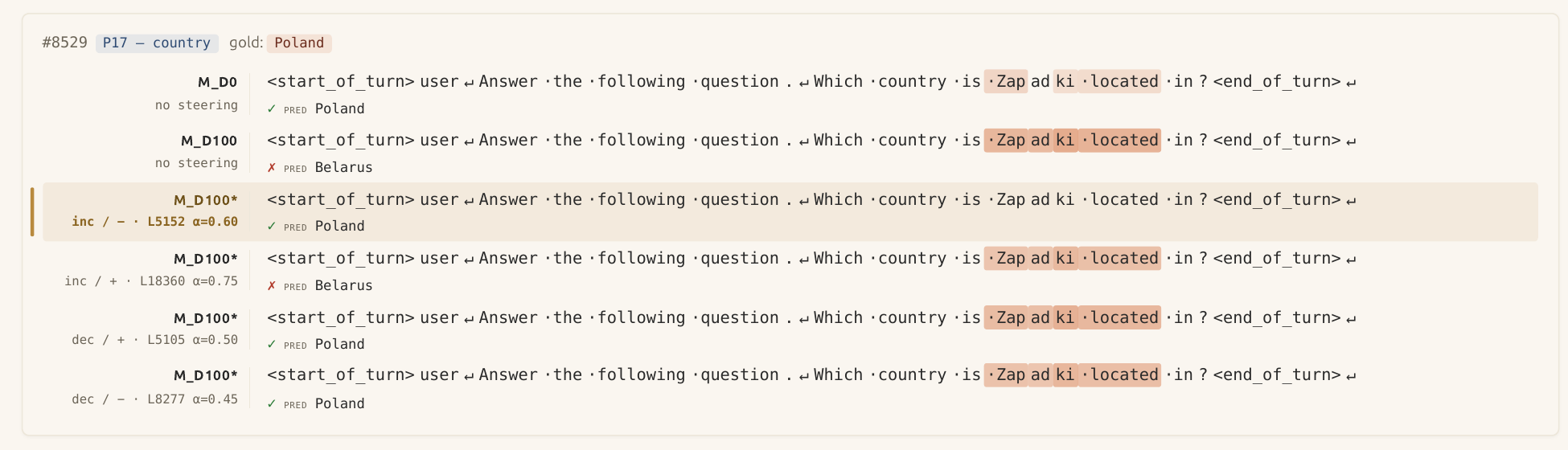}\\[1pt]
{\small\textbf{Gemma 2 9B}, latent L5152 (increasing, negative steering)}\\[6pt]
\includegraphics[width=\textwidth,height=0.25\textheight,keepaspectratio]{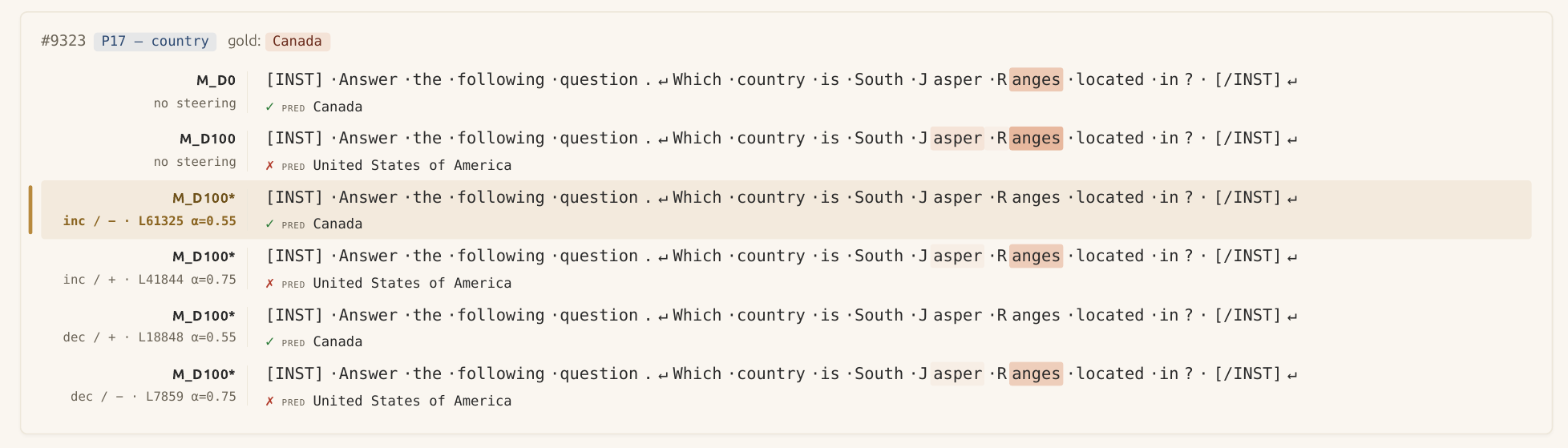}\\[1pt]
{\small\textbf{Mistral 7B v0.3}, latent L61325 (increasing, negative steering)}
\end{center}

\clearpage
\section{Steering Recovery Across Training}
\label{appendix:trajectory}

\begin{figure}[ht]
  \centering
  \includegraphics[width=\linewidth]{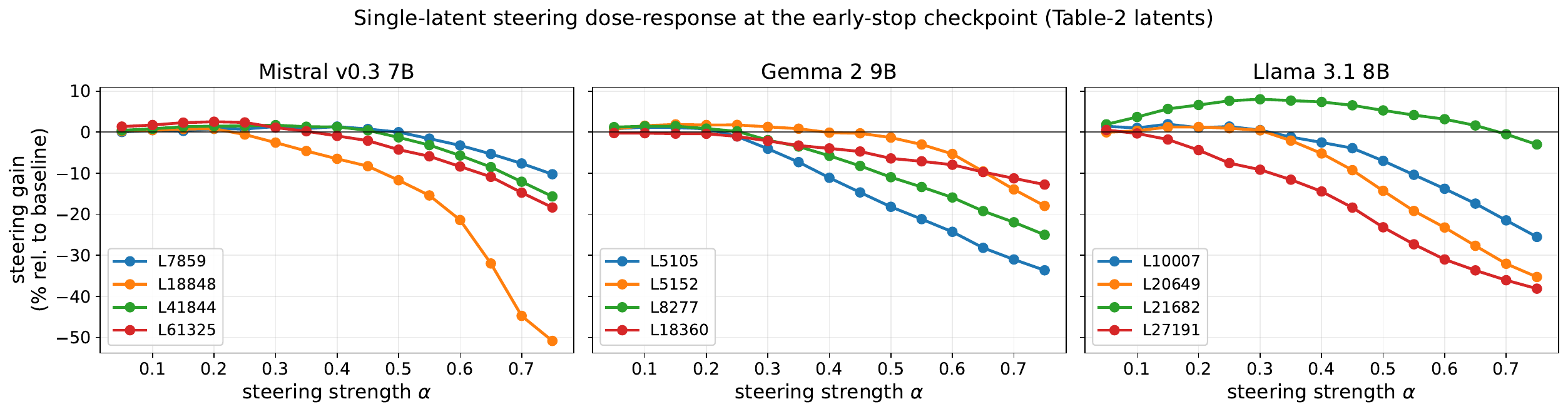}
  \caption{\textbf{Steering recovery vs.\ steering strength at the early-stop checkpoints (dev set).}
  Relative accuracy gain over the unsteered checkpoint as a function of steering
  strength $\alpha$, for the four Table~\ref{tab:cross_model_knowledge_recovery} ($\uparrow$Unknown) latents of
  each model, steered at their reported polarity. The horizontal line marks the unsteered early-stop baseline. The smooth,
  single-peaked curves show that the directions identified by MoRFI recover known facts
  even at the early-stop (non-overfit) checkpoints.}
  \label{fig:estop-strength}
\end{figure}

\begin{figure}[ht]
  \centering
  \includegraphics[width=\linewidth]{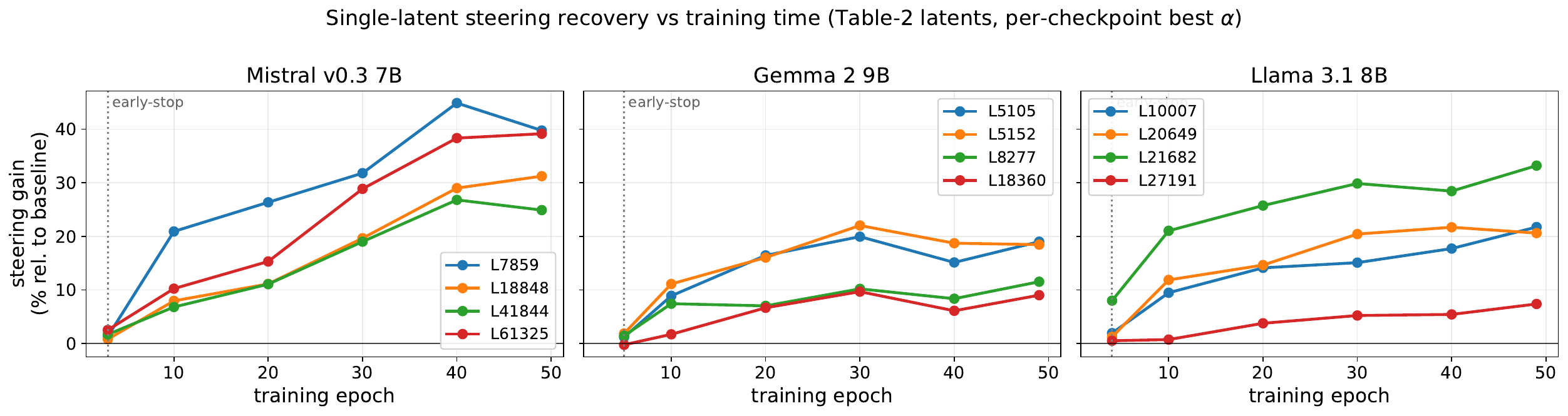}
\caption{\textbf{Single-latent steering recovery across training time (dev set).}
  For each model, the four Table~\ref{tab:cross_model_knowledge_recovery} ($\uparrow$Unknown) latents are steered
  at their reported polarity, with steering strength fit per checkpoint as in
  Algorithm~2. Each line shows one latent's relative accuracy gain over the
  unsteered checkpoint versus training epoch; the dotted line marks the early-stop epoch.
  The recovery is already present at early stopping and increases with continued
  training, approaching the converged (50-epoch) gains of Table~\ref{tab:cross_model_knowledge_recovery} (e.g., Llama
  latent~21682: $+8.0\%$ at early stop, rising toward $+33.4\%$ at convergence). The
  effect is thus a continuum across fine-tuning, growing with the amount of training.}
  \label{fig:estop-traj}
\end{figure}

\begin{figure}[ht]
  \centering
  \includegraphics[width=\linewidth]{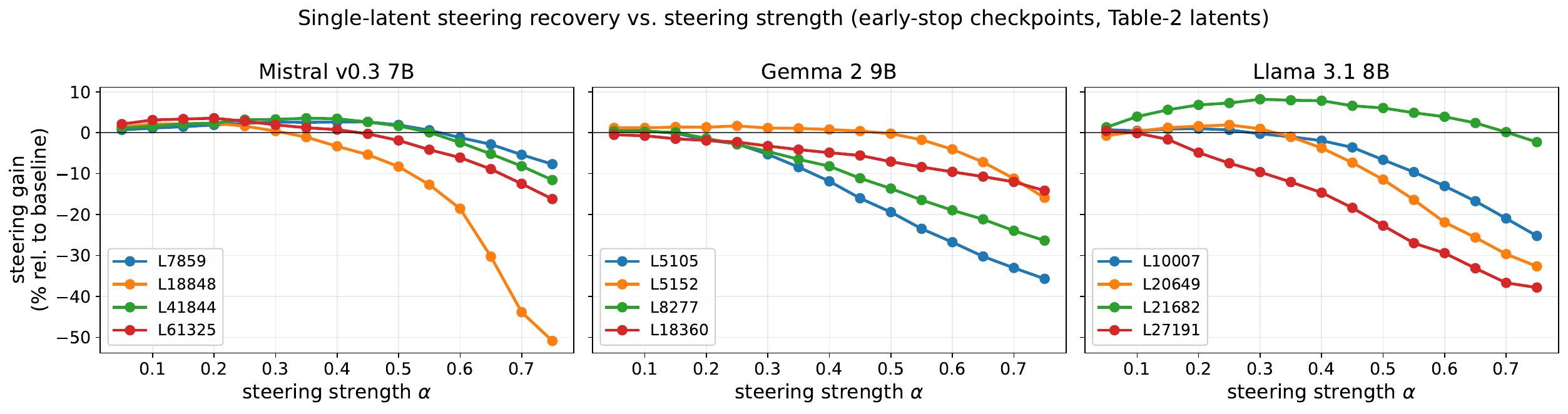}
  \caption{Test-split counterpart of Figure~\ref{fig:estop-strength}: relative accuracy gain versus steering strength $\alpha$ at the early-stop checkpoints, evaluated on the held-out test split.}
  \label{fig:estop-strength-test}
\end{figure}

\begin{figure}[ht]
  \centering
  \includegraphics[width=\linewidth]{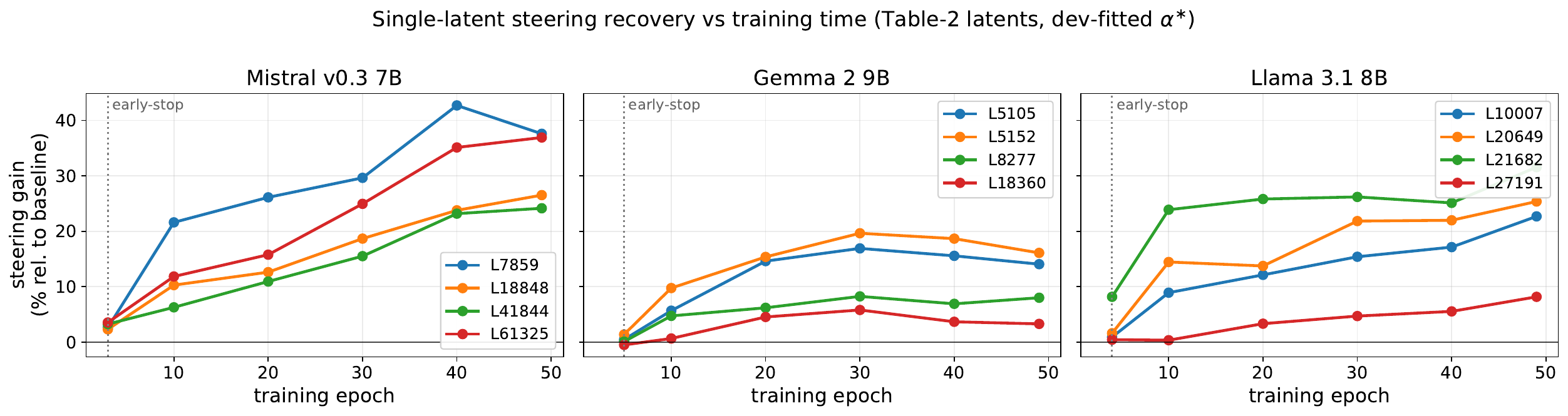}
  \caption{Test-split counterpart of Figure~\ref{fig:estop-traj}: single-latent steering recovery across training at the dev-fitted per-checkpoint strengths, evaluated once on the held-out test split.}
  \label{fig:estop-traj-test}
\end{figure}

\paragraph{Held-out evaluation} The gains above are not an artifact of selecting latents and fitting steering strengths on the dev set. Keeping the per-checkpoint strengths fitted on dev and evaluating each configuration once on the held-out test split (10{,}481 questions), 71 of 72 latent-checkpoint pairs remain positive, and all twelve latents are positive at the converged checkpoint, with per-model mean recoveries of +21.9\%/+10.4\%/+31.3\% (Llama/Gemma/Mistral) against +20.7\%/+14.5\%/+33.8\% on dev. The single exception is Gemma latent 18360 at its early-stop epoch ($-0.5\%$), the same latent that is near zero on dev at that epoch. Figures~\ref{fig:estop-strength-test} and~\ref{fig:estop-traj-test} mirror the two figures above on the test split.

\clearpage

\section{Steering Intervention Overshoot}
\label{appendix:overshoot}
The optimal steering strengths $\alpha^{*}$ produce interventions whose magnitude $\|\alpha^{*} s_{\ell} \mathbf{\Phi}_k\|$ exceeds the fine-tuning displacement $\|a_{D_0} - a_{D_{100}}\|$ at the injection (middle) layer $L_\mathrm{inj}$, where $a_{D_0}$ and $a_{D_{100}}$ are the mean middle-layer activations of $M_{D_0}$ and $M_{D_{100}}$ over the evaluation set. This stems from how the layer-wise scaling factor $s_{\ell}$ is calibrated. It is computed on a reference dataset (Alpaca) to give a common steering-strength unit across models, so $\alpha^{*} s_{\ell}$ carries an offset relative to the dev set activation scale and pushes activations beyond the $M_{D_0}$ level rather than restoring them exactly. This does not appear to affect knowledge recovery, consistent with self-repair-like behaviour downstream of the middle layer, and we flag a direct measurement of the overshoot as future work.

\begin{table}[!ht]
\centering
\resizebox{\textwidth}{!}{%
\begin{tabular}{@{}ll l ccc c@{}}
\toprule
\textbf{Direction} & \textbf{Feature Trend} & \textbf{Intervention} & \textbf{Top Latent} & $\boldsymbol{\alpha}$ & $\boldsymbol{\|\alpha s_{\ell} \mathbf{\Phi}_k\|}$ & \textbf{Overshoot} \\ \midrule
\multirow{15}{*}{\textbf{$\uparrow$Unknown}} & \multicolumn{6}{c}{\textbf{Llama 3.1 8B} \quad (Baseline: $\|a_{D_0} - a_{D_{100}}\|$ at $L_\mathrm{inj}$ = 2.03)} \\ \cmidrule(l){2-7}
& \multirow{2}{*}{\textbf{Increasing}} & Positive (+) & \href{https://www.neuronpedia.org/llama3.1-8b/15-llamascope-res-32k/27191}{27191} & 0.50 & 5.55 & $\mathbf{2.75\times}$ \\
&  & Negative (-) & \href{https://www.neuronpedia.org/llama3.1-8b/15-llamascope-res-32k/20649}{20649} & 0.35 & 3.89 & $\mathbf{1.93\times}$ \\
\cmidrule(l){2-7}
& \multirow{2}{*}{\textbf{Decreasing}} & Positive (+) & \href{https://www.neuronpedia.org/llama3.1-8b/15-llamascope-res-32k/10007}{10007} & 0.50 & 5.56 & $\mathbf{2.76\times}$ \\
&  & Negative (-) & \href{https://www.neuronpedia.org/llama3.1-8b/15-llamascope-res-32k/21682}{21682} & 0.70 & 7.78 & $\mathbf{3.86\times}$ \\
\cmidrule(l){2-7}
& \multicolumn{6}{c}{\textbf{Gemma 2 9B} \quad (Baseline: $\|a_{D_0} - a_{D_{100}}\|$ at $L_\mathrm{inj}$ = 45.29)} \\ \cmidrule(l){2-7}
& \multirow{2}{*}{\textbf{Increasing}} & Positive (+) & 18360 & 0.75 & 170.34 & $\mathbf{3.78\times}$ \\
&  & Negative (-) & 5152 & 0.60 & 136.25 & $\mathbf{3.02\times}$ \\
\cmidrule(l){2-7}
& \multirow{2}{*}{\textbf{Decreasing}} & Positive (+) & 5105 & 0.50 & 113.54 & $\mathbf{2.52\times}$ \\
&  & Negative (-) & 8277 & 0.45 & 102.19 & $\mathbf{2.27\times}$ \\
\cmidrule(l){2-7}
& \multicolumn{6}{c}{\textbf{Mistral v03 7B} \quad (Baseline: $\|a_{D_0} - a_{D_{100}}\|$ at $L_\mathrm{inj}$ = 1.49)} \\ \cmidrule(l){2-7}
& \multirow{2}{*}{\textbf{Increasing}} & Positive (+) & 41844 & 0.75 & 6.02 & $\mathbf{4.09\times}$ \\
&  & Negative (-) & 61325 & 0.55 & 4.42 & $\mathbf{3.00\times}$ \\
\cmidrule(l){2-7}
& \multirow{2}{*}{\textbf{Decreasing}} & Positive (+) & 18848 & 0.55 & 4.42 & $\mathbf{3.00\times}$ \\
&  & Negative (-) & 7859 & 0.75 & 6.02 & $\mathbf{4.09\times}$ \\
\bottomrule
\end{tabular}%
}
\caption{Steering intervention overshoot at $L_\mathrm{inj}$ for the Table~\ref{tab:cross_model_knowledge_recovery} latents. Overshoot $= \|\alpha s_{\ell} \mathbf{\Phi}_k\| / \|a_{D_0} - a_{D_{100}}\|$ at the middle layer, averaged over the evaluation dataset. All 12 latents overshoot the fine-tuning displacement, with mean ratios in $[1.93, 4.09]$.}
\label{tab:overshoot}
\end{table}

\section{Difference-in-Mean Baseline Comparison}
\label{appendix:baseline}
We compare MoRFI's selection against ranking latents by their difference-in-mean activation between $M_{D_0}$ and $M_{D_{100}}$. Table~\ref{tab:baseline_overlap} reports the selection overlap at matched set sizes, where $k$ is the size of MoRFI's $C/R{=}1$ set for each model and direction. Every latent MoRFI selects has the expected-sign difference-in-mean score, but the two criteria agree on only 18--41\% of selections. MoRFI keeps lower-magnitude latents that track the property monotonically and discards high-magnitude latents whose trajectory is non-monotonic or unstable across resamples.

\begin{table}[!ht]
\centering
\begin{tabular}{@{}lccccc@{}}
\toprule
\textbf{Model} & \textbf{Direction} & $k$ & \textbf{In baseline top-}$k$ & \textbf{Overlap} & \textbf{Jaccard} \\
\midrule
Llama 3.1 8B & $\uparrow$ & 160 & 28 & 18\% & 0.10 \\
Llama 3.1 8B & $\downarrow$ & 261 & 65 & 25\% & 0.14 \\
Gemma 2 9B & $\uparrow$ & 713 & 293 & 41\% & 0.26 \\
Gemma 2 9B & $\downarrow$ & 463 & 172 & 37\% & 0.23 \\
Mistral 7B v0.3 & $\uparrow$ & 642 & 199 & 31\% & 0.18 \\
Mistral 7B v0.3 & $\downarrow$ & 611 & 198 & 32\% & 0.19 \\
\bottomrule
\end{tabular}
\caption{Selection overlap between the difference-in-mean baseline and MoRFI at matched set sizes for the $\uparrow$Unknown property, where $k$ is the size of MoRFI's $C/R{=}1$ set per model and direction.}
\label{tab:baseline_overlap}
\end{table}

To test whether endpoint magnitude identifies causally effective latents, we steer the baseline's forty highest-ranked latents per direction among those MoRFI rejects, its most confident disagreements with our selection, on $M_{D_{100}}$ (50 epochs) at the uniform strength $\alpha=0.4$ of Algorithm~\ref{alg:idf}'s screening stage, in both polarities, on the dev set. Each combination of model, trend and steering polarity forms a cell of forty latents, giving twelve cells and 480 evaluations in total. Recovery is the relative accuracy gain over the unsteered checkpoint, following the convention of Tables~\ref{tab:llama31_8b_unk}--\ref{tab:cross_model_knowledge_recovery}. Only 11\%, 24\%, and 33\% of these latents (Llama, Gemma, and Mistral) improve dev accuracy at all (Table~\ref{tab:baseline_screen}), and the mean effect over all forty latents in a cell is negative in 11 of the 12 cells (Table~\ref{tab:baseline_cells}). At the same strength, MoRFI's top validated latent outperforms the strongest discarded latent in 11 of 12 cells (Table~\ref{tab:baseline_cells}). Conversely, five of the twelve Table~\ref{tab:cross_model_knowledge_recovery} latents, including three of Gemma's four, fall outside the baseline's top-$k$ altogether.

\begin{table}[!ht]
\centering
\begin{tabular}{@{}lcccc@{}}
\toprule
\textbf{Model} & \textbf{Impr.\ (of 160)} & \textbf{Mean (impr.)} & \textbf{MoRFI @ $\alpha{=}0.4$} & \textbf{Tuned} \\
\midrule
Llama 3.1 8B & 17 (11\%) & +4.4\% & +18.6\% & +20.7\% \\
Gemma 2 9B & 39 (24\%) & +1.3\% & +11.7\% & +14.5\% \\
Mistral 7B v0.3 & 52 (33\%) & +4.0\% & +22.7\% & +33.8\% \\
\bottomrule
\end{tabular}
\caption{Screening the baseline's top-40 discarded latents per trend-polarity cell at $\alpha=0.4$ on $M_{D_{100}}$, aggregated per model (160 latents each). ``Impr.'' counts latents exceeding the unsteered baseline and ``Mean (impr.)'' averages recovery over them; MoRFI columns give the mean recovery of the Table~\ref{tab:cross_model_knowledge_recovery} latents at the same screening strength and at their tuned strengths.}
\label{tab:baseline_screen}
\end{table}

\begin{table}[!ht]
\centering
\setlength{\tabcolsep}{4pt}
\resizebox{\textwidth}{!}{%
\begin{tabular}{@{}ll ccccc c@{}}
\toprule
\textbf{Trend} & \textbf{Steer} & \textbf{Impr.\ /40} & \textbf{Mean (impr.)} & \textbf{Mean (all 40)} & \textbf{Best discarded} & \textbf{MoRFI @ $\alpha{=}0.4$} & \textbf{Tuned} \\ \midrule
\multicolumn{8}{c}{\textbf{Llama 3.1 8B} \quad (unsteered baseline acc.\ 0.1785)} \\ \midrule
Increasing & $+$ & 6 & +3.0\% & $-$10.3\% & 6284 (+7.3\%) & 27191 (+7.2\%) & +7.1\% \\
Increasing & $-$ & 5 & +8.7\% & $-$7.3\% & 32644 (+17.2\%) & 20649 (+20.5\%) & +20.6\% \\
Decreasing & $+$ & 4 & +3.3\% & $-$10.3\% & 22975 (+5.2\%) & 10007 (+19.2\%) & +21.7\% \\
Decreasing & $-$ & 2 & +2.7\% & $-$7.0\% & 13122 (+4.8\%) & 21682 (+27.6\%) & +33.4\% \\ \midrule
\multicolumn{8}{c}{\textbf{Gemma 2 9B} \quad (unsteered baseline acc.\ 0.1905)} \\ \midrule
Increasing & $+$ & 10 & +1.1\% & $-$2.1\% & 16087 (+2.9\%) & 18360 (+4.6\%) & +9.0\% \\
Increasing & $-$ & 11 & +1.7\% & $-$1.7\% & 23053 (+5.3\%) & 5152 (+13.9\%) & +18.5\% \\
Decreasing & $+$ & 11 & +1.3\% & $-$1.9\% & 26009 (+3.7\%) & 5105 (+17.9\%) & +19.0\% \\
Decreasing & $-$ & 7 & +1.0\% & $-$2.5\% & 13826 (+3.3\%) & 8277 (+10.5\%) & +11.5\% \\ \midrule
\multicolumn{8}{c}{\textbf{Mistral 7B v0.3} \quad (unsteered baseline acc.\ 0.1570)} \\ \midrule
Increasing & $+$ & 6 & +3.9\% & $-$10.2\% & 43531 (+8.1\%) & 41844 (+16.0\%) & +24.9\% \\
Increasing & $-$ & 18 & +5.9\% & +0.1\% & 26824 (+20.8\%) & 61325 (+35.4\%) & +39.2\% \\
Decreasing & $+$ & 13 & +3.2\% & $-$4.8\% & 61796 (+10.2\%) & 18848 (+26.7\%) & +31.2\% \\
Decreasing & $-$ & 15 & +2.9\% & $-$2.7\% & 5199 (+7.4\%) & 7859 (+13.0\%) & +39.8\% \\
\bottomrule
\end{tabular}%
}
\caption{Per-cell results of the $\alpha=0.4$ screen. ``Impr.\ /40'' counts the baseline's top-40 discarded latents that exceed the unsteered baseline, and the means give their relative recovery over the improvers and over all 40; ``Best discarded'' and ``MoRFI @ $\alpha{=}0.4$'' give the strongest latent of each set at the screening strength, with MoRFI's tuned (Table~\ref{tab:cross_model_knowledge_recovery}) recovery in the last column.}
\label{tab:baseline_cells}
\end{table}

\end{document}